\documentclass{article}

 \usepackage[main, final]{neurips_2026}

\usepackage[utf8]{inputenc} 
\usepackage[T1]{fontenc}    
\usepackage{hyperref}       
\usepackage{url}            
\usepackage{booktabs}       
\usepackage{amsfonts}       
\usepackage{nicefrac}       
\usepackage{microtype}      
\usepackage{xcolor}         

\usepackage{amsmath} 
\usepackage{graphicx} 
\usepackage{makecell}
\usepackage{subcaption}
\usepackage{multirow}
\usepackage{wrapfig}
\usepackage{tcolorbox}

\tcbuselibrary{skins}      
\definecolor{findborder}{HTML}{14403B} 
\definecolor{findbg}{HTML}{F2F2F2}     
\newtcolorbox{findingbox}{
    enhanced,                 
    colframe=findborder,       
    colback=findbg,            
    boxrule=1pt,               
    arc=4pt,                   
    drop shadow=black!30,     
    left=8pt,                  
    right=8pt,                 
    top=6pt,                   
    bottom=6pt,                
    before skip=12pt,          
    after skip=12pt            
}

\definecolor{candbg}{RGB}{200, 220, 240}    
\definecolor{qbg}{RGB}{210, 230, 200}       
\definecolor{refbg}{RGB}{245, 215, 215}     
\definecolor{vidbg}{RGB}{200, 220, 240}    
\definecolor{audbg}{RGB}{255, 225, 195}    
\definecolor{sysbg}{RGB}{215, 235, 200}    
\definecolor{qbg}{RGB}{180, 215, 175}      
\definecolor{visanchor}{RGB}{200, 220, 240}    
\definecolor{audanchor}{RGB}{255, 225, 195}    
\definecolor{mixanchor}{RGB}{230, 215, 240}    
\definecolor{drop}{RGB}{200, 50, 50}
\definecolor{gain}{RGB}{50, 130, 50}

\definecolor{gold}{RGB}{200,150,0}
\definecolor{purpleblue}{HTML}{4A4FC4}   
\definecolor{ibotgreen}{rgb}{0.204,0.659,0.325}
\hypersetup{colorlinks=true, linkcolor=red, citecolor=ibotgreen, urlcolor=purpleblue}

\usepackage{hyperref}
\usepackage{fontawesome5}
\newcommand{\iconlink}[3]{\href{#1}{#2\hspace{0.3em}#3}}

\title{From Senses to Decisions: The Information Flow of Auditory and Visual Perception in Multimodal LLMs}

\author{%
  \textbf{Wish Suharitdamrong}$^{1}$\thanks{Corresponding author} \And
  \textbf{Muhammad Awais}$^{1,2}$ \And
  \textbf{Xiatian Zhu}$^{1,2}$ \And
  \textbf{Sara Atito}$^{1,2}$  \And
  \\
\\[-3.5em]
  \textnormal{$^1$Surrey Institute for People-Centred AI (PAI), University of Surrey, UK} \\
  \textnormal{$^2$Centre for Vision, Speech and Signal Processing (CVSSP), University of Surrey, UK} \\
  \texttt{\{w.suharitdamrong,xiatian.zhu,muhammad.awais,sara.atito\}@surrey.ac.uk}
}

\begin{document}

\maketitle

\vspace{-1.5em}
\begin{center}

  \iconlink{https://github.com/peterwisu/senses2decision}{\faGithub}{\href{https://github.com/peterwisu/senses2decision}{Code : https://github.com/peterwisu/senses2decision}}
\end{center}
\vspace{0.5em}

\begin{abstract}
Multimodal Large Language Models (MLLMs) can listen and see, but how do audio and visual signals actually travel through the network to shape an answer? Despite their growing role in research and real-world applications, the internal pathways through which audio and visual tokens influence the final prediction remain poorly understood. In this study, we examine audio-visual information flow inside Audio-Visual Large Language Models (AVLLMs), tracing how AVLLMs route, utilize, and integrate audio and visual information across two input configurations, audio-visual video and multiple interleaved audio-visual items. We find that for audio-visual video, AVLLMs follow the sequential information flow pathway established for VLMs and VideoLLMs, with audio and visual contribution flowing along this pathway in proportion to the task's reliance on each modality. In settings with multiple interleaved audio-visual items, this routing shifts to different parallel streams. Furthermore, we demonstrate that audio-visual and other token types can be discarded once their information is transferred to LLM, with minimal impact on the model's prediction or even slight improvement, generalizing across multiple tasks and datasets, enabling more efficient inference. These findings hold across multiple models and scales, Qwen2.5-Omni and Video-SALMONN2 Plus at 3B and 7B scales, leading to hypotheses on why these flow structures emerge. Together, these results deliver the first coherent picture of how AVLLMs orchestrate sound and sight inside the network and lay the groundwork for the next wave of interpretability, design, and efficiency advances in audio-visual and broader MLLMs.

\end{abstract}

\section{Introduction}
Multimodal Large Language Models (MLLMs)~\cite{team2023gemini,hurst2024gpt} have progressed rapidly, jointly processing auditory and visual information in models that can both listen and see, bringing machine perception closer to human perception. Earlier research developed each modality independently, leading to specialized Vision-Language models (VLMs)~\cite{liu2024improved,li2024llava,an2025llava,bai2025qwen3,tong2024cambrian,zhang2025videollama,wang2025internvl3} and Audio-Language models (ALMs)~\cite{gong2023listen,tang2023salmonn,ghosh2024gama,goel2025audio,chu2024qwen2,alex2025pal}, each operating effectively within its target modality. Recent Audio-Visual Large Language Models (AVLLMs)~\cite{xu2025qwen2,tang2025video,xu2025qwen3,fu2025vita,cheng2024videollama,team2026qwen3} integrate visual and auditory inputs to enable unified audio-visual understanding. These models can answer questions about audio-visual scenes and transcribe visually grounded speech, tasks requiring cross-modal reasoning across the audio and visual modalities. These models span input formats from single images, videos, or audio clips to audio-visual videos and multiple interleaved audio-visual items, reaching diverse real-world scenarios. Around these models, an active research landscape has emerged, including benchmarks probing audio-visual understanding across rich and complex scenarios~\cite{yang2025audio,li2025omnivideobench, zhou2025daily, li2024omnibench}, parameter-efficient fine-tuning methods~\cite{wei2025moka}, adaptation-based token compression~\cite{gong2025echoingpixels, ding2026omnisift}, and training-free token compression at inference~\cite{tao2025omnizip, li2026dash}.

In parallel, mechanistic interpretability has made significant progress in uncovering the internal mechanisms inside LLMs~\cite{nanda2023progress, elhage2021mathematical, rai2024practical, geva2023dissecting}. Similar techniques have recently been extended to MLLMs~\cite{basu2024understanding, nikankin2025same, neo2024towards, zhang2025cross, kim2025map, kaduri2025s, selvakumar2026audio}. In particular, attention knockout has been used to trace how cross-modal information flow emerges from image inputs in VLMs~\cite{zhang2025cross} and how spatiotemporal information flow emerges from video inputs in VideoLLMs~\cite{kim2025map}. Beyond these mechanistic studies, multi-image input handling has been actively studied in VLMs, surfacing failure modes and motivating mitigation strategies. Cross-image information leakage has been identified as a core failure mode, where visual content from different images entangles in the output~\cite{park2025mitigating}. Delimiter tokens have been examined and exploited as a mechanism to limit this entanglement~\cite{lee2026enhancing}. Performance on multi-image tasks has also been shown to degrade as the number of input images grows~\cite{das2026more}.

While AVLLMs introduce a new dimension to machine perception through the integration of sound and sight, the internal mechanisms underlying audio-visual integration remain largely unstudied. Concurrent work on AVLLMs~\cite{selvakumar2026audio} examines audio-visual captioning and reports that cross-modal integration concentrates in deep layers. In contrast, information flow studies in VLMs~\cite{zhang2025cross} and VideoLLMs~\cite{kim2025map} locate cross-modal integration at earlier-to-middle layers, where visual information flows to the prediction only through the language tokens. Whether the information flow in AVLLMs aligns with these VLM and VideoLLM findings or departs from them, and how AVLLMs distribute their reliance on audio versus visual inputs along this flow, remains an open question. In particular, no prior work has examined the role of audio along the information flow in MLLMs, and it is unclear whether audio behaves similarly to visual information or follows different pathways. Additionally, in the multi-input interleaved configuration, prior work in VLMs has characterized model behavior on multi-image inputs~\cite{park2025mitigating, lee2026enhancing, das2026more}, but the underlying information flow has not been examined, neither for multi-image inputs nor for the broader case where audio items are interleaved alongside visual items. In this study, we trace how audio and visual information jointly flow through AVLLMs to form the prediction, mapping these pathways in both input configurations and characterizing how each modality contributes along the way. Our key findings are as follows:
\begin{itemize}
    \item \textbf{Audio-visual information does not reach the deep layers:} Video attention in later layers is dominated by attention artifacts that disproportionately attract attention, making attention allocation an unreliable indicator of information flow.
        
    \item \textbf{Task requirements steer the model's audio-visual flow:} The contribution of each modality to the prediction and the strength of interaction between audio and video vary with what the task requires, with more visual, auditory, or audio-visual alignment content depending on which is needed to answer the question.
        
    \item \textbf{Multiple independent audio-visual inputs route through parallel paths:} Independent audio and visual items interleaved with text route information to the prediction along multiple parallel paths, rather than through a single sequential path as in single audio-visual videos.
    
    \item \textbf{Tokens can be discarded after their information is transferred:} Once a token's content has been passed on, it can be discarded with minimal impact on accuracy or even slight improvement. We show this across tasks and datasets, and across input configurations, with each token type discarded at the distinct layer where its information transfer completes.
\end{itemize}

\section{Preliminary on audio-visual large language models (AVLLMs)}\label{sec:preliminary}

\paragraph{Multimodal tokenization and sequence construction:}
AVLLMs process a video with its audio track and a text instruction through an autoregressive transformer over an interleaved token sequence. Let the video frames be $\mathcal{V} \in \mathbb{R}^{T \times H \times W \times 3}$, with $T$ frames at spatial resolution $H \times W$. The frames are passed through a vision encoder and projector to produce $N_V$ video tokens of dimension $d$, the audio track is processed by an audio encoder into $N_A$ audio tokens of the same dimension, and the text instruction is tokenized into $N_T$ text tokens. For a single audio-visual video input, AVLLMs preserve temporal alignment by interleaving audio and video tokens within fixed temporal windows. Let $C$ denote the number of windows, and let $\mathbf{V}_c$ and $\mathbf{A}_c$ be the visual and audio tokens within the $c$-th window. With \colorbox{sysbg}{system prompt}, \colorbox{vidbg}{video}, \colorbox{audbg}{audio}, and \colorbox{qbg}{question} segments, the full input sequence to the language model is

\begin{equation}
\label{eq:sync-layout}
\mathcal{I} \;=\; \Big[\; \underbrace{\colorbox{sysbg}{$s_1, \ldots, s_{N_S}$}}_{\text{system}} \;;\; \underbrace{\colorbox{vidbg}{$\mathbf{V}_1$}, \colorbox{audbg}{$\mathbf{A}_1$} \;;\; \ldots \;;\; \colorbox{vidbg}{$\mathbf{V}_C$}, \colorbox{audbg}{$\mathbf{A}_C$}}_{\text{single audio-visual video}} \;;\; \underbrace{\colorbox{qbg}{$q_1, \ldots, q_{N_Q}$}}_{\text{question}} \;\Big],
\end{equation}

where $s_1, \ldots, s_{N_S}$ are system-prompt tokens, $q_1, \ldots, q_{N_Q}$ are question tokens, and the total sequence length is $N = N_S + N_V + N_A + N_Q$. Beyond this single audio-visual video setting, AVLLMs also process \emph{multi-input} sequences with multiple independent audio and visual items interleaved with text, which we describe in Section~\ref{sec:multi-input}.

\paragraph{Causal self-attention:}
At each transformer layer $\ell$, the hidden states $\mathbf{H}^{\ell} \in \mathbb{R}^{N \times d}$ are projected into query, key, and value matrices $\mathbf{Q}^{\ell} = \mathbf{H}^{\ell}\mathbf{W}_Q^{\ell}$, $\mathbf{K}^{\ell} = \mathbf{H}^{\ell}\mathbf{W}_K^{\ell}$, $\mathbf{V}^{\ell} = \mathbf{H}^{\ell}\mathbf{W}_V^{\ell}$, where $\mathbf{W}_Q^{\ell}, \mathbf{W}_K^{\ell}, \mathbf{W}_V^{\ell} \in \mathbb{R}^{d \times d_h}$ are learnable and $d_h$ is the per-head dimension. The attention output is
\begin{equation}
\mathrm{Attention}(\mathbf{Q}^{\ell}, \mathbf{K}^{\ell}, \mathbf{V}^{\ell}) \;=\; \mathrm{softmax}\!\left(\frac{\mathbf{Q}^{\ell}(\mathbf{K}^{\ell})^{\top}}{\sqrt{d_h}} + \mathbf{M}\right) \mathbf{V}^{\ell},
\end{equation}
where $\mathbf{M} \in \mathbb{R}^{N \times N}$ is a causal mask enforcing autoregressive decoding.

\section{What attention patterns reveal about information flow?}
\label{sec:sinks}

\begin{figure}[h]
  \centering
  \includegraphics[width=\linewidth]{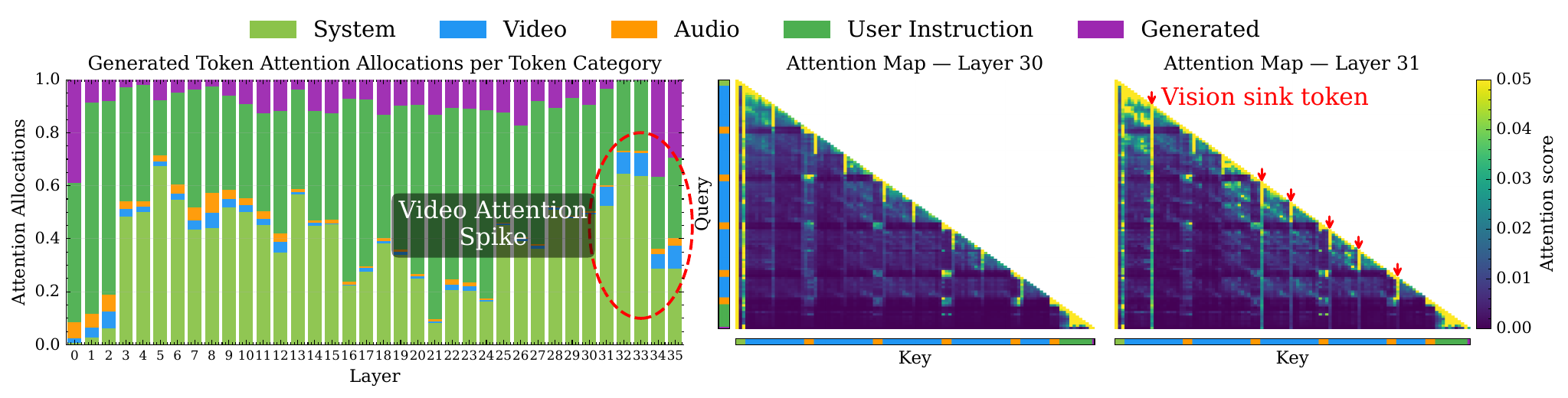}
    \caption{\textbf{Attention to video sharply spikes at layer 31.} (Left) Attention allocation of the last token per layer and token category. (Middle, Right) Attention maps at layers 30 and 31 of Qwen2.5-Omni 3B, with the vision sinks at layer 31 marked by red arrows.}
  \label{fig:attn}
\end{figure}

To trace how audio-visual information reaches the prediction, a natural starting point is to examine where the model directs its attention. We do this on multiple-choice question-answering (MCQ) tasks, where the prediction is a single token (the answer letter), using Qwen2.5-Omni~\cite{xu2025qwen2} at 3B scale as our subject model. We inspect the attention allocation of this last token which is the first generated token where the prediction is formed. Specifically, we track its allocation across layers and across token categories (\colorbox{sysbg}{system prompt}, \colorbox{vidbg}{video}, \colorbox{audbg}{audio}, \colorbox{qbg}{user instruction}). Figure~\ref{fig:attn} (left) shows that throughout most of the network, the last token attends predominantly to language tokens (\colorbox{sysbg}{system prompt} and \colorbox{qbg}{user instruction}), and attention to multimodal tokens fades through the layers. However, attention to video sharply spikes at layer 31 and remains elevated through the final layer.

\begin{wraptable}{r}{0.42\linewidth}
\centering
\small
\vspace{-1.2em}
\caption{Masking attention to video and audio tokens at later layers (31--35) leaves AV-SpeakerBench accuracy unchanged or slightly improved.}
\label{tab:sink-mask}
\setlength{\tabcolsep}{4pt}
\begin{tabular}{@{}lr@{}}
\toprule
\textbf{Mask} & \textbf{Accuracy} \\
\midrule
Original Causal Mask                  & 42.24 \\
Mask video for last token             & 42.24 \\
Mask video for all text               & 42.31 \\
Mask video and audio for all text     & 42.52 \\
\bottomrule
\end{tabular}
\vspace{-1em}
\end{wraptable}

\begin{figure}[h]
  \centering
  \includegraphics[width=\linewidth]{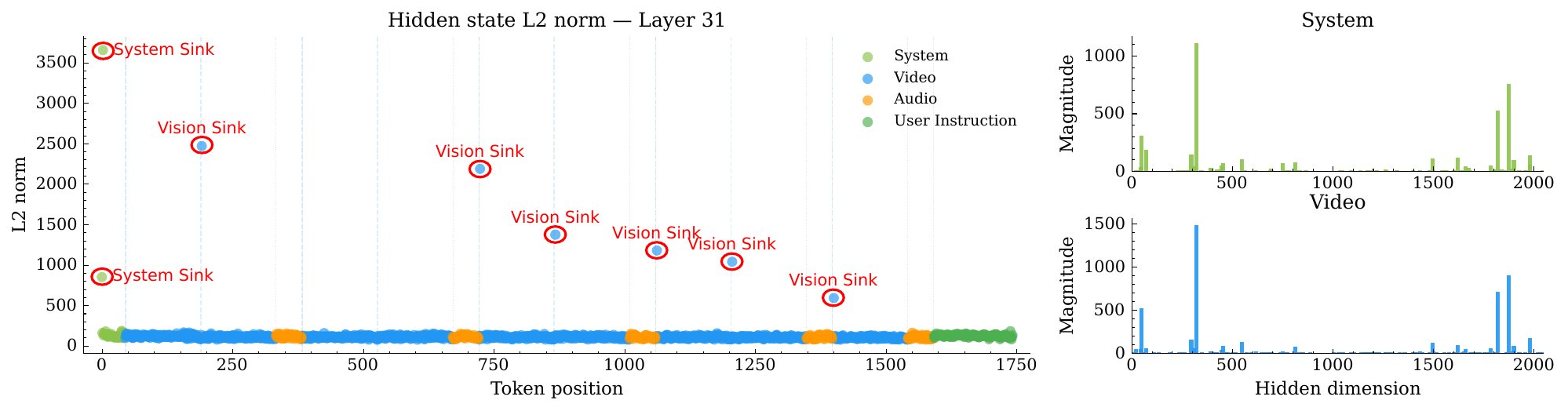}
\caption{\textbf{Vision sinks share the same hidden-state activation as the language sinks.} (Left) Hidden state $L_2$ norm at layer 31, with vision sink tokens marked by red circles. (Right) Magnitude per hidden dimension for a system sink and a vision sink, with massive activation peaks at dimensions \underline{318}, \underline{1874}, and \underline{1819} for both, on Qwen2.5-Omni 3B.}
  \label{fig:hidden}
\end{figure}

To understand why this spike emerges, we examine the attention maps at layers 30 and 31 (Figure~\ref{fig:attn} middle and right). The spike is driven by a sparse set of visual tokens generally at the first visual position of frames, receiving concentrated attention at layer 31 but absent at layer 30. This behavior of visual tokens matches the visual attention sinks identified in~\cite{kang2025see, luo2025sink}, and Figure~\ref{fig:hidden} confirms the sink-token characteristic, with these tokens generally exhibiting much larger $L_2$ norms than the rest of the sequence and activating the same hidden dimensions as the language sink tokens~\cite{xiao2023efficient, sun2024massive, gu2024attention} in the system prompt. Therefore, the attention to these visual tokens is a mechanical artifact of their massive activation, not a sign of meaningful visual information. This motivates the central question: if video attention in the later layers is dominated by sinks, does any audio-visual information actually flow to the prediction through these layers? To answer this, we apply three masking conditions at layers 31 through the final layer and measure their impact on AV-SpeakerBench~\cite{nguyen2025see}, an audio-visual video MCQ benchmark (Table~\ref{tab:sink-mask}). Across all three conditions, accuracy is unchanged or marginally improved. Despite the strong attention weights on the vision sinks, neither audio nor visual information flows through these later layers. This aligns with earlier evidence that attention weights do not necessarily reflect the importance of input tokens~\cite{jain2019attention, bai2021attentions}, and our results identify vision sinks as a concrete source of this mismatch in AVLLMs.

\begin{findingbox}
\textbf{Finding 1:} Attention allocation is not a reliable indicator of information flow in AVLLMs. Video attention in later layers is dominated by attention sinks, and audio-visual information does not flow through these deep layers.
\end{findingbox}

\section{How do audio and visual information flow in audio-visual videos?}
\label{sec:av-analysis}
In the previous section (Section~\ref{sec:sinks}), we showed that attention allocation does not reliably reveal information flow. To trace the information flow, we use causal interventions to investigate this, starting with the single audio-visual video input. Section~\ref{sec:q1} examines within- and cross-modal interactions, and Section~\ref{sec:q2} traces the route taken to the prediction. We first describe the experimental setup in Section~\ref{sec:setup}. We then extend the analysis to the multi-input interleaved configuration in Section~\ref{sec:multi-input}.

\subsection{Experimental setup}\label{sec:setup}
We use \emph{Attention Knockout}~\cite{geva2023dissecting}, a causal intervention that selectively blocks specific attention edges and measures the relative change in prediction probability. We apply it on AV-SpeakerBench~\citep{nguyen2025see}, an audio-visual four-way MCQ benchmark. To ensure each knockout measures a degradation rather than a coincidental change, we run all interventions only on samples the model predicts correctly.

\paragraph{Attention knockout:} \label{attention_knockout}
Given a source set of token positions $\mathcal{S}$ (the key side) and a target set $\mathcal{T}$ (the query side), we modify the causal mask $\mathbf{M}$ from Section~\ref{sec:preliminary} at a chosen subset of layers $\mathcal{L}$ such that $\mathbf{M}^{\ell}_{i,j} = -\infty$ for all $i \in \mathcal{T}$, $j \in \mathcal{S}$, $\ell \in \mathcal{L}$, blocking query positions in $\mathcal{T}$ from attending to key positions in $\mathcal{S}$ while leaving all other attention edges intact. We measure the effect via the relative change in the model's probability of the predicted answer letter, $\Delta p = (p_{\text{knockout}} - p_{\text{base}}) / p_{\text{base}}$, where $p_{\text{base}}$ is the probability under the original causal mask and $p_{\text{knockout}}$ is the probability after the mask modification. A large negative $\Delta p$ indicates the blocked pathway carries information critical to the prediction, while $\Delta p \approx 0$ indicates it is not necessary. These effects therefore identify the pathways that are necessary for the model to form its prediction. Following~\cite{geva2023dissecting,zhang2025cross, kim2025map}, we localize where in the network a pathway operates by applying the knockout within a sliding window of $k$ consecutive layers centered at each layer $\ell$ and sweeping $\ell$ across the full depth, yielding a curve of $\Delta p$ versus center layer. We use $k=7$ unless otherwise noted. We denote a knockout pathway as $\mathcal{S} \not\to \mathcal{T}$; for example, Video$\not\to$Question denotes blocking the question tokens from attending to the video tokens. We use $\mathcal{S} \not\leftrightarrow \mathcal{T}$ for bidirectional knockouts, where each set serves as both source and target of the other in the interleaved audio-video layout shown in Equation~\ref{eq:sync-layout}.

\paragraph{Dataset and tasks:}
\begin{table}[t]
\centering
\caption{\textbf{Five representative task categories in AV-SpeakerBench}, with example questions and all four options per example (separated by `;'). Category color indicates the cross-modal direction, \colorbox{visanchor}{visual anchor $\to$ audio answer}, \colorbox{audanchor}{audio anchor $\to$ visual answer}, and \colorbox{mixanchor}{mixed}.}
\label{tab:task_categories}
\footnotesize
\setlength{\tabcolsep}{4pt}
\renewcommand{\arraystretch}{1.1}
\begin{tabular}{@{}p{0.18\linewidth}p{0.78\linewidth}@{}}
\toprule
\textbf{Task} & \textbf{Example question} \\
\midrule
\multirow{2}{*}{\colorbox{visanchor}{Speech Recognition}} & What does the man in the blue shirt say just before he puts on a red jacket? \emph{Options:} A.~``No man''; B.~``Good at games''; C.~``Are you sure?''; D.~``No way'' \\
\midrule
\multirow{2}{*}{\colorbox{visanchor}{Speech Attributes}} & Among the people who speak, who speaks the most quietly overall? \emph{Options:} A.~The man with grey hair; B.~The man with a mustache; C.~The man with brown hair; D.~The man with glasses \\
\midrule
\multirow{2}{*}{\colorbox{audanchor}{Visual Recognition}} & When does the man say ``I have to tell you''? \emph{Options:} A.~Just before he sits down; B.~Just after he sits down; C.~Just before he takes off his glasses; D.~Just after he takes off his glasses \\
\midrule
\multirow{2}{*}{\colorbox{mixanchor}{Speaker Recognition}} & Who speaks right before the notebook is opened? \emph{Options:} A.~The man in black jacket says ``it's a tiger''; B.~The woman with blonde hair says ``I can't''; C.~The woman in blue jacket says ``something to do with how I die''; D.~The woman with blonde hair says ``I tried to draw something different'' \\
\midrule
\multirow{2}{*}{\colorbox{mixanchor}{Speaker Detection}} & Does the man in the black shirt speak after the woman hands him the ring? \emph{Options:} A.~No, he only cries; B.~No, he only laughs; C.~Yes, he tries to defend himself; D.~Yes, he tries to blame the woman \\
\bottomrule
\end{tabular}
\end{table}

AV-SpeakerBench is a speaker-centric audio-visual benchmark covering audio perception and visual understanding capabilities. Each task follows an \emph{anchor--target} design~\citep{nguyen2025see} that requires cross-modal understanding, where the \emph{anchor} is a cue in the question text pointing to an event in one modality (e.g., a visual action or a spoken phrase), and the \emph{answer} must be read off the opposite modality at the moment the anchor identifies. We select a subset of the benchmark's tasks and group them into five representative categories (Table~\ref{tab:task_categories}). Full details of the task selection and grouping are in Appendix~\ref{appendix:avspeakerbench}.

\paragraph{Models:}
We use Qwen2.5-Omni~\cite{xu2025qwen2} at 3B scale as the main subject model throughout our analysis. Results for Qwen2.5-Omni 7B and Video-SALMONN2 Plus~\cite{tang2025video} at 3B and 7B scales are reported in Appendices~\ref{appendix:qwen7b},~\ref{appendix:salmonn3b}, and~\ref{appendix:salmonn7b}. Full experimental details, including input settings and inference setup, are provided in Appendix ~\ref{appendix:experimental-details}.

\subsection{Do the modalities interact within themselves or with each other, and where?}
\label{sec:q1}

\begin{figure}[h]
  \centering
  \includegraphics[width=\linewidth]{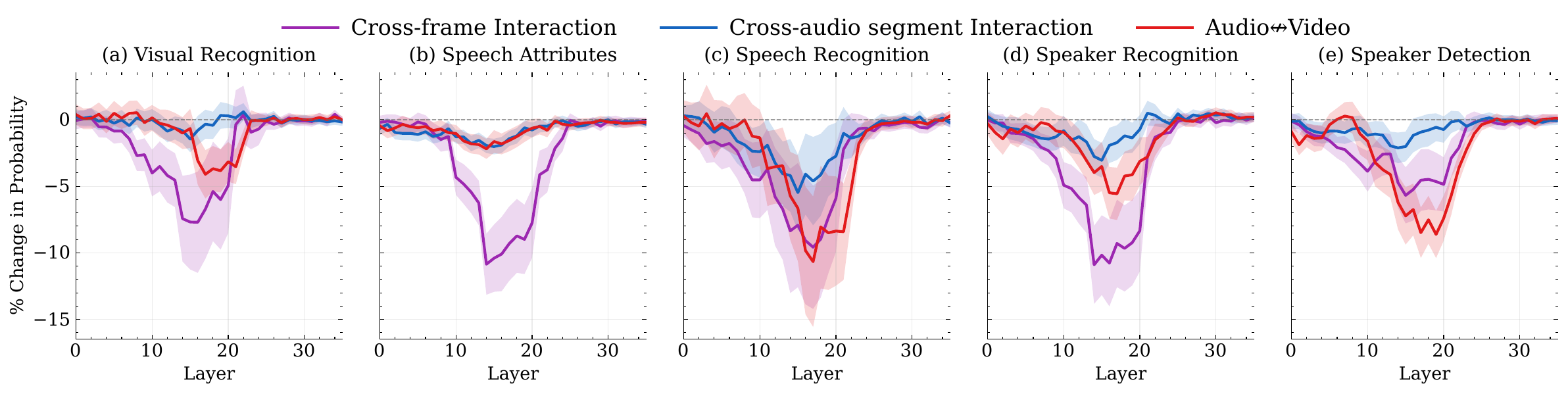}
\caption{\textbf{Within- and cross-modal interactions concentrate at early-to-middle layers.} Change in prediction probability when disconnecting within-modality (Cross-frame, Cross-audio) and direct cross-modal (Audio$\not\leftrightarrow$Video) pathways, across layers and five AV-SpeakerBench tasks. Cross-frame attention contributes across all tasks, while cross-modal effects vary by task.}
  \label{fig:knockout-sweep-1}
\end{figure}

First, we investigate whether and where the two modalities interact within themselves and with each other. Following~\cite{kim2025map}, we block cross-frame attention within the video stream, and we additionally block cross-chunk attention within the audio stream and bidirection cross-modal interaction between both modalities (Audio$\not\leftrightarrow$Video). Figure~\ref{fig:knockout-sweep-1} shows several patterns. First, all interactions are concentrated in the early-to-middle layers, with cross-modal exchange peaking shortly alongside within-modality interaction. Second, cross-frame attention contributes substantially to almost every task. This is expected for visually-grounded tasks like Visual Recognition and Speaker Recognition, but it also holds for tasks with audio-related answers. For example, in Speech Attributes, the question asks ``who speaks the most quietly?'' with the answer being a person's identity (e.g., ``the man with glasses''). Although the answer concerns audio (relative loudness), the model must first identify \emph{which speaker} corresponds to each visual descriptor, which is a visual task, before it can answer. Third, bidirectional cross-modal interaction varies by task. It carries substantial information for tasks that require fine-grained audio--visual alignment (Speech Recognition, Speaker Detection), where it operates shortly alongside cross-frame attention, but contributes little for tasks that can be solved through visual information alone. Fourth, cross-audio interaction has minimal impact across all tasks, likely because audio tokens within each chunk have already temporally interacted in the audio encoder before reaching the LLM, while video tokens lack such pre-interaction and rely on cross-frame attention within the LLM for temporal context.

\subsection{How and where does audio-visual information travel to the prediction?}
\label{sec:q2}

\begin{figure}[h]
  \centering
  \includegraphics[width=\linewidth]{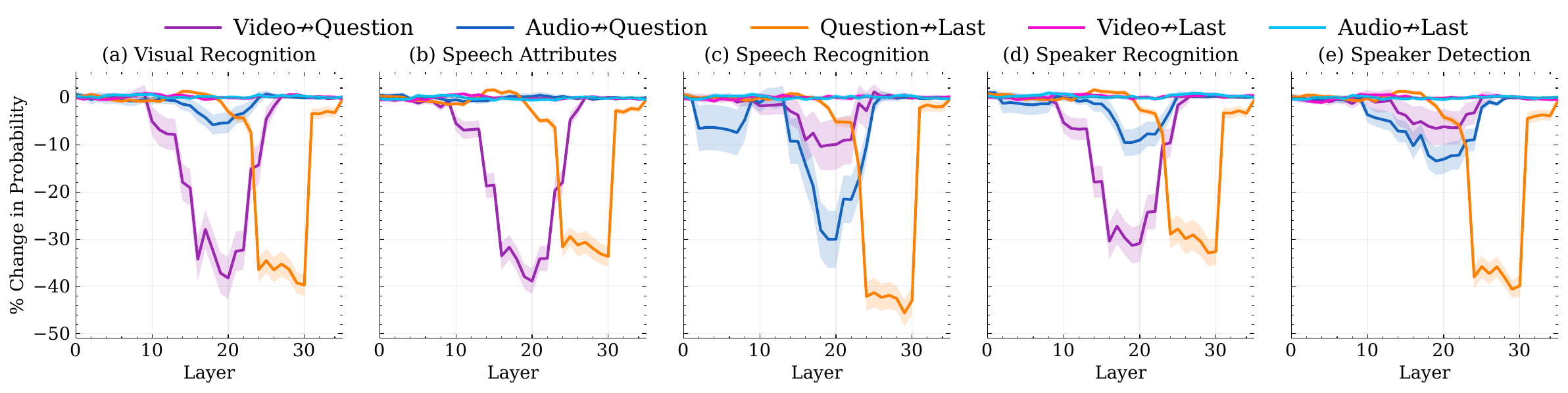}
\caption{\textbf{Overall audio-visual information flow in AVLLMs.} Change in prediction probability across knockouts targeting the question and the last token. Source$\not\to$Target indicates blocking attention edges from source tokens to target tokens. The flow follows a single sequential pathway with no direct flow from the modalities to the last token.}
  \label{fig:knockout-sweep-2}
\end{figure}

Next, we trace how audio and visual information reach the prediction. Figure~\ref{fig:knockout-sweep-2} reveals a clean indirect route, Modalities $\to$ Question $\to$ Last, aligning with prior findings on VLMs~\cite{zhang2025cross} and VideoLLMs~\cite{kim2025map}. At mid layers, video and audio transfer their information into the question tokens, picking up where the modality interactions of Section~\ref{sec:q1} left off. Since the question is positioned after the modalities in the sequence, it acts as the aggregation point for audio-visual content. The contribution of each modality along this route is shaped by the task requirements. Tasks that can be solved through visual contexts (Visual Recognition, Speech Attributes, Speaker Recognition) flow primarily through video, while tasks requiring fine-grained audio information (Speech Recognition, Speaker Detection) draw from both modalities simultaneously. At late layers, the question then carries this combined content to the last token, where the model prediction is formed. A deeper analysis of which question components (correct option, incorrect options, non-option question) carry the audio-visual content is provided in Appendix~\ref{appendix:more_knockout}.

\begin{findingbox}
\textbf{Finding 2:} Audio-visual information follows a single sequential pathway. Within- and cross-modal interactions at early-to-mid layers transfer audio-visual content into the question as the aggregation point, with their relative contribution shaped by the task requirements. Subsequently, the question carries this information to the model's prediction.
\end{findingbox}

\section{How does information flow across multiple interleaved inputs?}
\label{sec:multi-input}
In Section~\ref{sec:av-analysis}, we traced information flow in the single audio-visual video input. However, real-world prompts often arrive as multiple independent images and audio clips interleaved with text instructions. We investigate this multi-input interleaved setting, where these independent audio-visual items are interleaved with the question text. Section~\ref{sec:multi-flow} traces how information from these independent sources flows to the prediction, and Section~\ref{sec:multi-decision} reveals the information flow to the model's decision. We first describe the experimental setup in Section~\ref{sec:multi-setup}.

\subsection{Experimental setup for multiple interleaved audio-visual inputs}
\label{sec:multi-setup}

\paragraph{Dataset and tasks:}
\begin{table}
\centering
\caption{\textbf{Example multi-input interleaved prompts in AV-Odyssey.} The model matches a single \colorbox{refbg}{reference} in one modality against four \colorbox{candbg}{candidates} in the opposite modality, with the \colorbox{qbg}{question} text describing the task.}
\label{tab:avodyssey_structures}
\footnotesize
\setlength{\tabcolsep}{4pt}
\renewcommand{\arraystretch}{1.2}
\begin{tabular}{@{}p{0.08\linewidth}p{0.88\linewidth}@{}}
\toprule
\textbf{Direction} & \textbf{Example prompt} \\
\midrule
A Ref $\to$ I Cand & Which instrument illustrated in images in \colorbox{candbg}{[img1][img2][img3][img4]} \colorbox{qbg}{do you think best matches audio} \colorbox{refbg}{[audio1]}? \emph{Options:} second image; fourth image; third image; first image \\
\midrule
I Ref $\to$ A Cand & Which audio among \colorbox{candbg}{[audio1][audio2][audio3][audio4]} \colorbox{qbg}{best matches the scene shown in} \colorbox{refbg}{[img1]}? \emph{Options:} first audio; second audio; fourth audio; third audio \\
\bottomrule
\end{tabular}
\end{table}

AV-Odyssey~\cite{gong2024av} is a multi-input audio-visual benchmark where the inputs contain multiple independent images and audio clips interleaved with the question text. We focus on the matching subset where the model matches a single \emph{reference} (Ref) item in one modality against four \emph{candidate} (Cand) items in the opposite modality, across two task directions, audio reference to image candidates (A Ref $\to$ I Cand) and image reference to audio candidates (I Ref $\to$ A Cand) as shown in Table~\ref{tab:avodyssey_structures}. Within this subset, samples vary in the order of candidates, reference, and question text. We use the most common ordering in the dataset, \emph{candidates, question, reference}, and report results averaged across the selected tasks. The knockout results for individual tasks, task selection details, and additional dataset information are in Appendices~\ref{appendix:qwen3b-avodyssey} and~\ref{appendix:multi-orderings}.

\paragraph{Input structure:}
Multi-input interleaved samples consist of several independent items (images and audio clips) interleaved with the question text without temporal alignment. We extend the notation from Section~\ref{sec:preliminary} as follows: $\mathbf{C}_c$ denotes the tokens of the $c$-th candidate, $\mathbf{Q}$ the tokens of the question text, $\mathbf{R}$ the tokens of the reference, and $\mathbf{O}_o$ the tokens of the $o$-th option letter. Unlike Equation~\ref{eq:sync-layout} where the question text and option letters form a single block, the interleaved structure here separates them, with the reference appearing between the question text and the option letters. The full input sequence with \colorbox{candbg}{candidates}, \colorbox{qbg}{question}, and \colorbox{refbg}{reference} segments, is

\begin{equation}
\label{eq:multi-layout}
\mathcal{I} \;=\; \Big[\; \underbrace{s_1, \ldots, s_{N_S}}_{\text{system}} \;;\; \underbrace{\colorbox{candbg}{$\mathbf{C}_1, \mathbf{C}_2, \mathbf{C}_3, \mathbf{C}_4$}}_{\text{candidates}} \;;\; \underbrace{\colorbox{qbg}{$\mathbf{Q}$}}_{\text{question}} \;;\; \underbrace{\colorbox{refbg}{$\mathbf{R}$}}_{\text{reference}} \;;\; \underbrace{\mathbf{O}_1, \mathbf{O}_2, \mathbf{O}_3, \mathbf{O}_4}_{\text{options}} \;\Big].
\end{equation}

\subsection{How does the model route information across multiple interleaved inputs?}
\label{sec:multi-flow}

\begin{figure}[h]
  \centering
  \includegraphics[width=\linewidth]{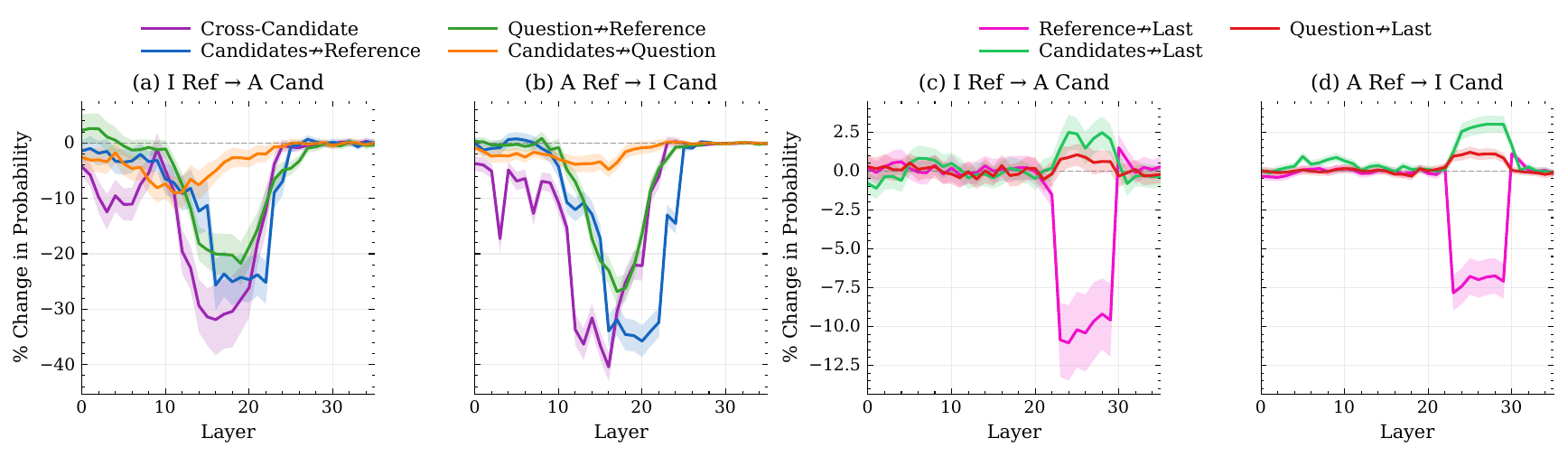}
\caption{\textbf{Multi-input interleaved information aggregates at the late-positioned token.} At mid layers, candidates interact among themselves (Cross-Candidate), and both candidates and question transfer their content to the reference. At late layers, only the reference reaches the last token.}
  \label{fig:multi-sweep-12}
\end{figure}
To trace this flow, we apply attention knockout (Section~\ref{attention_knockout}) across candidates, question, and reference. Figure~\ref{fig:multi-sweep-12} reveals the route Candidates + Question $\to$ Reference $\to$ Last. At mid layers, candidates exchange information among themselves (Cross-Candidate), then both the candidates and the question flow into the reference. The candidates and question reach the reference independently, with no flow between them. At late layers, only the reference flows to the model's prediction; blocking the candidates or question from the last token has negligible effect. The reference therefore plays the role that the question played in audio-visual videos (Section~\ref{sec:q2}), with the late-positioned token serving as the aggregation point in both settings. Since the prediction is one of four answer-choice letters, how does the reference's content translate into a specific option? We answer this in Section~\ref{sec:multi-decision}.

\subsection{How is the final answer selected?}
\label{sec:multi-decision}

\begin{figure}[h]
  \centering
  \includegraphics[width=\linewidth]{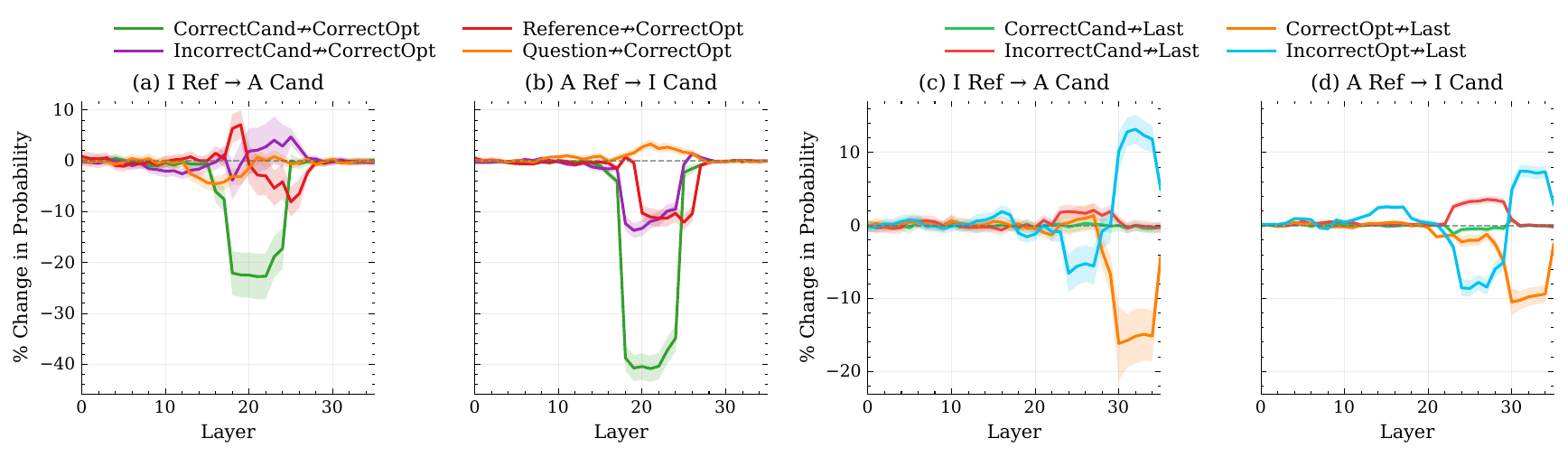}
\caption{\textbf{The prediction flows through the option letters.} (a-b) At mid layers, the correct option letter (CorrectOpt) aggregates from the correct and incorrect candidates (CorrectCand, IncorrectCand) and the reference. (c-d) At late layers, the last token reads from both correct and incorrect option letters, with the competition between them driving the prediction.}
  \label{fig:multi-sweep-34}
\end{figure}
Next, we trace how the model selects the correct option. We knock out pathways into the correct option (CorrectOpt) and from the option letters into the last token. At mid layers (Figure~\ref{fig:multi-sweep-34}a-b), CorrectOpt draws primarily from the correct candidate, with smaller contributions from the incorrect candidates and reference, and no flow from the question. At late layers (Figure~\ref{fig:multi-sweep-34}c-d), both correct and incorrect (IncorrectOpt) option letters flow to the last token. Blocking CorrectOpt~$\not\to$~Last suppresses the correct prediction, while blocking IncorrectOpt~$\not\to$~Last \emph{increases} it, indicating the incorrect options actively compete with the correct one. The decision therefore flows through the option letters, not directly from the candidates. Together, Sections~\ref{sec:multi-flow} and~\ref{sec:multi-decision} reveal two parallel paths to the prediction: (1) Candidates + Question $\to$ Reference $\to$ Last and (2) Candidates $\to$ Option letters $\to$ Last. At mid layers, both paths aggregate candidate content, but only the reference receives question content, so the question reaches the prediction exclusively via the reference. At late layers, the last token integrates from both paths independently.

\begin{findingbox}
\textbf{Finding 3:} Unlike the single sequential path in audio-visual videos, multi-input interleaved information flows through two parallel paths to the prediction, each with its own late-positioned aggregation point, integrating independently at the last token.
\end{findingbox}

\section{Do we still need multimodal and text tokens after information transfer?}
\label{sec:droptoken}

Sections~\ref{sec:av-analysis} and~\ref{sec:multi-input} establish that audio, visual, and non-option question tokens transfer their content to the late-positioned aggregation token (with the non-option question analysis for video reported in Appendix~\ref{appendix:question-components}). Combined with the attention sink result of Section~\ref{sec:sinks}, this implies that these tokens become dispensable from the sequence once their content has been transferred and can be discarded. Previously,~\cite{zhang2025cross} demonstrated this token removal for image tokens in VLMs and~\cite{kim2025map} validated effective pathways in VideoLLMs through attention masking. We extend these findings to AVLLMs by discarding audio, visual, and non-option question tokens, with each token type discarded at the distinct layer where its information transfer completes, applying this to both the single audio-visual video and multi-input interleaved configurations. We evaluate discarding video, audio, and non-option question tokens (single video) or candidates, reference, and non-option question tokens (multi-input), individually or all together. Each setting covers the task analyzed in Sections~\ref{sec:av-analysis} and~\ref{sec:multi-input}. We further evaluate generalization on cross-task and cross-dataset settings using AV-SpeakerBench~\cite{nguyen2025see}, AV-Odyssey~\cite{gong2024av}, and WorldSense~\cite{hong2025worldsense} (details in Appendix~\ref{appendix:datasets}). Table~\ref{tab:tokendrop} shows that discarding has minimal impact on model prediction and generalizes across tasks and datasets, improving model efficiency. The complete results on the cross-dataset WorldSense benchmark are provided in Appendix ~\ref{appendix:token_discard}.

\begin{findingbox}
\textbf{Finding 4:} Multimodal and text tokens can be discarded after their information is transferred, with each token type discarded once its information transfer is complete, with minimal impact on accuracy or slight improvement.
\end{findingbox}

\begin{table}[h]
\centering
\caption{\textbf{Effect of discarding multi-modal tokens on task accuracy and inference efficiency across analyzed task, cross-task, and cross-dataset settings.} $L$ denotes the layer after which the tokens are discarded. Multi-input results are reported in both matching directions reference $\to$ candidates ( I$\to$A, A$\to$I). Numbers in parentheses show change from baseline ({\color{gain}green = improvement}, {\color{drop}red = drop}); \underline{underlined} indicates no change. Best per column in \textbf{bold}. \textbf{Sp.}: Speech; \textbf{Vis.}: Visual; \textbf{Rec.}: Recognition; \textbf{Count.}: Counting; \textbf{Vid.}: Video; \textbf{Aud.}: Audio; \textbf{Transp.}: Transportation; \textbf{Ques}: non-option Question tokens; \textbf{All}: all token types together.}
\label{tab:tokendrop}
\footnotesize
\setlength{\tabcolsep}{3pt}
\renewcommand{\arraystretch}{1.1}
\begin{tabular}{@{}lccccccc@{}}
\toprule
\multicolumn{8}{c}{\textbf{Video} \textit{(AV-SpeakerBench knockout, Discard Video \& Audio at $L=26$, Ques at $L=29$; cross-dataset from WorldSense)}} \\
\midrule
\multirow{2}{*}[-0.5ex]{\textbf{Config / Task}} & \multicolumn{2}{c}{\textit{Tasks in Knockout}} & \multicolumn{2}{c}{\textit{Cross-task}} & \multicolumn{2}{c}{\textit{Cross-dataset}} & \multirow{2}{*}[-0.5ex]{\shortstack[c]{\textbf{Avg. Prefill}\\\textbf{Latency (ms)}}} \\
\cmidrule(lr){2-3} \cmidrule(lr){4-5} \cmidrule(lr){6-7}
& \textbf{Sp. Rec.} & \textbf{Vis. Rec.} & \textbf{Vis. Count.} & \textbf{Sp. Count.} & \textbf{Vid. Emotion} & \textbf{Aud. Change} & \\
\midrule
\emph{Baseline}      & 50.25 & 46.58 & 43.9 & 26.39 &  66.67& 42.22 & 2288.65 \\
Discard Ques         & \underline{50.25} & 46.83 \tiny{({\color{gain}+0.25})} & \textbf{44.39} \tiny{({\color{gain}+0.49})} & \underline{26.39} & \underline{66.67} & 40.00 \tiny{({\color{drop}-2.22})} & 2279.97 \\
Discard Audio        & \underline{50.25} & \textbf{47.55} \tiny{({\color{gain}+0.97})} & \textbf{44.39} \tiny{({\color{gain}+0.49})} & \textbf{26.74} \tiny{({\color{gain}+0.35})} & \underline{66.67} & 40.00 \tiny{({\color{drop}-2.22})} & 2232.45 \\
Discard Video        & \textbf{50.75} \tiny{({\color{gain}+0.50})} & 46.10 \tiny{({\color{drop}-0.48})} & \underline{43.9} & \textbf{26.74} \tiny{({\color{gain}+0.35})} & \underline{66.67} & 40.00 \tiny{({\color{drop}-2.22})} & 2098.75 \\
Discard All          & 49.75 \tiny{({\color{drop}-0.50})} & 46.59 \tiny{({\color{gain}+0.01})} & 42.93 \tiny{({\color{drop}-0.97})} & 26.04 \tiny{({\color{drop}-0.35})} & \underline{66.67} & \textbf{\underline{42.22}} & 2089.47 \\
\midrule
\multicolumn{8}{c}{\textbf{Multi-input} \textit{(AV-Odyssey knockout, Discard Cand at $L=25$, Discard Ref at $L=31$, Discard Ques at $L=29$)}} \\
\midrule
\multirow{3}{*}[-1.5ex]{\textbf{Config / Task}} & \multicolumn{2}{c}{\textit{Tasks in Knockout}} & \multicolumn{4}{c}{\textit{Cross-task}} & \multirow{3}{*}[-1.5ex]{\shortstack[c]{\textbf{Avg. Prefill}\\\textbf{Latency (ms)}}} \\
\cmidrule(lr){2-3} \cmidrule(lr){4-7}
& \multicolumn{2}{c}{\textbf{Animal Rec.}} & \multicolumn{2}{c}{\textbf{Bird Rec.}} & \multicolumn{2}{c}{\textbf{Transp. Rec.}} & \\
\cmidrule(lr){2-3} \cmidrule(lr){4-5} \cmidrule(lr){6-7}
& \textbf{A$\to$I} & \textbf{I$\to$A} & \textbf{A$\to$I} & \textbf{I$\to$A} & \textbf{A$\to$I} & \textbf{I$\to$A} & \\
\midrule
\emph{Baseline}      & 61.00 & 38.00 & 29.41 & 33.67 & 46.67 & 23.16 & 558.75 \\
Discard Ques         & 62.00 \tiny{({\color{gain}+1.00})} & \underline{38.00} & 30.39 \tiny{({\color{gain}+0.98})} & \textbf{34.69} \tiny{({\color{gain}+1.02})} & 44.76 \tiny{({\color{drop}-1.91})} & 24.21 \tiny{({\color{gain}+1.05})} & 550.41 \\
Discard Ref          & \textbf{63.00} \tiny{({\color{gain}+2.00})} & \textbf{40.00} \tiny{({\color{gain}+2.00})} & \underline{29.41} & \underline{33.67} & \underline{46.67} & \textbf{25.26} \tiny{({\color{gain}+2.10})} & 552.11 \\
Discard Cand         & \textbf{63.00} \tiny{({\color{gain}+2.00})} & 39.00 \tiny{({\color{gain}+1.00})} & \textbf{32.35} \tiny{({\color{gain}+2.94})} & 31.63 \tiny{({\color{drop}-2.04})} & \textbf{47.62} \tiny{({\color{gain}+0.95})} & \textbf{25.26} \tiny{({\color{gain}+2.10})} & 533.07 \\
Discard All          & \textbf{63.00} \tiny{({\color{gain}+2.00})} & \underline{38.00} & \textbf{32.35} \tiny{({\color{gain}+2.94})} & \underline{33.67} & \underline{46.67} & 24.21 \tiny{({\color{gain}+1.05})} & 530.62 \\
\bottomrule
\end{tabular}
\end{table}

\section{Discussion, future works and limitations}
\label{sec:discussion}
\textbf{Discussion:}
We present the first comprehensive analysis of information flow in AVLLMs across single audio-visual video and multi-input interleaved configurations. In both configurations, modality content reaches the prediction by being aggregated into a token positioned later in the sequence. A plausible explanation for this aggregator emergence is causal attention, which makes tokens positioned later in the sequence structurally available to absorb upstream content, paired with the finding from~\cite{kim2025map} that modality tokens progressively align with linguistic embeddings at mid layers, which could plausibly provide the semantic basis for cross-modal information transfer. Concurrent work~\cite{selvakumar2026audio} reports that cross-modal integration concentrates in deep layers in captioning. Our analysis offers an alternative view in the question-answering setting, where attention to video tokens at the late layers is dominated by sinks (Section~\ref{sec:sinks}) and discarding different token segments at the distinct layers where their transfer completes has minimal impact on performance (Section~\ref{sec:droptoken}). The actual integration takes place much earlier, at mid layers where modality tokens transfer to the aggregator. What appears as deep-layer integration in captioning may therefore reflect attention to sinks rather than meaningful integration, though captioning and question-answering may engage distinct routing at late layers.

\textbf{Additional analyses:}
Furthermore, we show in Appendix~\ref{appendix:sink} that vision sinks emerge later in the network than the language sink. In addition, we show in Appendix~\ref{appendix:fps-spatial-ablation} that our findings remain consistent across different frame rates and spatial resolutions, and in Appendix~\ref{appendix:qwen3b-videomme} that they hold on VideoMME tasks that rely less on audio information. Finally, we show in Appendices~\ref{appendix:qwen3b-worldsense},~\ref{appendix:qwen7b-worldsense},~\ref{appendix:salmonn3b-worldsense}, and~\ref{appendix:salmonn7b-worldsense} that the same information flow holds on WorldSense across all models.

\textbf{Future work:}
Our findings open several research directions. First, our finding that tokens can be discarded after their information is transferred (Section~\ref{sec:droptoken}) opens a new direction for AVLLM efficiency through token compression at the LLM's internal layers, complementing existing input-level methods that compress audio-visual tokens before they enter the LLM~\cite{gong2025echoingpixels, ding2026omnisift, tao2025omnizip, li2026dash}. Second, the task-dependent modality contribution we observe (Section~\ref{sec:av-analysis}) raises the question of whether steering AVLLMs to rebalance modality reliance could improve performance for tasks where one modality is underused. Third, visual bias in AVLLMs has been reported from ~\cite{selvakumar2026audio} through counterfactual analysis where audio and video are intentionally mismatched. Extending our information flow analysis to these conditions is a natural next step for understanding where visual bias emerges along the routing pathway. Fourth, complementing attention knockout with other mechanistic techniques such as causal patching~\cite{meng2022locating} could provide further evidence on how these pathways operate under the model's original inference conditions.

\textbf{Limitations:}
Our analysis operates in the MCQ setting, where the prediction is a single answer letter. Open-ended generation tasks such as captioning or free-form dialogue may engage distinct pathways that our setup does not capture.

\section*{Acknowledgments}
We acknowledge EuroHPC Joint Undertaking for awarding us access to MareNostrum5 as BSC, Spain. Use as many instances of the pattern MareNostrum5 at BSC, Spain as the number
of systems awarded via EuroHPC. We also acknowledge the support of the Surrey Institute for
People-Centred Artificial Intelligence (PAI) at the University of Surrey through a doctoral studentship.

{
\small
\bibliographystyle{abbrv}
\bibliography{references_neurips}

}


\newpage
\appendix
\label{appendix}

\newcommand{\appdots}{\leaders\hbox to 0.7em{\hss.\hss}\hfill}

\newcommand{\appsec}[2]{%
  \par\addvspace{0.6em}%
  \noindent\hyperref[#1]{\textbf{\makebox[2.2em][l]{\textcolor{red}{\ref*{#1}}}\textcolor{black}{#2}}}%
  \nobreak\appdots\nobreak\hyperref[#1]{\textbf{\textcolor{black}{\pageref*{#1}}}}\par}

\newcommand{\appsubsec}[2]{%
  \par\addvspace{0.15em}%
  \noindent\hspace{2.2em}\hyperref[#1]{\makebox[2.8em][l]{\textcolor{red}{\ref*{#1}}}\textcolor{black}{#2}}%
  \nobreak\appdots\nobreak\hyperref[#1]{\textcolor{black}{\pageref*{#1}}}\par}

\noindent{\large\bfseries Appendix Contents}
\vspace{0.4em}

\appsec{appendix:relatedwork}{Related Works}
  \appsubsec{appendix:relatedwork-avllm}{Audio-Visual Large Language Models (AVLLMs)}
  \appsubsec{appendix:relatedwork-mechinterp}{Mechanistic Interpretability of LLMs and MLLMs}

\appsec{appendix:sink}{Additional Visualizations of Attention Sink Tokens}

\appsec{appendix:experimental-details}{Experimental Details}

\appsec{appendix:datasets}{Dataset Details}
  \appsubsec{appendix:avspeakerbench}{AV-SpeakerBench}
  \appsubsec{appendix:worldsense}{WorldSense}
  \appsubsec{appendix:av_odyssey}{AV-Odyssey}
  \appsubsec{appendix:videomme-data}{VideoMME}

\appsec{appendix:window-ablation}{Ablation on Window Size for Attention Knockout}

\appsec{appendix:fps-spatial-ablation}{Ablation on FPS and Spatial Resolution}

\appsec{appendix:token_discard}{Cross-Dataset Generalization of Token Discard on WorldSense}

\appsec{appendix:qwen3b-additional}{Additional Knockout Analyses on Qwen2.5-Omni 3B}
  \appsubsec{appendix:more_knockout}{Extended Knockout Analyses on AV-SpeakerBench}
  \appsubsec{appendix:qwen3b-videomme}{VideoMME}
  \appsubsec{appendix:qwen3b-worldsense}{WorldSense}
  \appsubsec{appendix:qwen3b-avodyssey}{AV-Odyssey}

\appsec{appendix:qwen7b}{Knockout Analyses: Generalization to Qwen2.5-Omni 7B}
  \appsubsec{appendix:qwen7b-avspeakerbench}{AV-SpeakerBench}
  \appsubsec{appendix:qwen7b-worldsense}{WorldSense}
  \appsubsec{appendix:qwen7b-avodyssey}{AV-Odyssey}

\appsec{appendix:salmonn3b}{Knockout Analyses: Generalization to Video-SALMONN2 3B Plus}
  \appsubsec{appendix:salmonn3b-avspeakerbench}{AV-SpeakerBench}
  \appsubsec{appendix:salmonn3b-worldsense}{WorldSense}

\appsec{appendix:salmonn7b}{Knockout Analyses: Generalization to Video-SALMONN2 7B Plus}
  \appsubsec{appendix:salmonn7b-avspeakerbench}{AV-SpeakerBench}
  \appsubsec{appendix:salmonn7b-worldsense}{WorldSense}

\clearpage

\section{Related works}\label{appendix:relatedwork}

\subsection{Audio-visual large language models (AVLLMs)}
\label{appendix:relatedwork-avllm}

Audio-visual large language models (AVLLMs) extend the multimodal LLM paradigm to jointly process audio and visual inputs. The first generation of AVLLMs~\cite{cheng2024videollama, tang2025video, ye2024cat, liu2025ola} couples dedicated audio and visual encoders with a language model to support text-based audio-visual question answering and dialogue. Building on this foundation, omni models~\cite{xu2025qwen2, xu2025qwen3, team2026qwen3, ye2025omnivinci, li2025uni, fu2025vita} introduce temporal alignment between audio and visual streams and unify perception with end-to-end speech generation, pushing AVLLMs toward real-time multimodal interaction. Alongside these architectural advances, a growing body of work targets the inference efficiency of AVLLMs by compressing audio and visual tokens prior to the language model~\cite{gong2025echoingpixels, ding2026omnisift, tao2025omnizip, li2026dash}. Distinct from these prior works, we perform a mechanistic interpretability study of AVLLMs in this study, tracing how audio and visual information flow through the model to form the prediction.

\subsection{Mechanistic interpretability of LLMs and MLLMs}
\label{appendix:relatedwork-mechinterp}

Mechanistic interpretability~\cite{sharkey2025open, rai2024practical, nanda2023progress, elhage2021mathematical, geva2023dissecting} is the study of how internal computations in neural networks give rise to their behavior. It has become an  become a popular research direction that recently extends from LLMs to MLLMs. In LLMs, prior studies have uncovered the algorithms underlying grokking~\cite{nanda2023progress} and the attention head circuits behind transformer computation~\cite{elhage2021mathematical}. Among the methodological tools developed in this field, attention knockout~\cite{geva2023dissecting} causally intervenes on attention pathways to identify the routes information takes through the model. Within MLLMs, recent work has examined information storage~\cite{basu2024understanding}, modality-specific circuits~\cite{nikankin2025same},visual information processing~\cite{neo2024towards, kaduri2025s} and visual representations across different integration paradigms~\cite{suharitdamrong2026hidden}. Most relevant to our study, \cite{zhang2025cross} and \cite{kim2025map} apply attention knockout to trace information flow in image VLMs and VideoLLMs, respectively. Concurrent work~\cite{selvakumar2026audio} investigates AVLLMs through counterfactual analysis on audio-visual captioning and ~\cite{jung2026probing}  uses causal tracing to locate where cross-modal information is stored within token representations.  Distinct from these prior works, we trace the internal mechanism of how audio and visual information flow inside AVLLMs in both audio-visual video and multi-input interleaved scenarios to form the prediction in this study.

\section{Additional visualizations of attention sink tokens}\label{appendix:sink}
\begin{figure}[htbp]
  \centering
  \includegraphics[width=\linewidth]{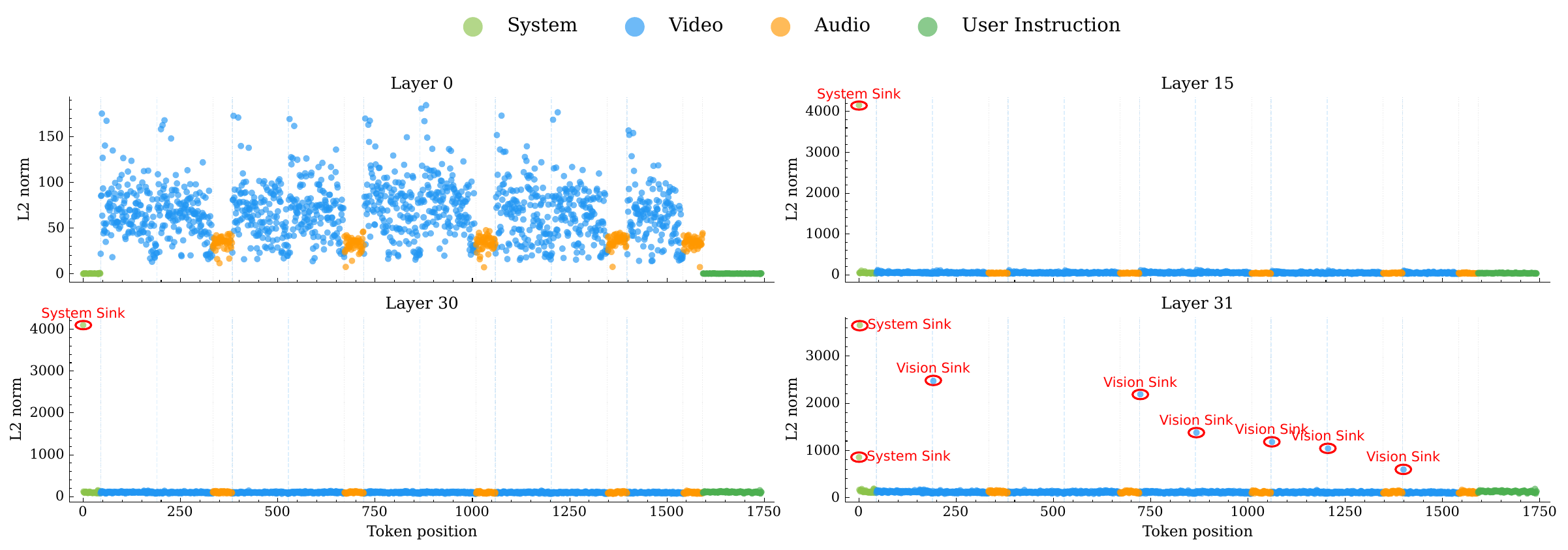}
\caption{L2 norm distribution across token positions at four representative layers (0, 15, 30, 31). Tokens are colored by type (system, video, audio, user instruction). High-norm sink tokens are highlighted with red circles. The language sink emerges at layer 15 in the system prompt region, while vision sinks emerge later at layer 31 in the video sequence.}
  \label{fig:appendix_sink}
\end{figure}
To trace when each type of sink emerges across layers, we analyze the L2 norm distribution at four representative layers (0, 15, 30, 31), shown in Figure~\ref{fig:appendix_sink}. At layer 0, vision and audio tokens already have higher norms than language tokens, since they have already passed through the modality encoders before entering the LLM. By layer 15, a single high-norm token has emerged in the system prompt region, corresponding to the language sink described in~\cite{xiao2023efficient}. This language sink persists through layer 30, while modality tokens remain at low norms with no vision sink yet present. At layer 31, vision sinks emerge sharply at specific positions in the video sequence, marking the layer at which the visual anchors identified in Section~\ref{sec:sinks} first appear. This progression confirms that vision sinks emerge later than the language sink, despite sharing the same mechanical signature.

\section{Experimental details}
\label{appendix:experimental-details}

\paragraph{Models and inputs:} We analyze Qwen2.5-Omni 3B ~\cite{xu2025qwen2} as the primary model, with additional results on its 7B variants and Video-SALMONN2 Plus 3B and 7B~\cite{tang2025video} in Appendix~\ref{appendix:qwen7b}, ~\ref{appendix:salmonn3b} and~\ref{appendix:salmonn7b}. All models are loaded from their official Hugging Face checkpoints. Videos are sampled at 2 FPS with up to 128 frames per video, where each frame is represented by up to a 12$\times$12 grid of visual tokens, and standalone images are represented by up to a 24$\times$24 grid of visual tokens.

\paragraph{Inference setup:} All experiments use greedy decoding (\texttt{do\_sample=False}) for deterministic outputs and are conducted on a single NVIDIA H100 GPU. To ensure each knockout measures a real degradation rather than a coincidental change, we run all interventions only on samples that the model predicts correctly under the no-knockout baseline. A typical knockout run takes around 30 minutes for the audio-visual video setting and 15 minutes for the multi-input interleaved setting.

\section{Dataset details}
\label{appendix:datasets}

This section provides additional details on the datasets used in our analysis. AV-SpeakerBench (Section~\ref{appendix:avspeakerbench}) is the audio-visual video benchmark used in the main paper, AV-Odyssey (Section~\ref{appendix:av_odyssey}) is the multi-input audio-visual benchmark also used in the main paper, and WorldSense (Section~\ref{appendix:worldsense}) is an additional audio-visual video benchmark used in this appendix to provide further evidence on the generality of our findings.

\subsection{AV-SpeakerBench (audio-visual video)}
\label{appendix:avspeakerbench}

AV-SpeakerBench~\citep{nguyen2025see} is a benchmark for evaluating audio--visual reasoning in models that jointly process video and speech. Each sample is a four-way multiple-choice question paired with an audio-visual video, and the benchmark contains 12 tasks across audio-centric, speaker-centric, and visual-centric domains.

\paragraph{Cross-modal anchor design:} A key feature of AV-SpeakerBench is that each question is constructed around an explicit \emph{anchor--target} structure that forces genuine audio--visual integration. The anchor is a cue described in the question text that points to a specific moment in one modality, and the answer must be read off the opposite modality at that moment. We summarize the three task types as follows:

\begin{itemize}
    \item \textbf{Audio-centric tasks} (\colorbox{visanchor}{visual anchor $\to$ audio answer}): the question describes a visual cue (e.g., \emph{``after the man in the grey shirt wiggles his fingers''}), and the model must first locate that visual moment, then listen to the audio within that window to extract the answer (e.g., counting how often a phrase is spoken).
    \item \textbf{Visual-centric tasks} (\colorbox{audanchor}{audio anchor $\to$ visual answer}): the question describes an audio cue, typically a spoken phrase (e.g., \emph{``after the woman says This is very datable''}), and the model must locate that moment in the audio, then inspect the video at that timestamp to extract a visual answer (e.g., counting visible people).
    \item \textbf{Speaker-centric tasks} (\colorbox{mixanchor}{mixed}): the question may use either a visual or audio anchor, and the answer choices differ in the opposite modality (e.g., visually distinct speakers who utter different lines). Solving these requires jointly tracking identity, timing, and modality, which prevents unimodal shortcuts.
\end{itemize}

This anchor structure makes AV-SpeakerBench particularly well-suited for our analysis, since the cross-modal pathway is built directly into the question design rather than emerging incidentally from the content.

\paragraph{Task selection for attention knockout:}  For our analysis, we focus on tasks that exhibit the cross-modal anchor structure described above and exclude tasks where the question can be largely answered from a single modality without the anchor playing a meaningful role. The 8 selected tasks, totaling 2{,}281 samples, are organized into 5 categories used throughout Section~\ref{sec:av-analysis}. Table~\ref{tab:avspeakerbench-selected-tasks} (left) lists the selected tasks with their domain and category, and (right) reports the per-category sample counts. Categories with multiple tasks (Visual Recognition, Speech Attributes) merge tasks that share the same anchor direction and answer modality; the other three categories correspond to a single task each. The knockout results on AV-SpeakerBench are reported in Section~\ref{sec:av-analysis} for Qwen2.5-Omni 3B, with extended analyses in Appendix~\ref{appendix:more_knockout}, and in Appendices~\ref{appendix:qwen7b-avspeakerbench}, \ref{appendix:salmonn3b-avspeakerbench}, and~\ref{appendix:salmonn7b-avspeakerbench} for Qwen2.5-Omni 7B and Video-SALMONN2 Plus 3B and 7B, respectively.

\begin{table}[htbp]
\caption{\textbf{Left:} The 8 tasks selected from AV-SpeakerBench, with their domain, sample counts, and the category each task is grouped into for our analysis. \textbf{Right:} The 5 categories with their aggregated sample counts. Domain color indicates the cross-modal direction: \colorbox{visanchor}{visual anchor $\to$ audio answer}, \colorbox{audanchor}{audio anchor $\to$ visual answer}, and \colorbox{mixanchor}{mixed}.}
\label{tab:avspeakerbench-selected-tasks}
\footnotesize
\setlength{\tabcolsep}{4pt}
\renewcommand{\arraystretch}{1.15}
\begin{minipage}[t]{0.62\linewidth}
\vspace{0pt}
\centering
\begin{tabular}{@{}llcl@{}}
\toprule
\textbf{Domain} & \textbf{Task} & \textbf{\# Samples} & \textbf{Category} \\
\midrule
\multirow{4}{*}{\colorbox{visanchor}{Audio-centric}}
    & Speech Intensity      & 206 & Speech Attributes \\
    & Speech Pitch          & 206 & Speech Attributes \\
    & Speech Rate           & 209 & Speech Attributes \\
    & Speech Recognition    & 201 & Speech Recognition \\
\midrule
\multirow{2}{*}{\colorbox{mixanchor}{Speaker-centric}}
    & Speaker Detection     & 427 & Speaker Detection \\
    & Speaker Recognition   & 422 & Speaker Recognition \\
\midrule
\multirow{2}{*}{\colorbox{audanchor}{Visual-centric}}
    & Activity Recognition  & 206 & Visual Recognition \\
    & Attribute Recognition & 204 & Visual Recognition \\
\midrule
\textbf{Total} & & \textbf{2{,}281} & \\
\bottomrule
\end{tabular}
\end{minipage}%
\hfill
\begin{minipage}[t]{0.36\linewidth}
\vspace{0pt}
\centering
\begin{tabular}{@{}lc@{}}
\toprule
\textbf{Category} & \textbf{\# Samples} \\
\midrule
\colorbox{visanchor}{Speech Attributes}  & 621 \\
\colorbox{visanchor}{Speech Recognition} & 201 \\
\colorbox{audanchor}{Visual Recognition} & 410 \\
\colorbox{mixanchor}{Speaker Recognition} & 422 \\
\colorbox{mixanchor}{Speaker Detection}  & 427 \\
\midrule
\textbf{Total} & \textbf{2{,}281} \\
\bottomrule
\end{tabular}
\end{minipage}
\end{table}

\paragraph{Task selection for cross-task evaluation:} In addition to the analysis tasks, for Section~\ref{sec:droptoken} we use Speech Counting and Visual Counting from AV-SpeakerBench. Table~\ref{tab:avspeakerbench-crosstask} reports the domain and sample count for each.

\begin{table}[htbp]
\centering
\caption{Cross-task evaluation tasks from AV-SpeakerBench used in Section~\ref{sec:droptoken}. Domain color indicates the cross-modal direction: \colorbox{visanchor}{visual anchor $\to$ audio answer}, \colorbox{audanchor}{audio anchor $\to$ visual answer}.}
\label{tab:avspeakerbench-crosstask}
\footnotesize
\setlength{\tabcolsep}{4pt}
\renewcommand{\arraystretch}{1.15}
\begin{tabular}{@{}llc@{}}
\toprule
\textbf{Domain} & \textbf{Task} & \textbf{\# Samples} \\
\midrule
\colorbox{visanchor}{Audio-centric}  & Speech Counting & 288 \\
\colorbox{audanchor}{Visual-centric} & Visual Counting & 205 \\
\midrule
\textbf{Total} & & \textbf{493} \\
\bottomrule
\end{tabular}
\end{table}

\subsection{WorldSense (audio-visual video)}
\label{appendix:worldsense}

WorldSense~\citep{hong2025worldsense} is a benchmark of audio-visual videos paired with multiple-choice questions, designed to evaluate joint reasoning over visual, audio, and temporal information. The benchmark covers 26 task types organized into three task domains: Recognition (identifying entities or events), Reasoning (inferring causal or relational structure), and Understanding (comprehending temporal or contextual states).

\paragraph{Task selection for attention knockout:} WorldSense contains videos ranging from a few seconds to over eight minutes. To keep the audio-visual stream short enough for tractable knockout sweeps and to ensure that all selected samples fit within a comparable input length, we filter to videos under one minute long. From this filtered set, we select the 10 task types marked under \emph{Knockout} in Table~\ref{tab:worldsense-tasks}, totaling 418 samples. The selected videos have a mean duration of 47.4 seconds (median 49 seconds, range 17--60 seconds). Compared to AV-SpeakerBench, WorldSense provides substantially fewer samples per task (typically 25--64 per task type, against 200--400 in AV-SpeakerBench). As a result, the per-task knockout curves on WorldSense are visibly noisier than those on AV-SpeakerBench. The knockout results on WorldSense are reported in Appendices~\ref{appendix:qwen3b-worldsense}, \ref{appendix:qwen7b-worldsense}, \ref{appendix:salmonn3b-worldsense}, and~\ref{appendix:salmonn7b-worldsense} for Qwen2.5-Omni 3B and 7B and Video-SALMONN2 Plus 3B and 7B, respectively.

\paragraph{Task selection for cross-dataset evaluation:} For the cross-dataset evaluation of token discarding, we use all 26 task types of WorldSense under the same one-minute filter, totaling 1{,}047 samples. Section~\ref{sec:droptoken} reports Video Emotions and Audio Change, and the results on all tasks are reported in Appendix~\ref{appendix:token_discard}.

\begin{table}[htbp]
\centering
\caption{WorldSense tasks used in our analysis, restricted to videos under one minute. \emph{Knockout} marks the tasks used for attention knockout. All tasks are used for the cross-dataset evaluation of token discarding in Section~\ref{sec:droptoken} and Appendix~\ref{appendix:token_discard}.}
\label{tab:worldsense-tasks}
\footnotesize
\setlength{\tabcolsep}{6pt}
\renewcommand{\arraystretch}{1.1}
\begin{tabular}{@{}llcc@{}}
\toprule
\textbf{Domain} & \textbf{Task} & \textbf{\# Samples} & \textbf{Knockout} \\
\midrule
\multirow{11}{*}{Recognition}
    & Action Counting                & 46 &              \\
    & Anomaly Recognition            & 27 &              \\
    & Attribute Recognition          & 52 & $\checkmark$ \\
    & Audio Counting                 & 38 & $\checkmark$ \\
    & Audio Recognition              & 88 &              \\
    & Audio Source Localization      & 64 & $\checkmark$ \\
    & Event Recognition              & 30 & $\checkmark$ \\
    & Object Counting                & 47 &              \\
    & Object Existence Recognition   & 31 &              \\
    & Scene Recognition              & 27 & $\checkmark$ \\
    & Temporal Localization          & 56 &              \\
\midrule
\multirow{8}{*}{Reasoning}
    & Attribute Reasoning            & 42 &              \\
    & Audio Change                   & 45 &              \\
    & Causal Reasoning               & 37 &              \\
    & Emotion Change                 & 37 & $\checkmark$ \\
    & Hallucination                  & 21 &              \\
    & Object State Change            & 25 & $\checkmark$ \\
    & Relation Reasoning             & 20 &              \\
    & Temporal Prediction            & 36 &              \\
\midrule
\multirow{7}{*}{Understanding}
    & Event Sorting                  & 46 & $\checkmark$ \\
    & Human Emotions                 & 29 &              \\
    & Human Interaction              & 36 &              \\
    & Human-object Interaction       & 41 &              \\
    & Spatial Relation               & 58 & $\checkmark$ \\
    & Text and Diagram Understanding & 41 & $\checkmark$ \\
    & Video Emotions                 & 27 &              \\
\midrule
\textbf{Total} & & \textbf{1{,}047} & \textbf{418} \\
\bottomrule
\end{tabular}
\end{table}

\subsection{AV-Odyssey (multi-input audio-visual interleaved)}
\label{appendix:av_odyssey}

AV-Odyssey~\citep{gong2024av} is a benchmark for audio-visual understanding that contains 26 tasks spanning multiple reasoning skills, where each sample interleaves multiple independent images and audio clips with text. Among these, we focus on the matching subset where the model selects an answer by comparing a single \colorbox{refbg}{reference} item in one modality against multiple \colorbox{candbg}{candidate} items in the opposite modality, which aligns with the multi-input setting analyzed in Section~\ref{sec:multi-input}. This appendix provides additional details on how we process AV-Odyssey for our analysis: Section~\ref{appendix:multi-orderings} describes the procedure used to assign each sample an input structure label, and Section~\ref{appendix:av_odyssey_tasks} lists the specific tasks we select and reports the distribution of input structures across them.

\subsubsection{Input structure assignment for AV-Odyssey}
\label{appendix:multi-orderings}

AV-Odyssey samples vary in how the \colorbox{candbg}{candidate} media, \colorbox{refbg}{reference} media, and \colorbox{qbg}{question} text are arranged within the prompt. To support per-ordering analysis, we automatically classify each sample's input structure through the following procedure:

\begin{enumerate}
    \item \textbf{Parse media placeholders.} We scan the prompt to locate each media placeholder (e.g., \texttt{[img1]}, \texttt{[audio1]}) and identify the single \colorbox{refbg}{reference} media along with the \colorbox{candbg}{candidate} media in the opposite modality.
    \item \textbf{Split into ordered segments.} Using the located media placeholders as boundaries, we split the prompt into an ordered sequence of segments alternating between media spans (\colorbox{candbg}{candidates} and \colorbox{refbg}{reference}) and text spans.
    \item \textbf{Identify the question text.} Among the text segments, we assign the role of \emph{\colorbox{qbg}{question} text} to the segment that describes the actual matching task. If the \colorbox{refbg}{reference} media is the final media in the prompt, we prefer the text segment immediately preceding it (which typically contains the matching prompt, e.g., \emph{``which best matches''}); otherwise, we select the longest text segment by word count. All remaining text segments are treated as padding and excluded from the analysis.
    \item \textbf{Construct the structure label.} The final segment ordering, for example \colorbox{candbg}{candidates}, \colorbox{qbg}{question}, \colorbox{refbg}{reference}, is recorded as the sample's structure label.
\end{enumerate}

We use the resulting structure labels to bucket samples for both the main-paper analysis (which uses the most common ordering, \colorbox{candbg}{candidates}, \colorbox{qbg}{question}, \colorbox{refbg}{reference}) and the per-ordering breakdowns reported in this appendix.

\subsubsection{Selected tasks from AV-Odyssey}
\label{appendix:av_odyssey_tasks}

\paragraph{Task selection for attention knockout:} For our knockout analysis, we select 7 tasks from AV-Odyssey that use either a single \colorbox{refbg}{reference} media against multiple \colorbox{candbg}{candidates} of the opposite modality or vice versa. Table~\ref{tab:av-odyssey-selected-tasks} lists the 7 selected tasks, totaling 1{,}304 samples. Within each task, the prompts can also vary in how the \colorbox{candbg}{candidates}, \colorbox{qbg}{question} text, and \colorbox{refbg}{reference} are ordered. Table~\ref{tab:av-odyssey-structures} shows the aggregate distribution across all selected samples, and Table~\ref{tab:av-odyssey-per-task-structures} provides the per-task breakdown. The dominant ordering is \colorbox{candbg}{candidates},\colorbox{qbg}{question},\colorbox{refbg}{reference}, which we use for the main paper analysis; the remaining orderings are analyzed individually in this appendix. The knockout results on AV-Odyssey are reported in Section~\ref{sec:multi-input} for Qwen2.5-Omni 3B averaged across tasks, with per-task results in Appendix~\ref{appendix:qwen3b-avodyssey}, and in Appendix~\ref{appendix:qwen7b-avodyssey} for Qwen2.5-Omni 7B.

\paragraph{Task selection for cross-task evaluation:} In addition to the tasks used for attention knockout, for Section~\ref{sec:droptoken} we use Bird Recognition and Transportation Recognition from AV-Odyssey. Table~\ref{tab:av-odyssey-crosstask} reports the sample count and input layout distribution for each.

\begin{table}[htbp]
\centering
\caption{The 7 tasks selected from AV-Odyssey for attention knockout. A Ref → I Cand denotes audio reference to image candidates , and I Ref → A Cand denotes image reference to audio candidates .}
\label{tab:av-odyssey-selected-tasks}
\footnotesize
\setlength{\tabcolsep}{4pt}
\renewcommand{\arraystretch}{1.1}
\begin{tabular}{@{}lccc@{}}
\toprule
\textbf{Task} & \textbf{\# Samples} & \textbf{A Ref $\to$  I Cand} & \textbf{I Ref $\to$  A Cand} \\
\midrule
Instrument Recognition       & 200  & 49\% & 51\% \\
Animal Recognition           & 200  & 50\% & 50\% \\
Material Recognition         & 200  & 100\% & --- \\
Hazard Recognition           & 108  & 100\% & --- \\
Action Recognition           & 196  & 100\% & --- \\
Music Genre Classification   & 200  & 52\% & 48\% \\
Film and Music Matching      & 200  & 100\% & --- \\
\midrule
\textbf{Total} & \textbf{1{,}304} & \textbf{77\%} & \textbf{23\%} \\
\bottomrule
\end{tabular}
\end{table}

\begin{table}[htbp]
\centering
\caption{Aggregate distribution of prompt structures across all 1{,}304 selected samples.}
\label{tab:av-odyssey-structures}
\footnotesize
\setlength{\tabcolsep}{4pt}
\renewcommand{\arraystretch}{1.2}
\begin{tabular}{@{}lcc@{}}
\toprule
\textbf{Structure} & \textbf{\# Samples} & \textbf{Proportion} \\
\midrule
\colorbox{candbg}{candidates} , \colorbox{qbg}{question}      , \colorbox{refbg}{reference} & 1{,}001 & 77\% \\
\colorbox{qbg}{question}      , \colorbox{refbg}{reference}   , \colorbox{candbg}{candidates}  &   161 & 12\% \\
\colorbox{refbg}{reference}   , \colorbox{candbg}{candidates} , \colorbox{qbg}{question}  &   124 & 10\% \\
\colorbox{refbg}{reference}   , \colorbox{qbg}{question}      , \colorbox{candbg}{candidates}  &    18 & 1\% \\
\bottomrule
\end{tabular}
\end{table}

\begin{table}[htbp]
\centering
\caption{Per-task distribution of prompt structures across the 7 selected tasks. \emph{Cand} = \colorbox{candbg}{candidates}, \emph{Q} = \colorbox{qbg}{question} text, \emph{Ref} = \colorbox{refbg}{reference}. Each row reports the sample counts for each structure observed in that task. Empty cells indicate the structure does not appear in that task.}
\label{tab:av-odyssey-per-task-structures}
\footnotesize
\setlength{\tabcolsep}{4pt}
\renewcommand{\arraystretch}{1.3}
\begin{tabular}{@{}>{\centering\arraybackslash}m{0.20\linewidth}>{\centering\arraybackslash}m{0.16\linewidth}>{\centering\arraybackslash}m{0.16\linewidth}>{\centering\arraybackslash}m{0.16\linewidth}>{\centering\arraybackslash}m{0.16\linewidth}@{}}
\toprule
& \multicolumn{4}{c}{\textbf{Structure}} \\
\cmidrule(lr){2-5}
\textbf{Task} & \textbf{\colorbox{candbg}{Cand}, \colorbox{qbg}{Q}, \colorbox{refbg}{Ref}} & \textbf{\colorbox{qbg}{Q}, \colorbox{refbg}{Ref}, \colorbox{candbg}{Cand}} & \textbf{\colorbox{refbg}{Ref}, \colorbox{candbg}{Cand}, \colorbox{qbg}{Q}} & \textbf{\colorbox{refbg}{Ref}, \colorbox{qbg}{Q}, \colorbox{candbg}{Cand}} \\
\midrule
Instrument Recognition       &  46 & 154 & --- & --- \\
\midrule
Animal Recognition           & 200 & --- & --- & --- \\
\midrule
Material Recognition         & 151 &   7 &  35 &   7 \\
\midrule
Hazard Recognition           &  97 & --- & --- &  11 \\
\midrule
Action Recognition           & 196 & --- & --- & --- \\
\midrule
Music Genre Classification   & 164 & --- &  36 & --- \\
\midrule
Film and Music Matching      & 147 & --- &  53 & --- \\
\midrule
\textbf{Total} & \textbf{1{,}001} & \textbf{161} & \textbf{124} & \textbf{18} \\
\bottomrule
\end{tabular}
\end{table}

\begin{table}[htbp]
\centering
\caption{Cross-task evaluation tasks from AV-Odyssey used in Section~\ref{sec:droptoken}. A Ref → I Cand denotes audio reference to image candidates , and I Ref → A Cand denotes image reference to audio candidates.}
\label{tab:av-odyssey-crosstask}
\footnotesize
\setlength{\tabcolsep}{4pt}
\renewcommand{\arraystretch}{1.1}
\begin{tabular}{@{}lccc@{}}
\toprule
\textbf{Task} & \textbf{\# Samples} & \textbf{A Ref $\to$  I Cand} & \textbf{I Ref $\to$  A Cand} \\
\midrule
Bird Recognition           & 200 & 51\% & 49\% \\
Transportation Recognition & 200 & 52\% & 48\% \\
\bottomrule
\end{tabular}
\end{table}

\subsection{VideoMME (audio-visual video)}
\label{appendix:videomme-data}

VideoMME~\citep{fu2025video} is a video understanding benchmark of four-way multiple-choice questions over videos of short, medium, and long duration. Most of its questions concern visual content, so the answer can be read from the video alone even when the audio track is provided. We include it to examine whether the task-dependent modality contribution observed in Section~\ref{sec:q2} holds on video-reliant tasks, where audio is present in the input but not needed to answer. We use the videos with their audio tracks and without subtitles.

\paragraph{Task selection for attention knockout:} To keep the input length comparable to the other benchmarks, we restrict the analysis to short videos, which are under two minutes long. From this set, we select the three vision-based task types listed in Table~\ref{tab:videomme-selected-tasks}, totaling 421 samples. The knockout results on VideoMME are reported in Appendix~\ref{appendix:qwen3b-videomme}.

\begin{table}[htbp]
\centering
\caption{The 3 tasks selected from VideoMME for attention knockout, restricted to short videos under two minutes.}
\label{tab:videomme-selected-tasks}
\footnotesize
\setlength{\tabcolsep}{6pt}
\renewcommand{\arraystretch}{1.15}
\begin{tabular}{@{}lc@{}}
\toprule
\textbf{Task} & \textbf{\# Samples} \\
\midrule
Attribute Perception & 122 \\
Action Recognition   & 131 \\
Object Recognition   & 168 \\
\midrule
\textbf{Total} & \textbf{421} \\
\bottomrule
\end{tabular}
\end{table}

\clearpage
\section{Ablation on window size for attention knockout}

\begin{figure}[h]
  \centering
  \includegraphics[width=\linewidth]{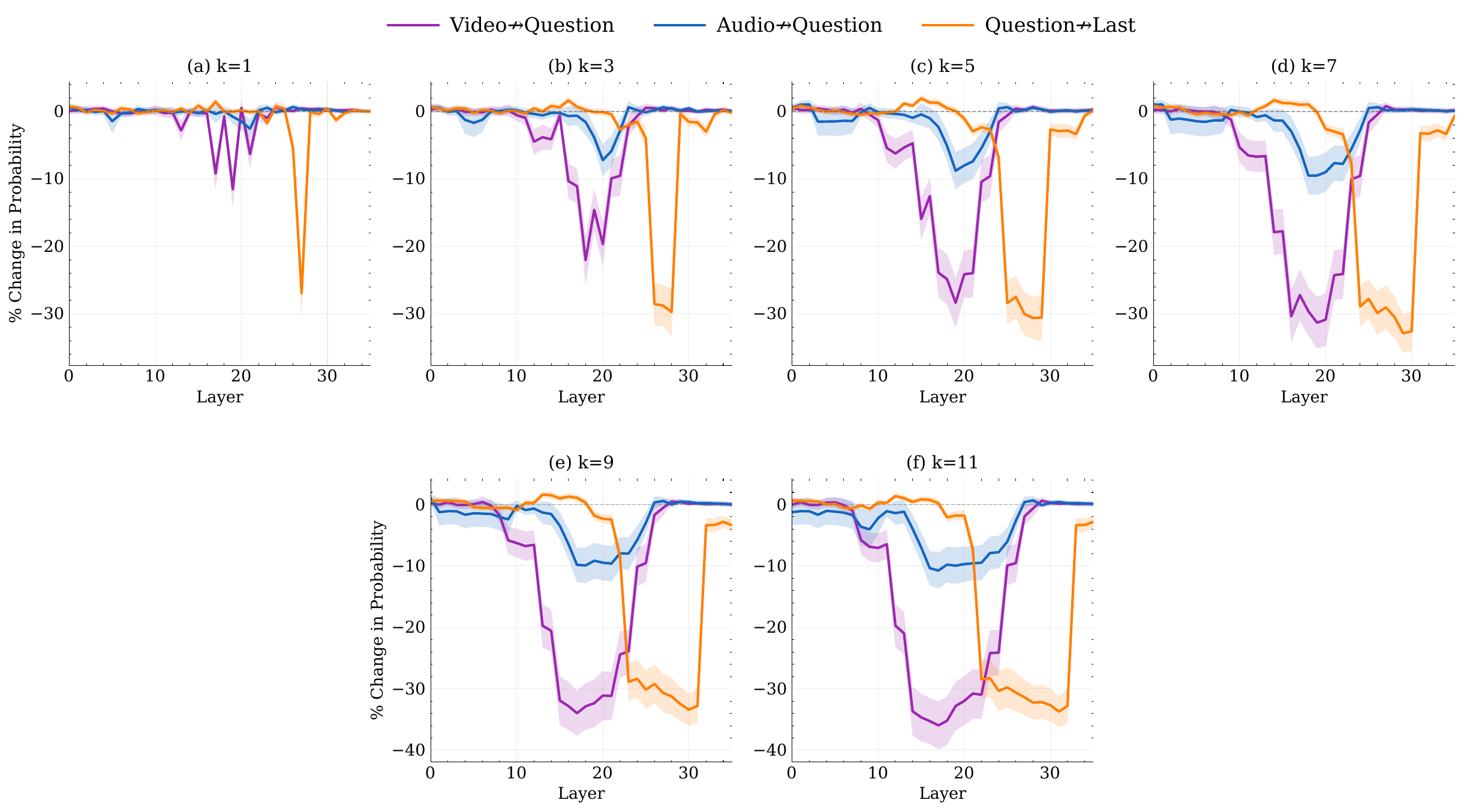}
\caption{Window size ablation on the Speaker Recognition task. Each panel shows the relative change in prediction probability for three pathways (Video$\not\to$Question, Audio$\not\to$Question, Question$\not\to$Last) under a different window size $k$. Source$\not\to$Target indicates blocking attention edges from source tokens to target tokens.}
  \label{fig:appendix_window}
\end{figure}

Following~\cite{geva2023dissecting}, attention knockout is applied within a sliding window of $k$ layers around each target layer. We ablate $k \in \{1, 3, 5, 7, 9, 11\}$ on the Speaker Recognition task to select the most informative window size. Figure~\ref{fig:appendix_window} shows that small windows ($k=1, 3$) produce shallow noisy drops as the narrow block is easily bypassed by remaining attention edges. Large windows ($k=11$) instead blur the layer localization, with the Question$\not\to$Last drop spanning a much wider range and losing the recovery point where the pathway concludes. Both $k=7$ and $k=9$ give clean and well-localized drops, but we adopt $k=7$ as it preserves the recovery point of the Question$\not\to$Last pathway, whereas $k=9$ tends to merge the recovery into the broader drop similarly to $k=11$. The choice of $k=7$ over $k=9$ also accounts for model scale. The 7B variants have only 28 layers compared to 36 in the 3B model, so a window of $k=9$ would cover a relatively larger portion of the network in 7B and obscure the precise layer where each pathway operates. Adopting $k=7$ across both scales preserves localization accuracy in 7B while remaining valid for the 3B model. 
\label{appendix:window-ablation}

\clearpage

\section{Ablation on FPS and spatial resolution}\label{appendix:fps-spatial-ablation}

The main analysis samples videos at 2 FPS at the default spatial resolution (Appendix~\ref{appendix:experimental-details}). We vary the frame rate (Appendix~\ref{appendix:fps}) and the spatial resolution (Appendix~\ref{appendix:spatial}) separately on AV-SpeakerBench with Qwen2.5-Omni 3B to verify that the information flow reported in Section~\ref{sec:q2} is robust to other input configurations. All other settings follow Section~\ref{sec:setup} and Appendix~\ref{appendix:experimental-details}.

\subsection{FPS}\label{appendix:fps}

\begin{figure}[htbp]
  \centering
  \includegraphics[width=\linewidth]{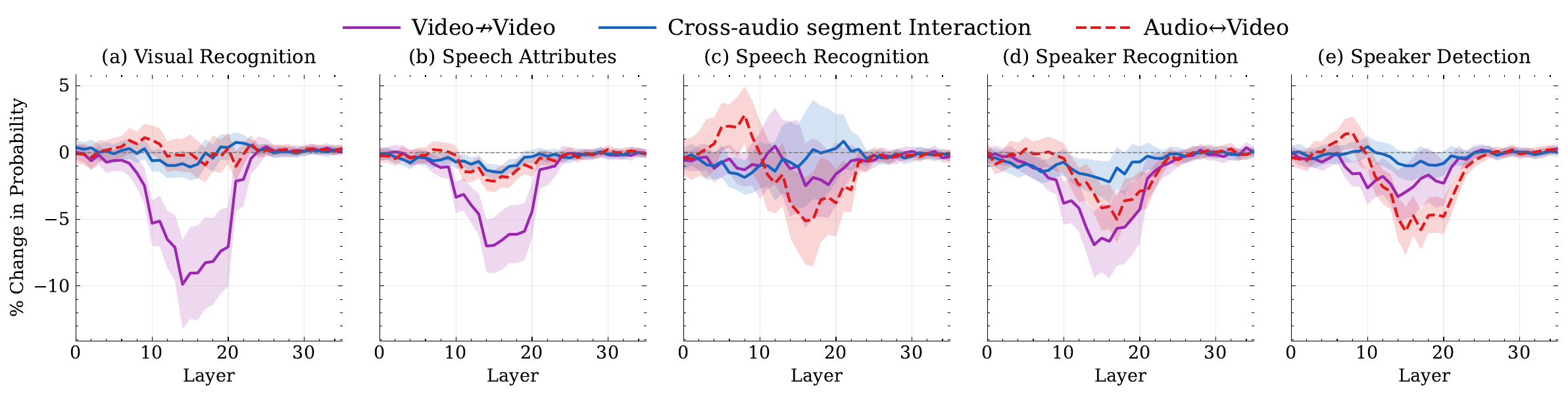}\\[4pt]
  \includegraphics[width=\linewidth]{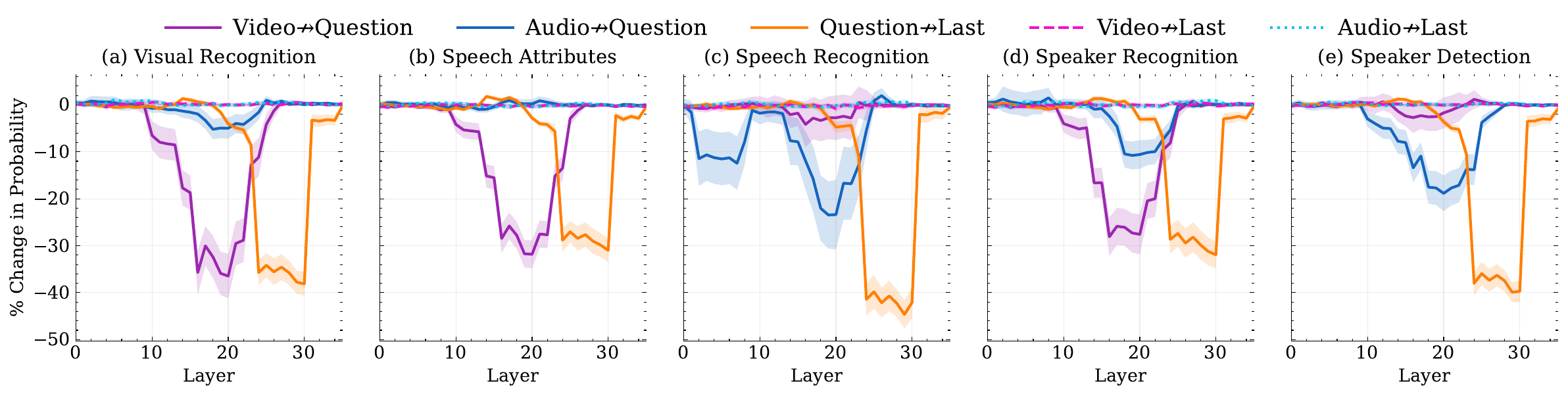}
\caption{\textbf{Qwen2.5-Omni 3B on AV-SpeakerBench at 1 FPS.} \textit{Top:} knockout of within- and cross-modal pathways (Cross-frame, Cross-audio, Audio$\not\leftrightarrow$Video). \textit{Bottom:} knockout of modality and question pathways into the question and the last token (Video$\not\to$Question, Audio$\not\to$Question, Question$\not\to$Last, Video$\not\to$Last, Audio$\not\to$Last). Source$\not\to$Target indicates blocking attention edges from source tokens to target tokens.}
  \label{fig:appendix_1fps_1}
\end{figure}

\begin{figure}[htbp]
  \centering
  \includegraphics[width=\linewidth]{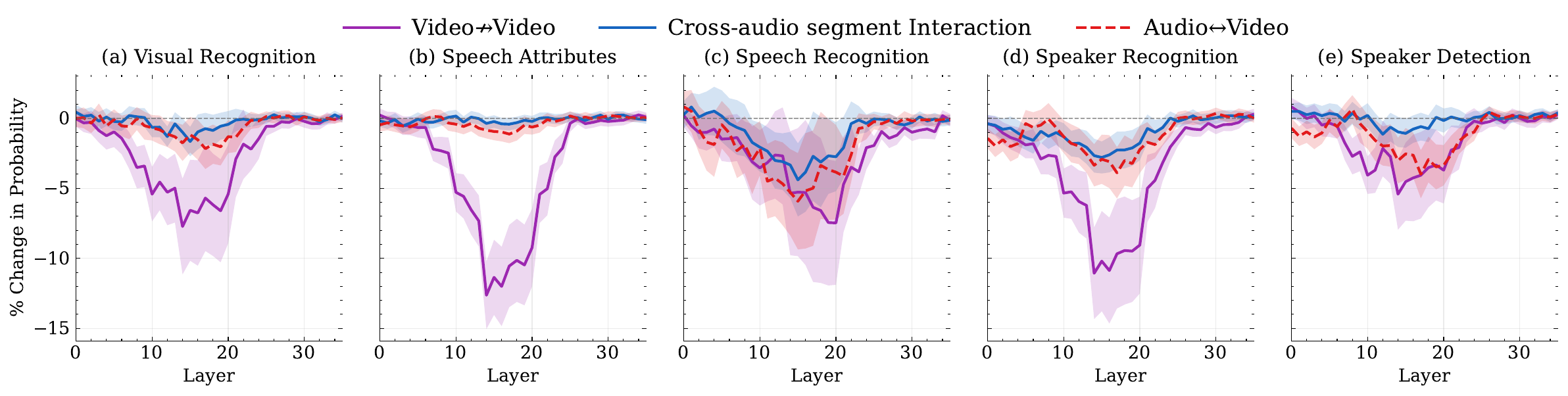}\\[4pt]
  \includegraphics[width=\linewidth]{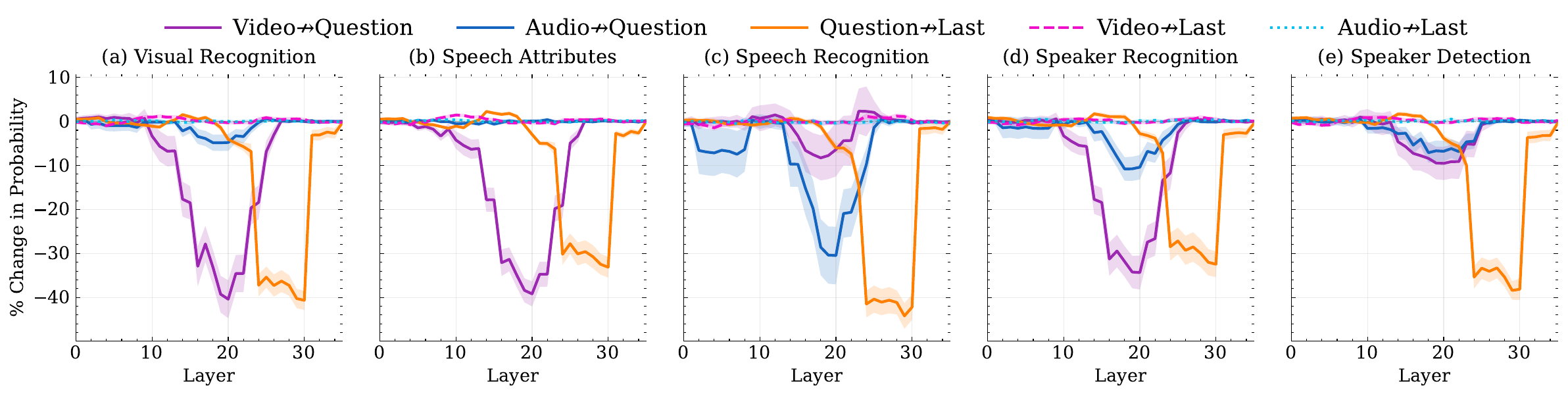}
\caption{\textbf{Qwen2.5-Omni 3B on AV-SpeakerBench at 4 FPS.} \textit{Top:} knockout of within- and cross-modal pathways (Cross-frame, Cross-audio, Audio$\not\leftrightarrow$Video). \textit{Bottom:} knockout of modality and question pathways into the question and the last token (Video$\not\to$Question, Audio$\not\to$Question, Question$\not\to$Last, Video$\not\to$Last, Audio$\not\to$Last). Source$\not\to$Target indicates blocking attention edges from source tokens to target tokens.}
  \label{fig:appendix_4fps_1}
\end{figure}

The frame rate determines the temporal alignment granularity between audio and video tokens, as it sets how many frames each temporal window contains for the same audio segment. We evaluate frame rates of 1 FPS and 4 FPS at the default spatial resolution, halving and doubling the default rate. Figures~\ref{fig:appendix_1fps_1} and~\ref{fig:appendix_4fps_1} report the knockout results. At both frame rates, the route, the layers where each transfer localizes, and the task-dependent modality contribution all remain consistent with Section~\ref{sec:q2}. Only the magnitudes vary, following how much visual content each frame rate provides. At 1 FPS, the video contribution to the question weakens across all three tasks, most clearly on Speech Recognition, where the Video$\not\to$Question drop falls from $-10\%$ at 2 FPS to $-4.2\%$, consistent with sparser frames providing less visual content. At 4 FPS, the denser frames slightly strengthen the video contribution on the visually grounded tasks, while the audio contribution and the Question$\not\to$Last drop remain close to the default setting.

\subsection{Spatial resolution}\label{appendix:spatial}

\begin{figure}[htbp]
  \centering
  \includegraphics[width=\linewidth]{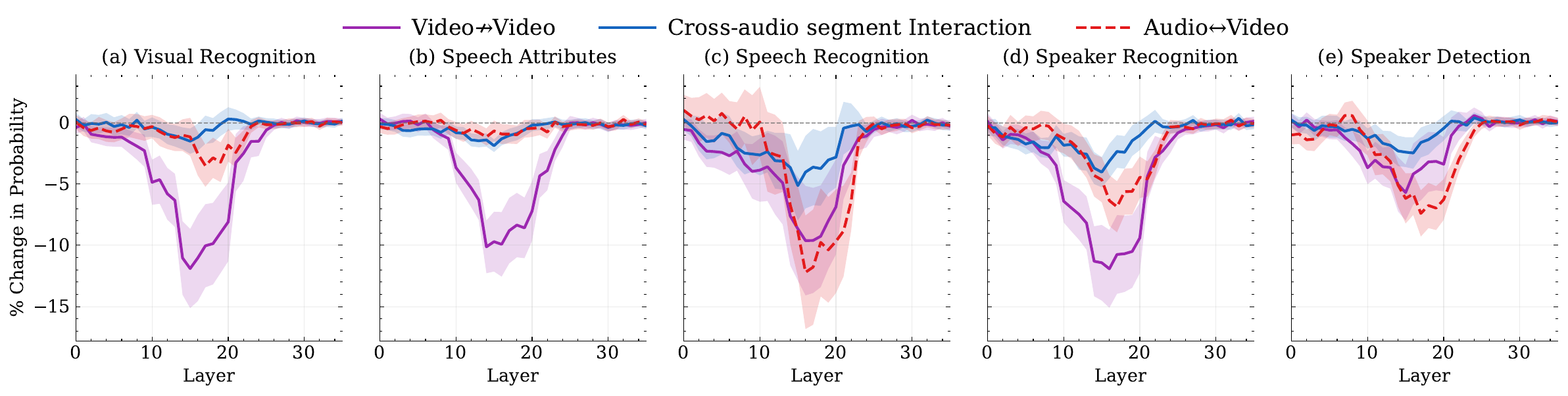}\\[4pt]
  \includegraphics[width=\linewidth]{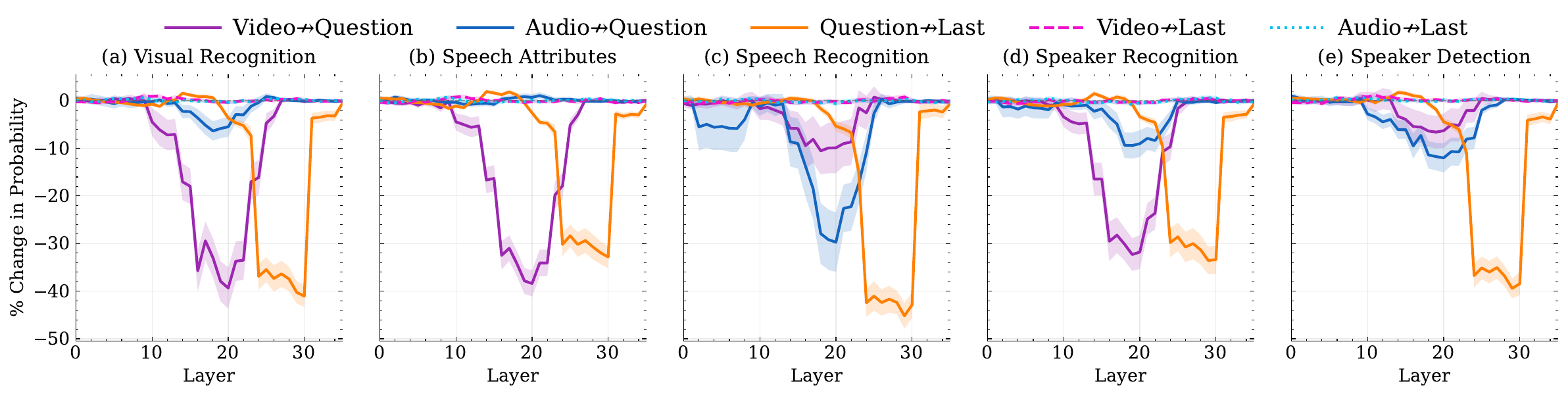}
\caption{\textbf{Qwen2.5-Omni 3B on AV-SpeakerBench at 2 FPS with reduced spatial resolution} (168$\times$280 pixels per frame). \textit{Top:} knockout of within- and cross-modal pathways (Cross-frame, Cross-audio, Audio$\not\leftrightarrow$Video). \textit{Bottom:} knockout of modality and question pathways into the question and the last token (Video$\not\to$Question, Audio$\not\to$Question, Question$\not\to$Last, Video$\not\to$Last, Audio$\not\to$Last). Source$\not\to$Target indicates blocking attention edges from source tokens to target tokens.}
  \label{fig:appendix_spatial1}
\end{figure}

The spatial resolution determines how much visual detail each frame carries. We reduce the resolution of each frame from 224$\times$420 to 168$\times$280 pixels, which reduces the grid of visual tokens per frame from 12$\times$12 to 8$\times$8 and halves the per-frame visual token budget, while keeping the frame rate at 2 FPS. Figure~\ref{fig:appendix_spatial1} reports the knockout results. The route, the layers where each transfer localizes, and the task-dependent modality contribution all remain consistent with Section~\ref{sec:q2}. The magnitudes vary only slightly from the default setting, with differences of about 1\% across tasks, suggesting that the coarser frames still carry the visual content these tasks rely on.

Together, these ablations show that the sequential Modalities $\to$ Question $\to$ Last route, the layers where each transfer localizes, and the task-dependent modality contribution reported in Section~\ref{sec:q2} all hold across frame rates and spatial resolutions, with only the magnitudes varying according to the amount of visual content provided.

\section{More results on token discarding cross-dataset generalization on WorldSense}\label{appendix:token_discard}

Table~\ref{tab:tokendrop} reports two WorldSense categories, Video Emotions and Audio Change, as the cross-dataset evaluation of token discarding. Table~\ref{tab:worldsense_percategory} extends this evaluation to all 26 categories of WorldSense, using the same discarding configuration as in Section~\ref{sec:droptoken}. The overall accuracy remains stable and slightly improves in all four configurations, from 45.75 at baseline to 46.23 when discarding all token types.

\begin{table}[ht]
\centering
\caption{\textbf{Per-category cross-dataset results on WorldSense.} Expanded breakdown of the cross-dataset evaluation in Table~\ref{tab:tokendrop} (Video setting: Discard Video \& Audio at $L=26$, Ques at $L=29$). Numbers in parentheses show change from baseline ({\color{gain}green = improvement}, {\color{drop}red = drop}); \underline{underlined} indicates no change. Best per row in \textbf{bold}. \textbf{Ques}: non-option Question tokens; \textbf{All}: all token types together.}
\label{tab:worldsense_percategory}
\footnotesize
\setlength{\tabcolsep}{3pt}
\renewcommand{\arraystretch}{1.1}
\begin{tabular}{@{}lccccc@{}}
\toprule
\textbf{Category} & \textbf{Baseline} & \textbf{Discard Ques} & \textbf{Discard Audio} & \textbf{Discard Video} & \textbf{Discard All} \\
\midrule
Spatial Relation & 24.14 & \underline{24.14} & \underline{24.14} & \underline{24.14} & \underline{24.14} \\
Video Emotions & 66.67 & \underline{66.67} & \underline{66.67} & \underline{66.67} & \underline{66.67} \\
Event Sorting & 56.52 & \textbf{58.70} \tiny{({\color{gain}+2.18})} & \underline{56.52} & 54.35 \tiny{({\color{drop}-2.17})} & \underline{56.52} \\
Event Recognition & 46.67 & \underline{46.67} & 50.00 \tiny{({\color{gain}+3.33})} & \underline{46.67} & \textbf{53.33} \tiny{({\color{gain}+6.66})} \\
Attribute Recognition & 51.92 & \underline{51.92} & \underline{51.92} & \underline{51.92} & \underline{51.92} \\
Audio Recognition & 63.64 & 62.50 \tiny{({\color{drop}-1.14})} & 62.50 \tiny{({\color{drop}-1.14})} & \textbf{\underline{63.64}} & \textbf{\underline{63.64}} \\
Human-object Interaction & 43.90 & \underline{43.90} & \underline{43.90} & \underline{43.90} & \underline{43.90} \\
Audio Source Localization & 53.12 & \underline{53.12} & \underline{53.12} & 51.56 \tiny{({\color{drop}-1.56})} & \textbf{54.69} \tiny{({\color{gain}+1.57})} \\
Audio Change & 42.22 & 40.00 \tiny{({\color{drop}-2.22})} & 40.00 \tiny{({\color{drop}-2.22})} & 40.00 \tiny{({\color{drop}-2.22})} & \textbf{\underline{42.22}} \\
Relation Reasoning & 45.00 & \textbf{50.00} \tiny{({\color{gain}+5.00})} & \textbf{50.00} \tiny{({\color{gain}+5.00})} & \underline{45.00} & \textbf{50.00} \tiny{({\color{gain}+5.00})} \\
Action Counting & 30.43 & \textbf{32.61} \tiny{({\color{gain}+2.18})} & \underline{30.43} & \textbf{32.61} \tiny{({\color{gain}+2.18})} & \textbf{32.61} \tiny{({\color{gain}+2.18})} \\
Text and Diagram Understanding & 51.22 & \textbf{\underline{51.22}} & 46.34 \tiny{({\color{drop}-4.88})} & 48.78 \tiny{({\color{drop}-2.44})} & 46.34 \tiny{({\color{drop}-4.88})} \\
Attribute Reasoning & 47.62 & \textbf{50.00} \tiny{({\color{gain}+2.38})} & \underline{47.62} & \underline{47.62} & \underline{47.62} \\
Causal Reasoning & 48.65 & \underline{48.65} & \underline{48.65} & \underline{48.65} & \underline{48.65} \\
Human Emotions & 37.93 & \underline{37.93} & \textbf{41.38} \tiny{({\color{gain}+3.45})} & \underline{37.93} & \underline{37.93} \\
Temporal Localization & 33.93 & 35.71 \tiny{({\color{gain}+1.78})} & 35.71 \tiny{({\color{gain}+1.78})} & \textbf{37.50} \tiny{({\color{gain}+3.57})} & 35.71 \tiny{({\color{gain}+1.78})} \\
Anomaly Recognition & 51.85 & 48.15 \tiny{({\color{drop}-3.70})} & \textbf{\underline{51.85}} & \textbf{\underline{51.85}} & 48.15 \tiny{({\color{drop}-3.70})} \\
Audio Counting & 39.47 & 42.11 \tiny{({\color{gain}+2.64})} & 42.11 \tiny{({\color{gain}+2.64})} & \textbf{44.74} \tiny{({\color{gain}+5.27})} & 42.11 \tiny{({\color{gain}+2.64})} \\
Human Interaction & 30.56 & \underline{30.56} & \underline{30.56} & \underline{30.56} & \underline{30.56} \\
Temporal Prediction & 50.00 & \underline{50.00} & \underline{50.00} & \underline{50.00} & \underline{50.00} \\
Hallucination & 23.81 & \underline{23.81} & \underline{23.81} & \underline{23.81} & \underline{23.81} \\
Object Counting & 27.66 & 25.53 \tiny{({\color{drop}-2.13})} & 25.53 \tiny{({\color{drop}-2.13})} & \textbf{\underline{27.66}} & 25.53 \tiny{({\color{drop}-2.13})} \\
Object State Change & 56.00 & \underline{56.00} & \underline{56.00} & \underline{56.00} & \underline{56.00} \\
Emotion Change & 62.16 & \underline{62.16} & \textbf{64.86} \tiny{({\color{gain}+2.70})} & \underline{62.16} & \textbf{64.86} \tiny{({\color{gain}+2.70})} \\
Scene Recognition & 59.26 & \textbf{62.96} \tiny{({\color{gain}+3.70})} & \underline{59.26} & \underline{59.26} & \textbf{62.96} \tiny{({\color{gain}+3.70})} \\
Object Existence Recognition & 38.71 & \textbf{41.94} \tiny{({\color{gain}+3.23})} & \textbf{41.94} \tiny{({\color{gain}+3.23})} & \underline{38.71} & \underline{38.71} \\
\midrule
\textbf{Overall} & 45.75 & 46.13 \tiny{({\color{gain}+0.38})} & 45.94 \tiny{({\color{gain}+0.19})} & 45.85 \tiny{({\color{gain}+0.10})} & \textbf{46.23} \tiny{({\color{gain}+0.48})} \\
\bottomrule
\end{tabular}
\end{table}

\clearpage
\section{Additional results on Qwen2.5-Omni 3B}
\label{appendix:qwen3b-additional}

This appendix provides additional knockout analyses for Qwen2.5-Omni 3B, extending the AV-SpeakerBench~\cite{nguyen2025see} results from Section~\ref{sec:av-analysis} and reporting attention knockout on WorldSense~\cite{hong2025worldsense} and AV-Odyssey~\cite{gong2024av}. While AV-Odyssey is also analyzed in the main paper, the results there are averaged across tasks. Here we instead report per-task knockouts to show that the same pattern holds for each individual task.

\subsection{Extended knockout analyses on AV-SpeakerBench (audio-visual video)}
\label{appendix:more_knockout}
This subsection extends the AV-SpeakerBench analysis of Section~\ref{sec:av-analysis} with additional knockouts on cross-modal directions, joint audio-visual routing, and the question's internal components.

\subsubsection{Further analysis of cross-modal directions and joint audio-visual routing}
\label{appendix:v1}

\begin{figure}[htbp]
  \centering
  \includegraphics[width=\linewidth]{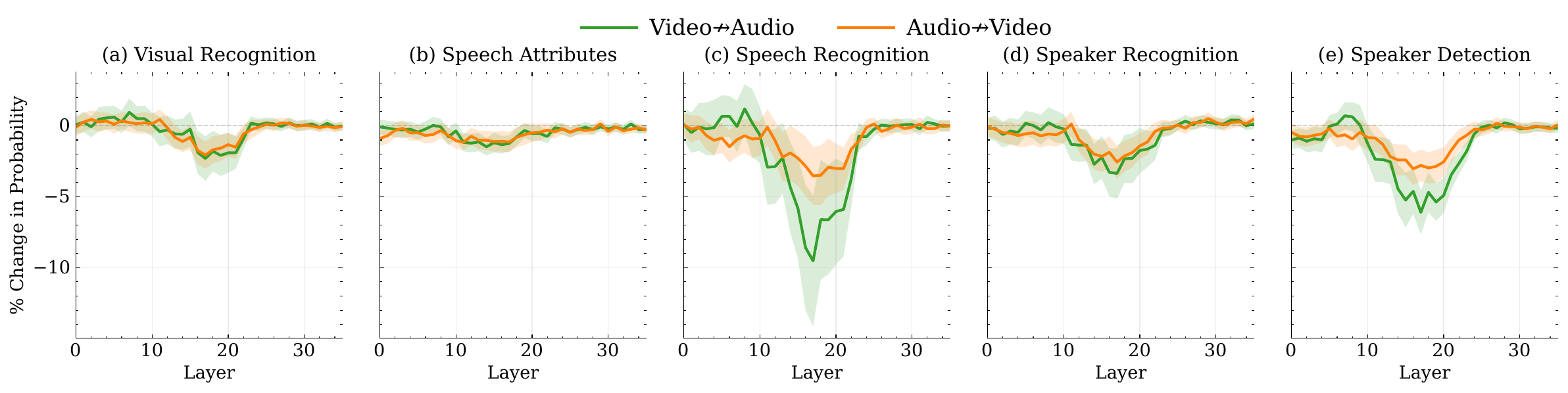}\\[4pt]
  \includegraphics[width=\linewidth]{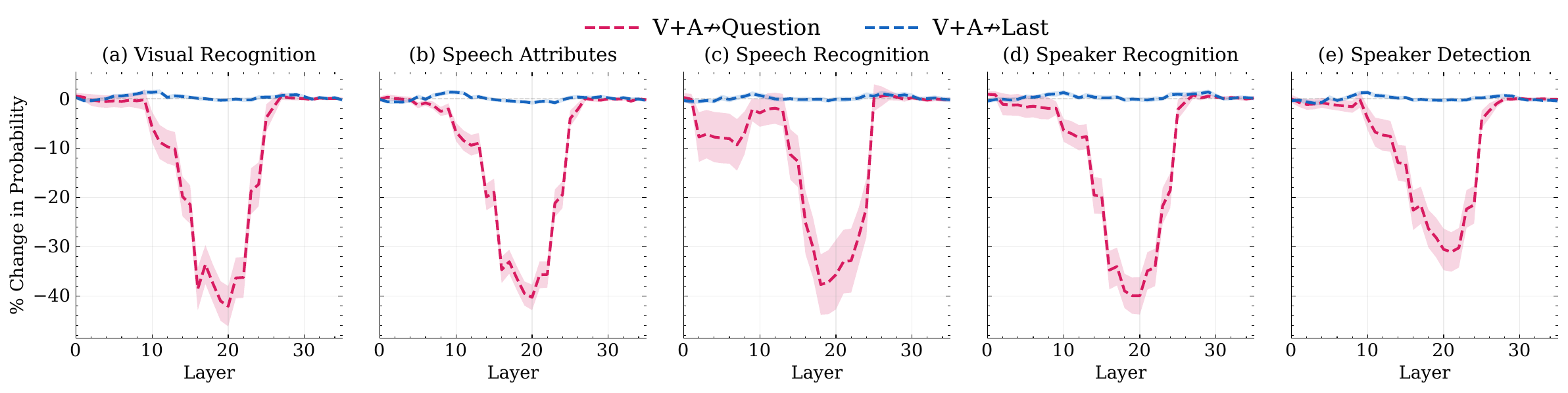}
  \caption{Additional knockout analysis on Qwen2.5-Omni 3B. \textit{Top:} direction-specific cross-modal knockouts (Video$\not\to$Audio and Audio$\not\to$Video). \textit{Bottom:} joint knockouts where audio and video serve as sources (V+A$\not\to$Question and V+A$\not\to$Last). Source$\not\to$Target indicates blocking attention edges from source tokens to target tokens.}
  \label{fig:appendix_v1}
\end{figure}

We extend the analysis of Section~\ref{sec:av-analysis} with two follow-up analyses. The top panel of Figure~\ref{fig:appendix_v1} decomposes the bidirectional cross-modal knockout of Section~\ref{sec:q1} into its two unidirectional components, Video$\not\to$Audio and Audio$\not\to$Video. The Video$\to$Audio direction is the dominant one on tasks that require fine-grained audio-visual alignment (Speech Recognition, Speaker Detection), accounting for most of the bidirectional cross-modal effect observed in Section~\ref{sec:q1}, while Audio$\to$Video plays a smaller role. The remaining tasks remain near zero in both directions, consistent with Section~\ref{sec:q1}. We hypothesize that this asymmetry stems from the input ordering of the time-aligned sequence, where each video frame is followed by its corresponding audio segment. Audio tokens can therefore attend back to the time-aligned video frame, while video tokens can only attend to audio segments from earlier time steps and not to the time-aligned audio segment, weakening the Audio$\to$Video direction. The bottom panel groups audio and video into a single source (V+A$\not\to$Question and V+A$\not\to$Last). V+A$\not\to$Question produces large mid-layer drops across all tasks, while V+A$\not\to$Last produces near-zero changes, confirming that audio-visual information reaches the prediction through the question rather than directly, in line with the Modalities $\to$ Question $\to$ Last route established in Section~\ref{sec:q2}.

\subsubsection{Further analysis of how audio-visual information reaches the prediction via question components}
\label{appendix:question-components}

\begin{figure}[htbp]
  \centering
  \includegraphics[width=\linewidth]{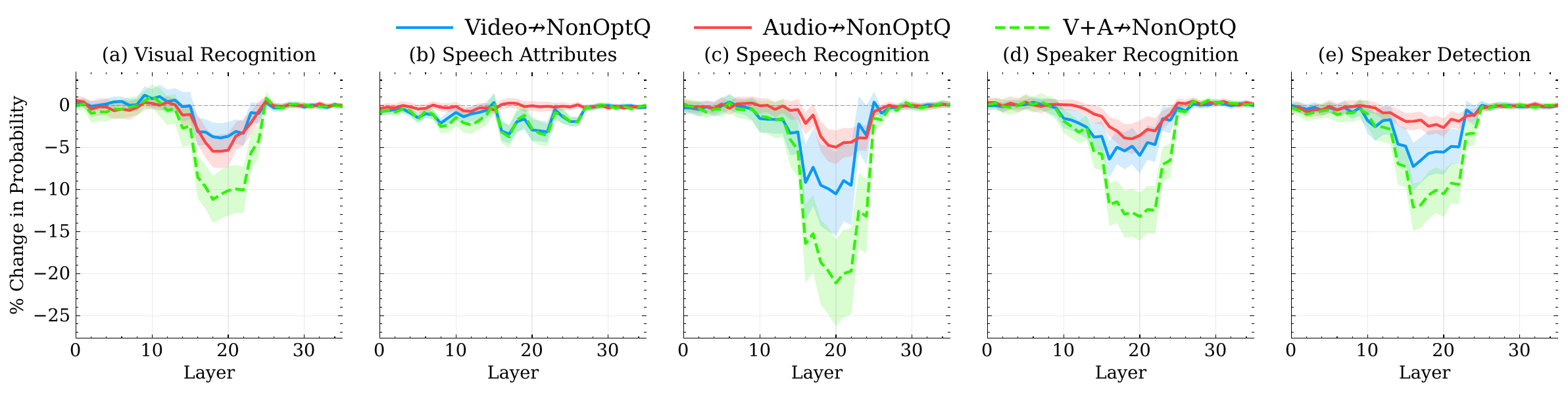}\\[4pt]
  \includegraphics[width=\linewidth]{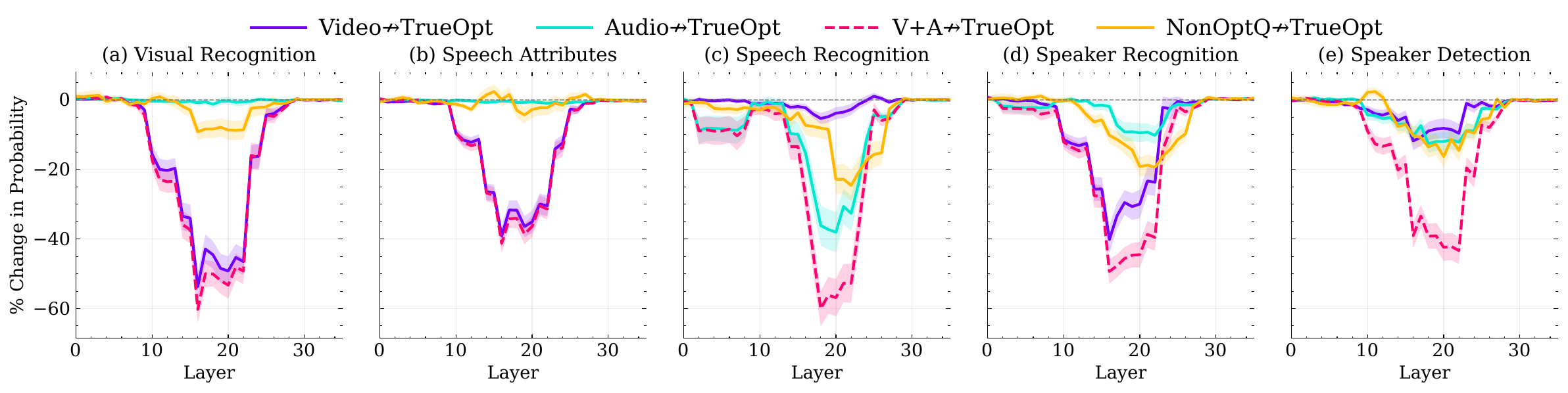}\\[4pt]
  \includegraphics[width=\linewidth]{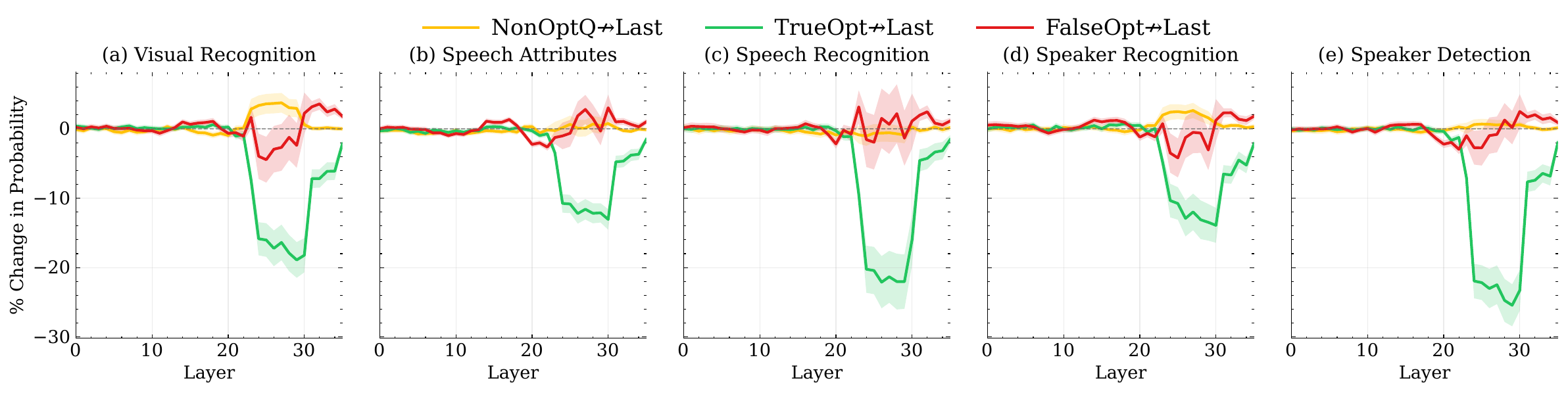}
  \caption{Component-level question-internal knockouts on Qwen2.5-Omni 3B. \textit{Top:} Source$\not\to$NonOptQ. \textit{Middle:} Source$\not\to$TrueOpt. \textit{Bottom:} Source$\not\to$Last. Source$\not\to$Target indicates blocking attention edges from source tokens to target tokens.}
  \label{fig:appendix_v4}
\end{figure}

Section~\ref{sec:q2} treats the question as a single aggregator. Here we decompose it into three components, the correct-option letter (TrueOpt), the incorrect-option letters (FalseOpt), and the non-option question text (NonOptQ), and trace how audio-visual content reaches the prediction within the question. Figure~\ref{fig:appendix_v4} reveals two routes converging at the correct-option letter. The direct route, Modalities $\to$ TrueOpt, sends modality content straight to the option letter. The indirect route, Modalities $\to$ NonOptQ $\to$ TrueOpt, first passes through the non-option question text before reaching the option letter. Both routes are active on Visual Recognition, Speech Recognition, Speaker Recognition, and Speaker Detection, while Speech Attributes relies on the direct route only, with negligible flow through the non-option question text. The correct-option letter therefore acts as the local aggregator, absorbing audio-visual content from these routes at mid layers before being read by the last token at late layers. The incorrect-option letters and the non-option question text never flow to the prediction directly, only through the correct-option letter. This component-level view is consistent with previous work in VideoLLMs~\cite{kim2025map}, which also reports the option token as the decisive integration point with both direct and indirect routes from video to the option. Our analysis extends this observation to the audio-visual setting, where the same dual-route structure governs how both modalities reach the option letter.

\clearpage

\subsection{VideoMME}\label{appendix:qwen3b-videomme}

\begin{figure}[htbp]
  \centering
  \includegraphics[width=0.8\linewidth]{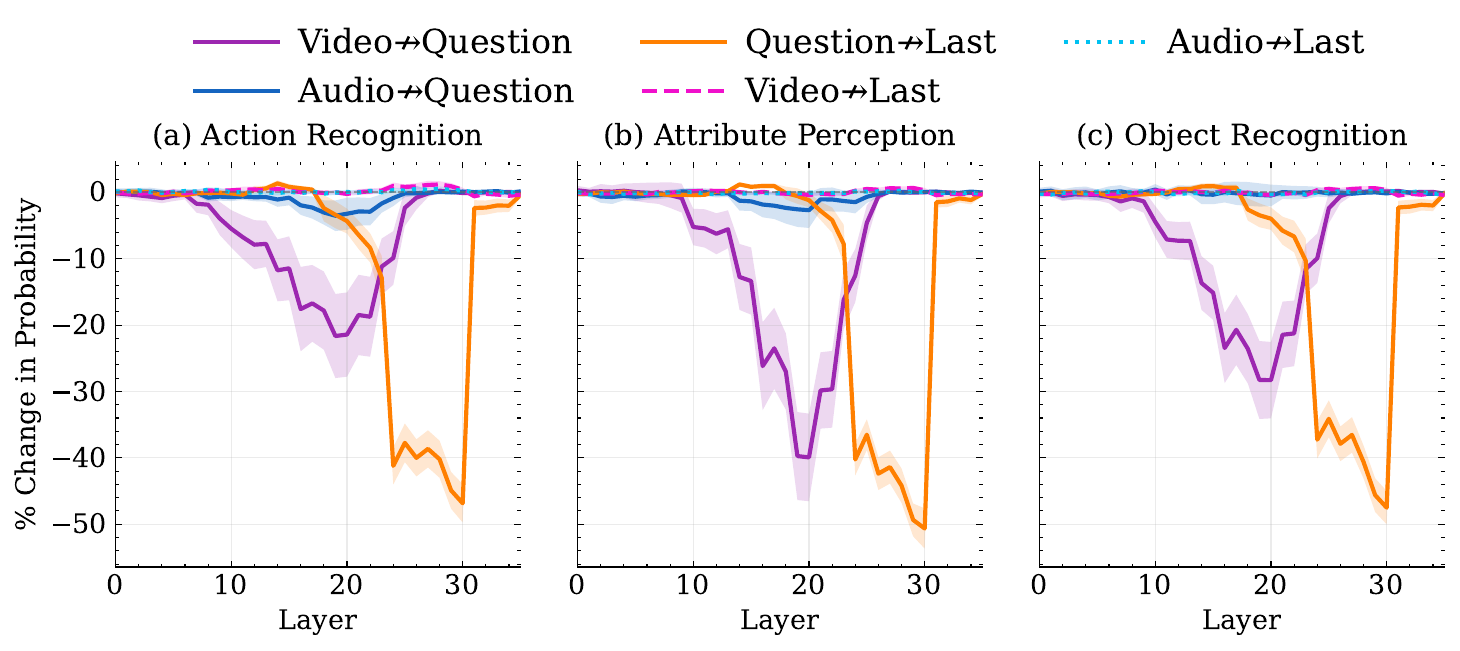}
\caption{\textbf{Qwen2.5-Omni 3B on VideoMME.} Knockout of modality and question pathways into the question and the last token (Video$\not\to$Question, Audio$\not\to$Question, Question$\not\to$Last, Video$\not\to$Last, Audio$\not\to$Last). Source$\not\to$Target indicates blocking attention edges from source tokens to target tokens.}
  \label{fig:video-mme2}
\end{figure}

We further examine whether our findings hold across benchmarks by repeating the analysis of Section~\ref{sec:av-analysis} on VideoMME~\cite{fu2025video} at 2 FPS, using three vision-based categories, Attribute Perception, Action Recognition, and Object Recognition (details in Appendix~\ref{appendix:videomme-data}). Figures~\ref{fig:video-mme1} and~\ref{fig:video-mme2} report the knockout results. The route and the layers where each transfer localizes remain consistent with Section~\ref{sec:q2}. All three categories can be answered from the video alone, and the flow follows accordingly, with the contribution coming almost entirely from video and little from audio, consistent with the task-dependent modality contribution observed on AV-SpeakerBench.

\begin{figure}[htbp]
  \centering
  \includegraphics[width=0.8\linewidth]{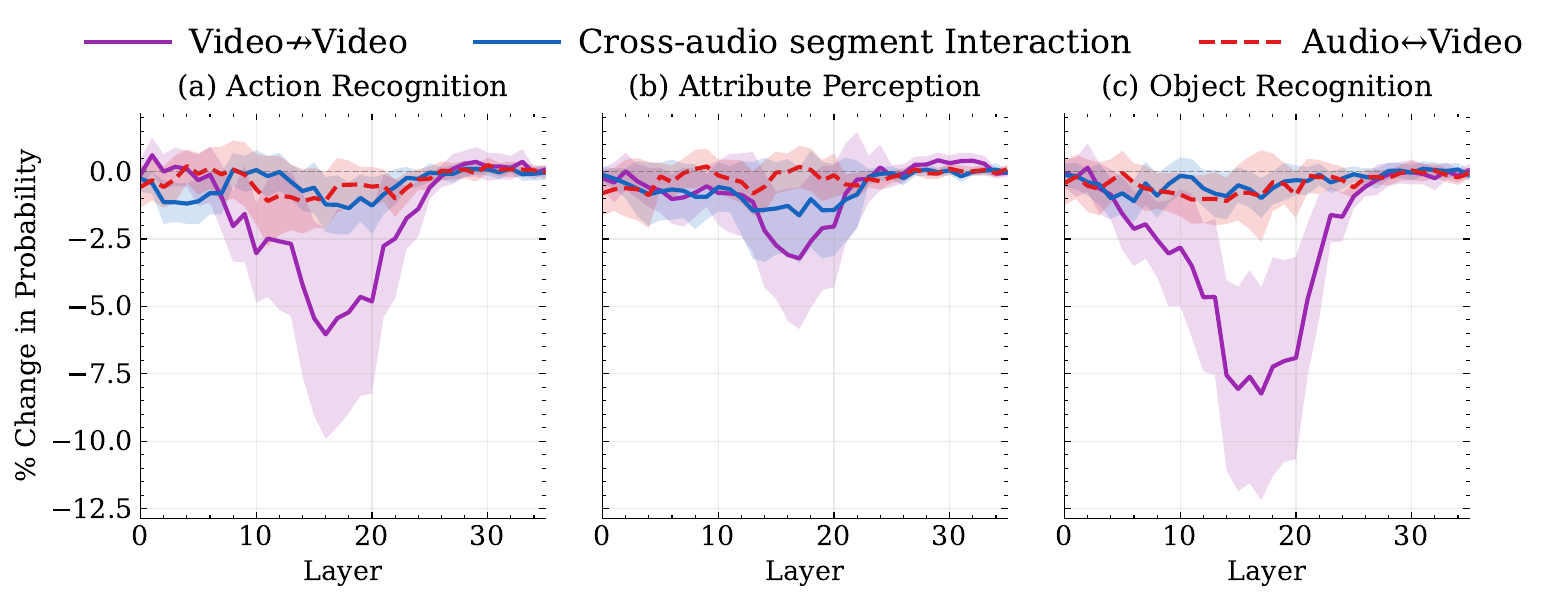}
\caption{\textbf{Qwen2.5-Omni 3B on VideoMME.} Knockout of within- and cross-modal pathways (Cross-frame, Cross-audio, Audio$\not\leftrightarrow$Video). Source$\not\to$Target indicates blocking attention edges from source tokens to target tokens.}
  \label{fig:video-mme1}
\end{figure}

\clearpage
\subsection{WorldSense (audio-visual video)}
\label{appendix:qwen3b-worldsense}

\begin{figure}[htbp]
  \centering
  \includegraphics[width=\linewidth]{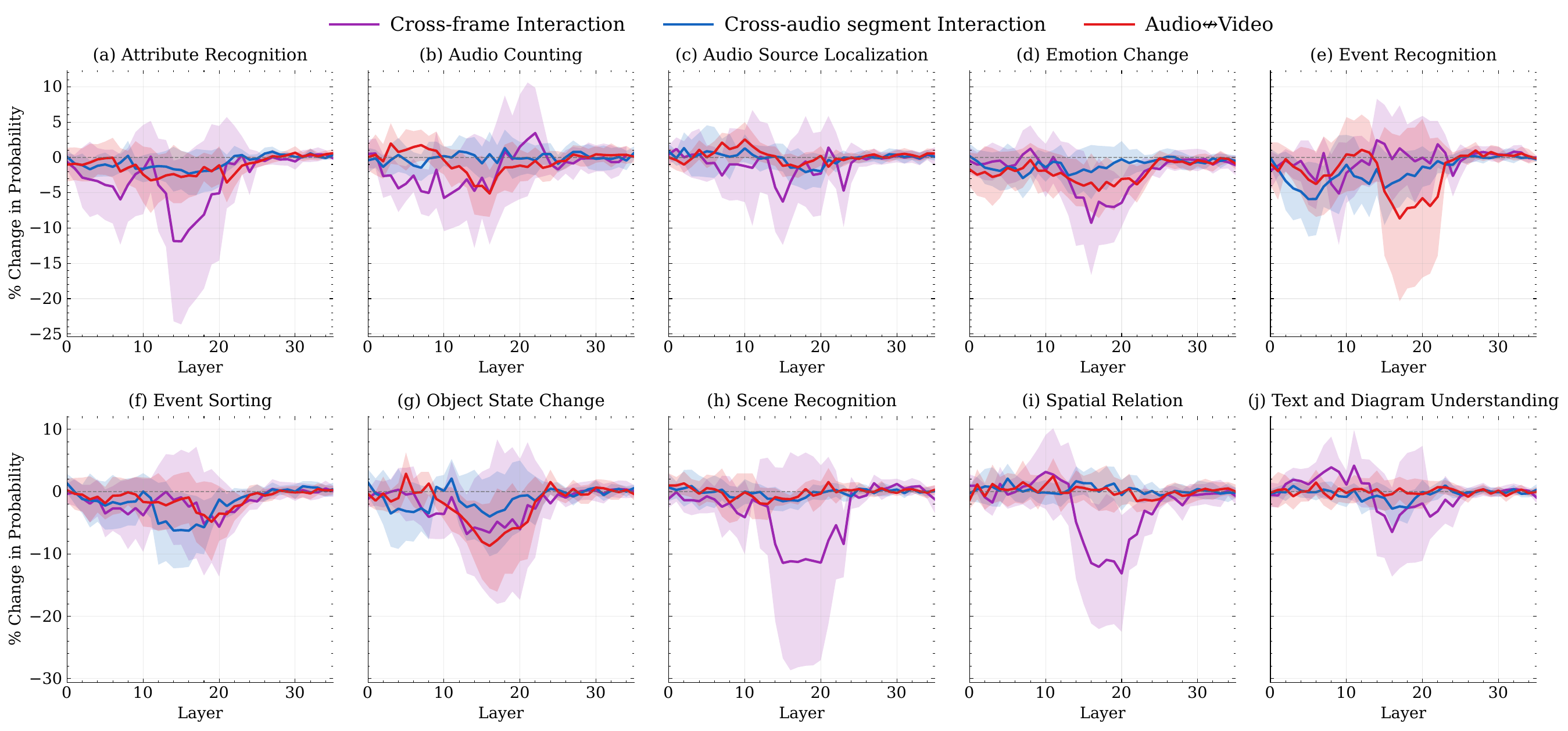}\\[4pt]
  \includegraphics[width=\linewidth]{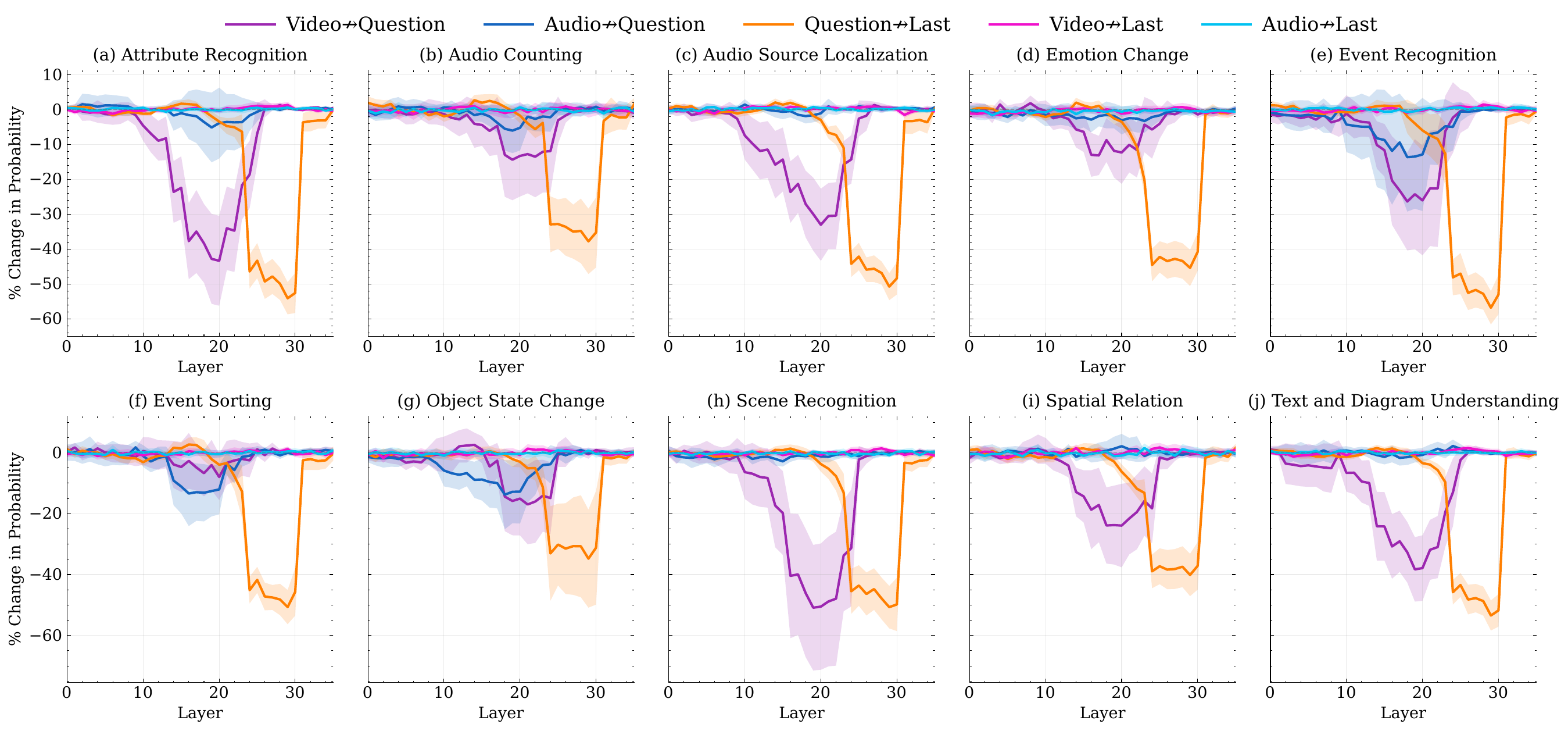}
  \caption{\textbf{Qwen2.5-Omni 3B on WorldSense.} Knockout of within- and cross-modal pathways (Cross-frame, Cross-audio segment, Audio$\leftrightarrow$Video) and of modality and question pathways into the last token (Video$\not\to$Question, Audio$\not\to$Question, Question$\not\to$Last, Video$\not\to$Last, Audio$\not\to$Last). Source$\not\to$Target indicates blocking attention edges from source tokens to target tokens.}
  \label{fig:qwen3B_worldsense_video_knock_group1}
\end{figure}

\begin{figure}[htbp]
  \centering
  \includegraphics[width=\linewidth]{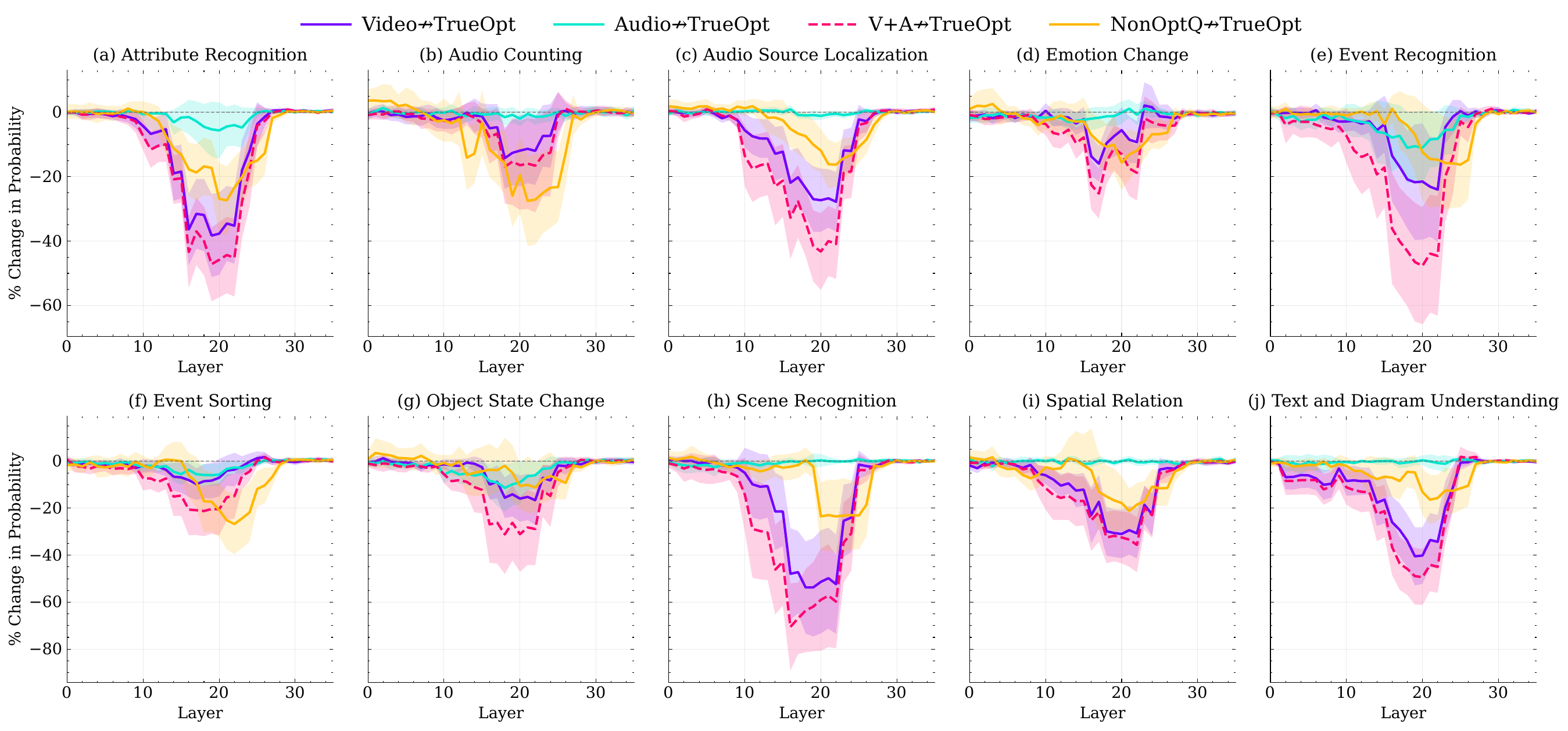}\\[4pt]
  \includegraphics[width=\linewidth]{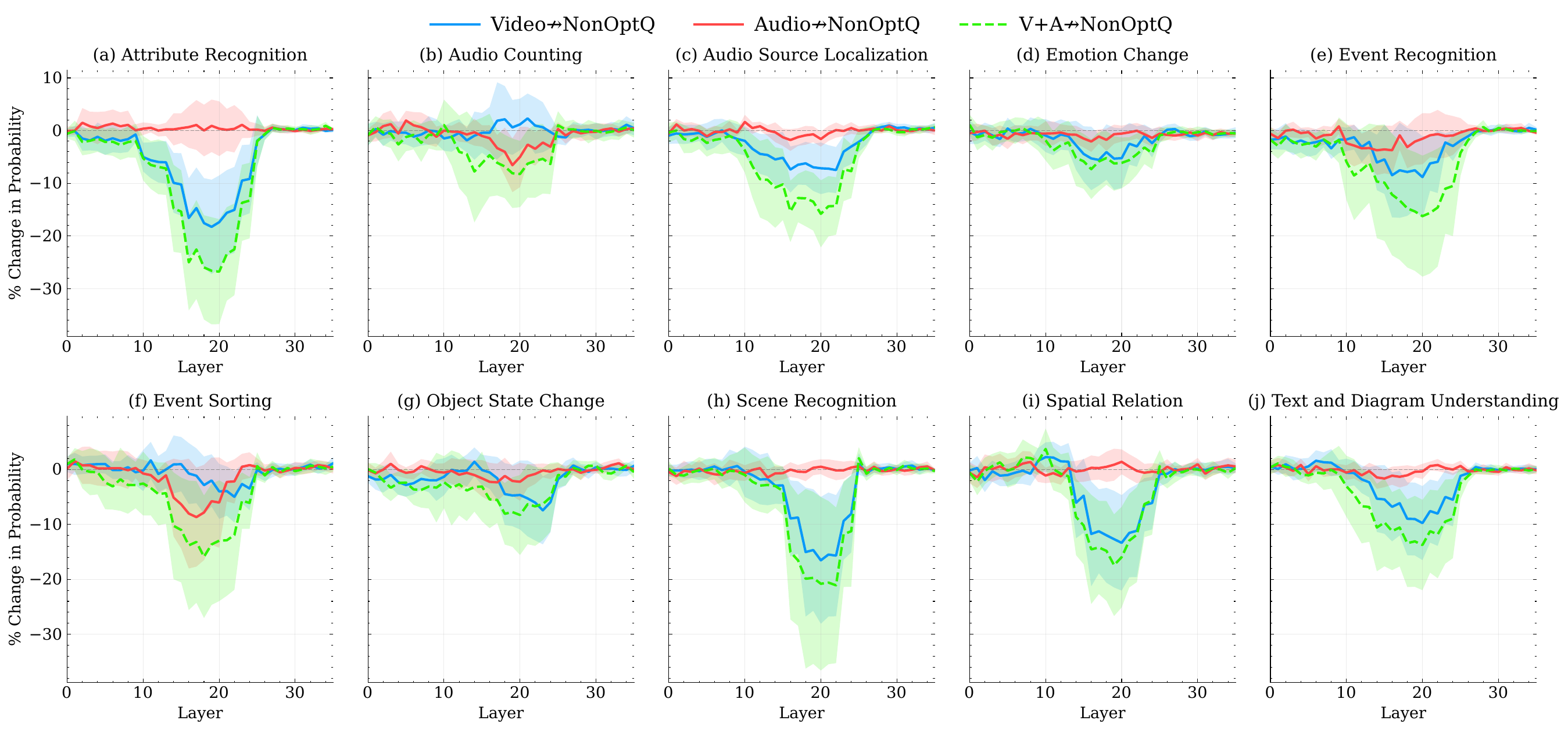}\\[4pt]
  \includegraphics[width=\linewidth]{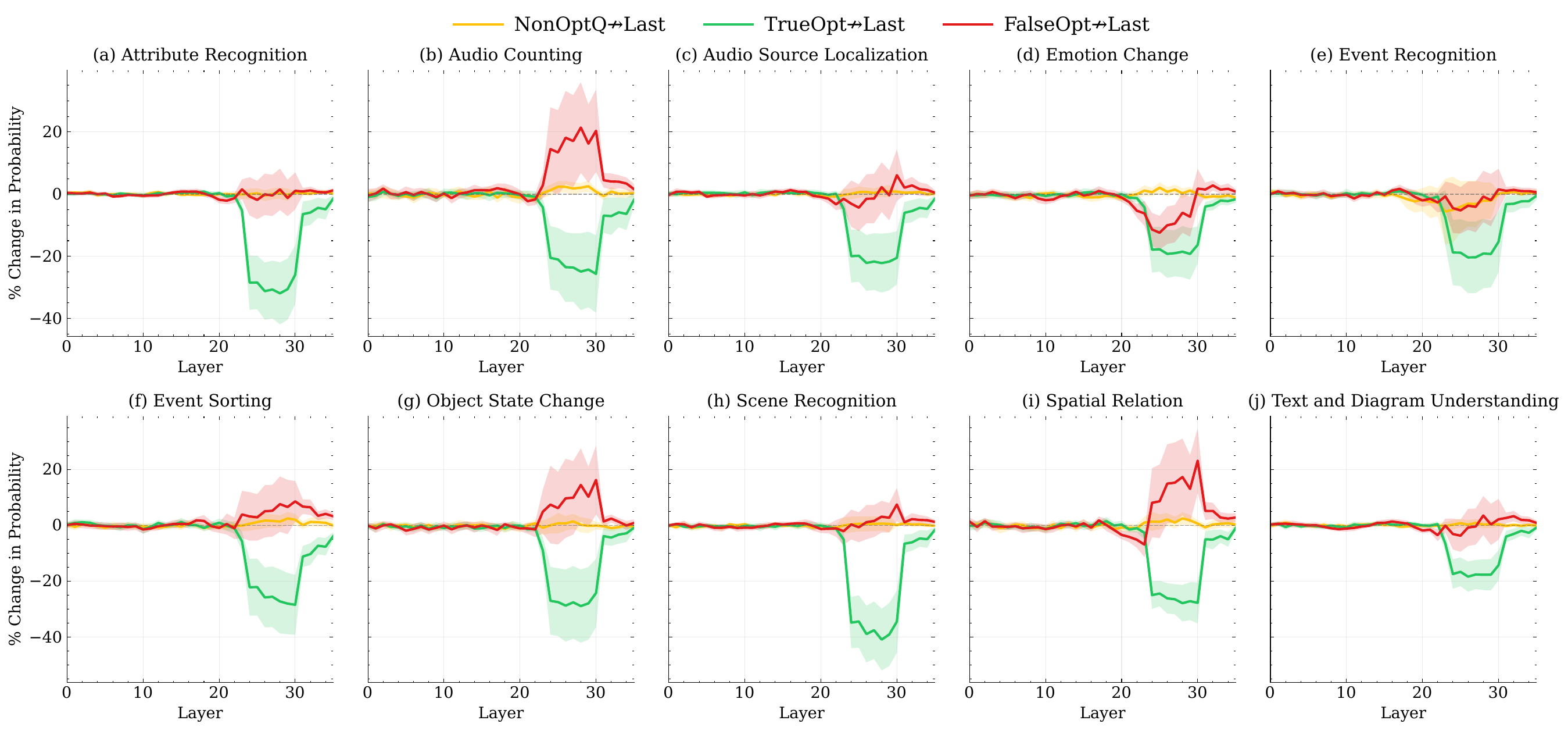}
  \caption{\textbf{Qwen2.5-Omni 3B on WorldSense.} Modality and question pathways into the correct option letter (Video$\not\to$TrueOpt, Audio$\not\to$TrueOpt, NonOptQ$\not\to$TrueOpt, V+A$\not\to$TrueOpt); modality pathways into the non-option question text (Video$\not\to$NonOptQ, Audio$\not\to$NonOptQ, V+A$\not\to$NonOptQ); and question-internal pathways into the last token (TrueOpt$\not\to$Last, FalseOpt$\not\to$Last, NonOptQ$\not\to$Last). Source$\not\to$Target indicates blocking attention edges from source tokens to target tokens.}
  \label{fig:qwen3B_worldsense_video_knock_group2}
\end{figure}

Figures~\ref{fig:qwen3B_worldsense_video_knock_group1} and~\ref{fig:qwen3B_worldsense_video_knock_group2} report the WorldSense knockout results. Figure~\ref{fig:qwen3B_worldsense_video_knock_group1} covers within- and cross-modal pathways together with modality and question pathways into the last token, while Figure~\ref{fig:qwen3B_worldsense_video_knock_group2} covers the question-internal pathways and the pathways into the correct option letter and the non-option question text.

\clearpage
\subsection{AV-Odyssey (multi-input audio-visual interleaved)}
\label{appendix:qwen3b-avodyssey}

\begin{figure}[htbp]
  \centering
  \includegraphics[width=\linewidth]{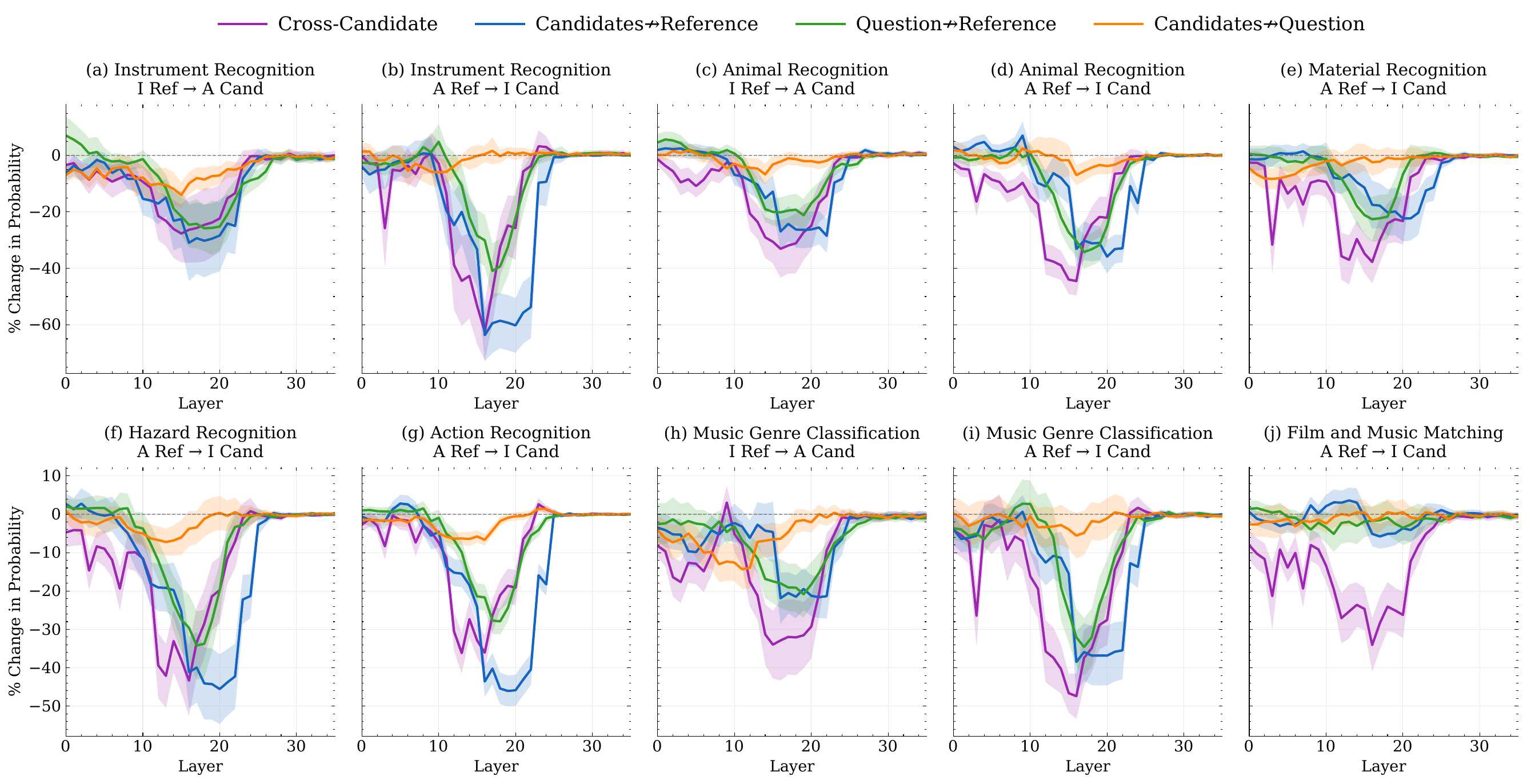}\\[4pt]
  \includegraphics[width=\linewidth]{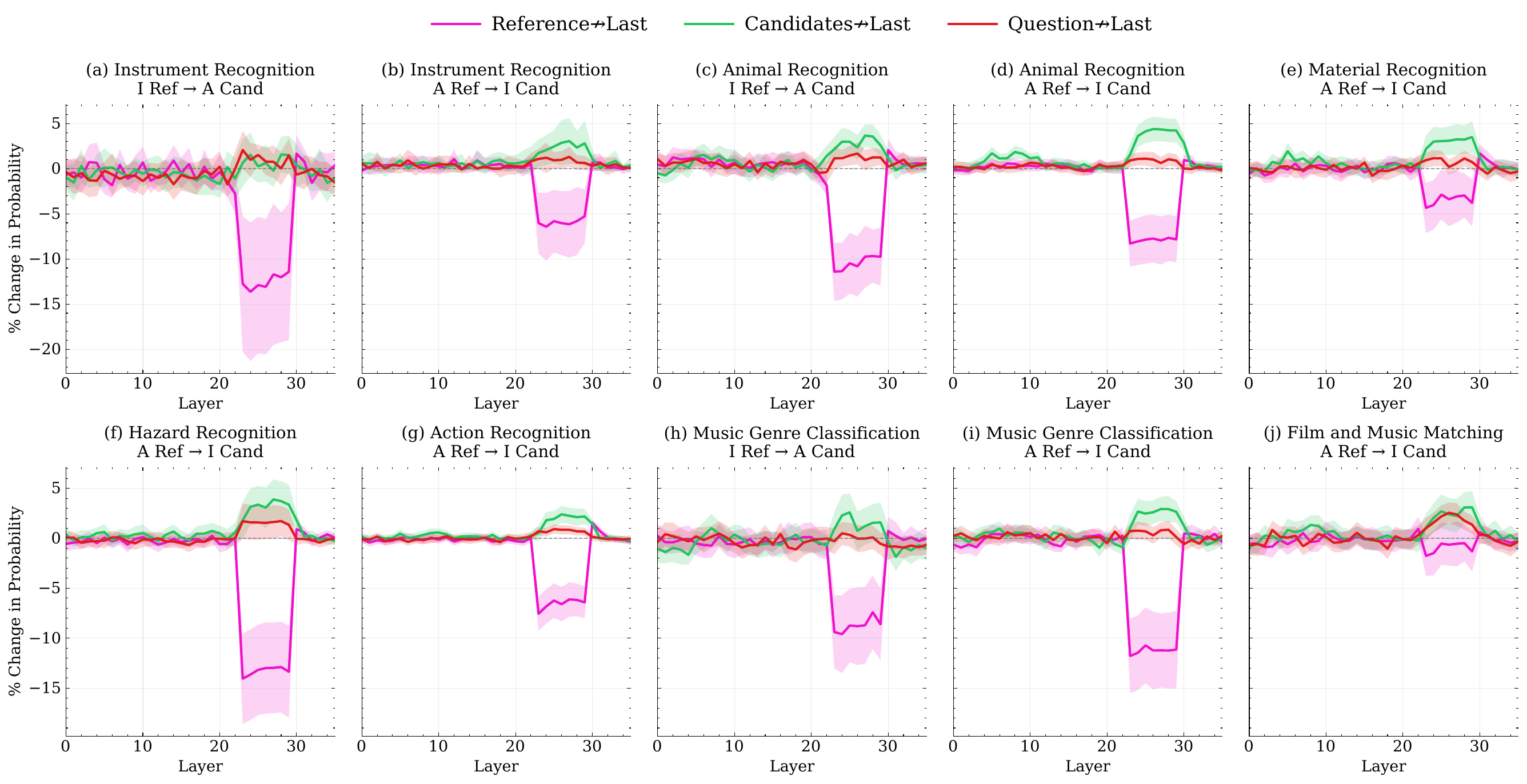}
  \caption{\textbf{Qwen2.5-Omni 3B per-task multi-input knockout (AV-Odyssey).} Pathways into the Reference and Question (Cross-Candidate, Candidates$\not\to$Reference, Question$\not\to$Reference, Candidates$\not\to$Question) and pathways into the last token (Reference$\not\to$Last, Candidates$\not\to$Last, Question$\not\to$Last). Each panel shows one task under one input ordering (I Ref $\to$ A Cand or A Ref $\to$ I Cand). Source$\not\to$Target indicates blocking attention edges from source tokens to target tokens.}
  \label{fig:qwen3B_task_group1}
\end{figure}

\begin{figure}[htbp]
  \centering
  \includegraphics[width=\linewidth]{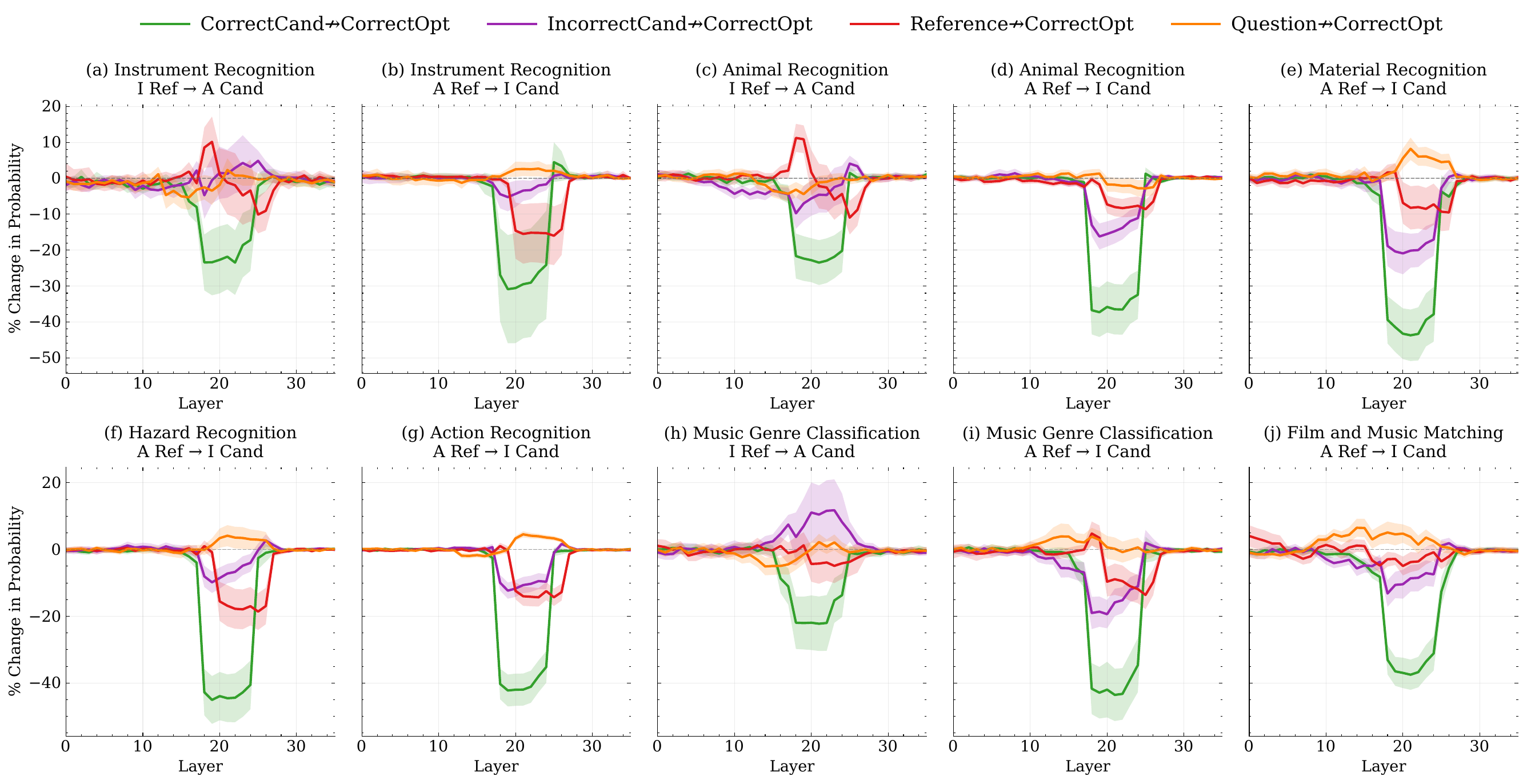}\\[4pt]
  \includegraphics[width=\linewidth]{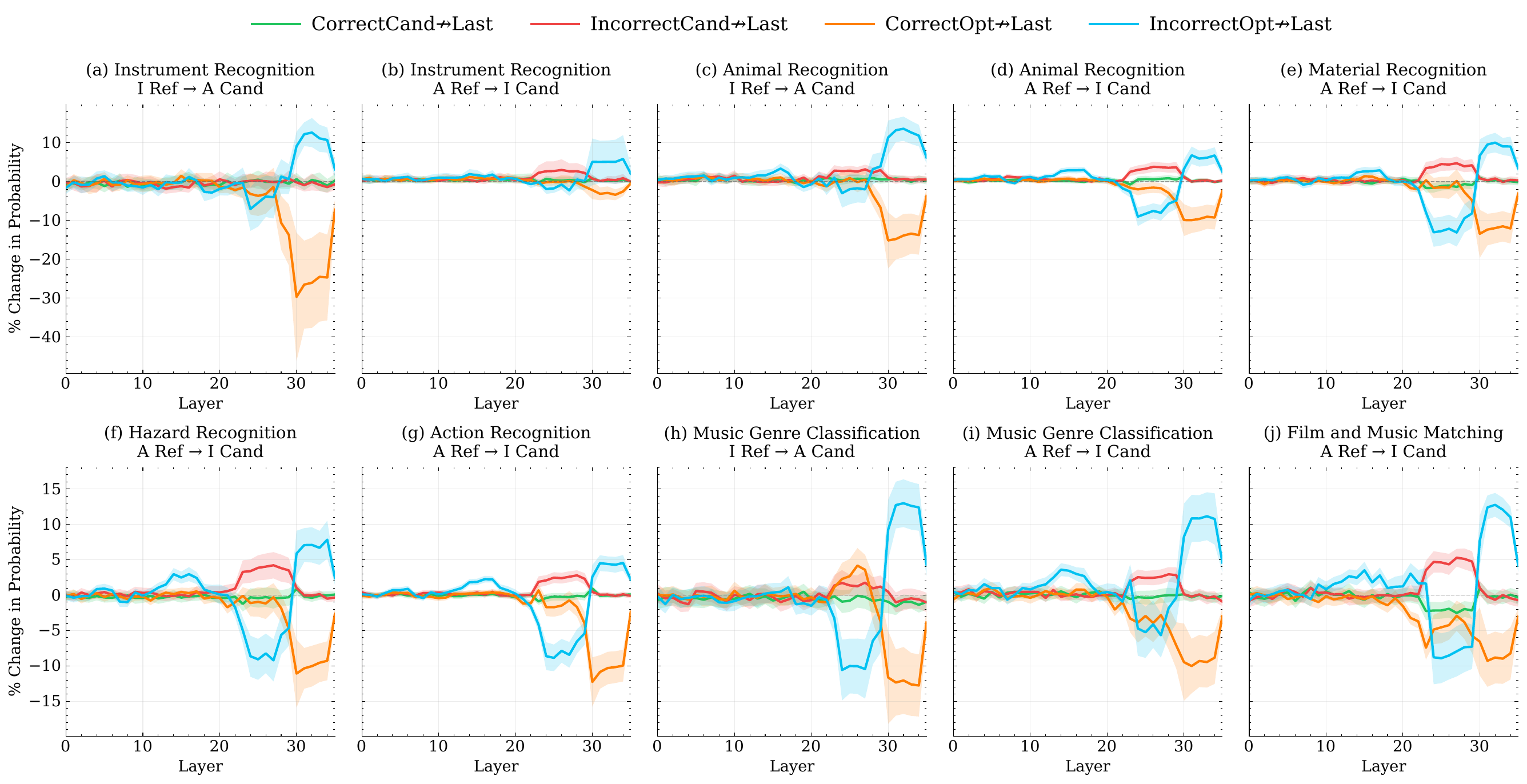}
  \caption{\textbf{Qwen2.5-Omni 3B per-task multi-input knockout (AV-Odyssey).} Pathways into the correct option letter (CorrectCand$\not\to$CorrectOpt, IncorrectCand$\not\to$CorrectOpt, Reference$\not\to$CorrectOpt, Question$\not\to$CorrectOpt) and finer-grained pathways into the last token (CorrectCand$\not\to$Last, IncorrectCand$\not\to$Last, CorrectOpt$\not\to$Last, IncorrectOpt$\not\to$Last). Each panel shows one task under one input ordering (I Ref $\to$ A Cand or A Ref $\to$ I Cand). Source$\not\to$Target indicates blocking attention edges from source tokens to target tokens.}
  \label{fig:qwen3B_task_group2}
\end{figure}

This subsection reports per-task multi-input interleaved knockouts on AV-Odyssey. The main paper reports results averaged across tasks, while Figures~\ref{fig:qwen3B_task_group1} and~\ref{fig:qwen3B_task_group2} break results down by individual task and input ordering (I Ref $\to$ A Cand and A Ref $\to$ I Cand) to show consistency across the benchmark. Figure~\ref{fig:qwen3B_task_group1} covers pathways into the Reference, the Question, and the last token, while Figure~\ref{fig:qwen3B_task_group2} covers pathways into the correct option letter and pathways from the correct and incorrect candidates and option letters into the last token.

\clearpage

\section{Generalization to Qwen2.5-Omni 7B}
\label{appendix:qwen7b}

This appendix reports knockout analyses for Qwen2.5-Omni 7B across all three datasets used in the paper, AV-SpeakerBench~\cite{nguyen2025see}, WorldSense~\cite{hong2025worldsense}, and AV-Odyssey~\cite{gong2024av}. The setup mirrors the analyses on Qwen2.5-Omni 3B in the main paper and Appendix~\ref{appendix:qwen3b-additional}, allowing a direct comparison across model scales. For AV-Odyssey, we report both the averaged knockouts (matching the main paper's reporting style) and the per-task knockouts for completeness.

\subsection{AV-SpeakerBench (audio-visual video)}
\label{appendix:qwen7b-avspeakerbench}

\begin{figure}[htbp]
  \centering
  \includegraphics[width=\linewidth]{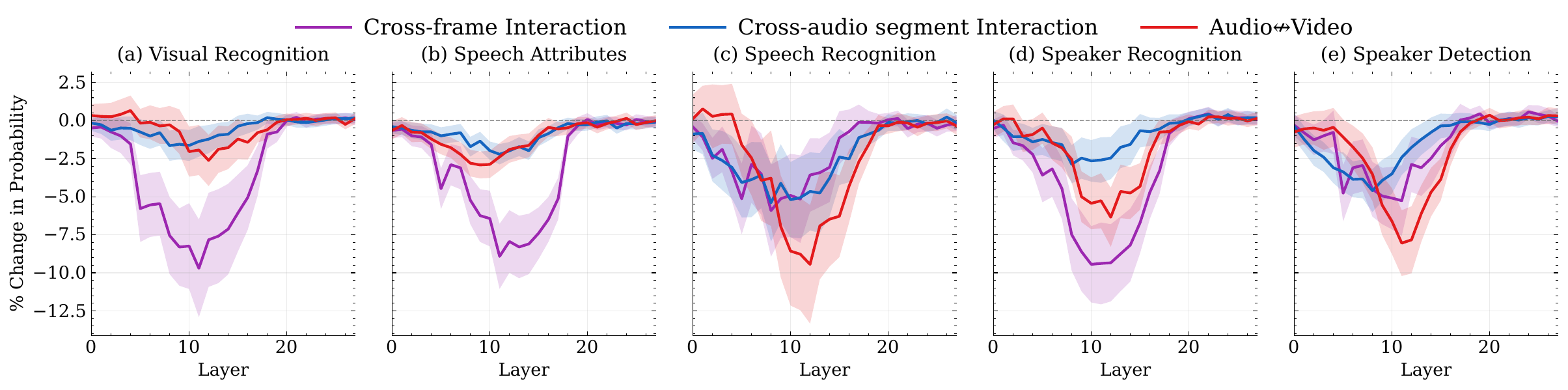}\\[4pt]
  \includegraphics[width=\linewidth]{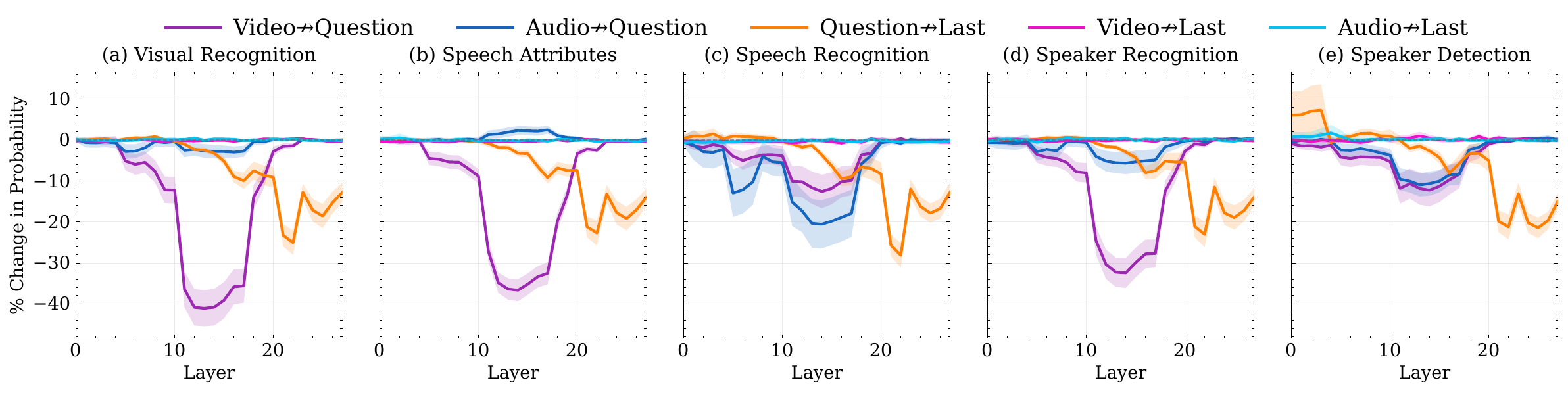}
\caption{\textbf{Qwen2.5-Omni 7B on AV-SpeakerBench.} Knockout of within- and cross-modal pathways (Cross-frame, Cross-audio segment, Audio$\leftrightarrow$Video) and of modality and question pathways into the last token (Video$\not\to$Question, Audio$\not\to$Question, Question$\not\to$Last, Video$\not\to$Last, Audio$\not\to$Last). Source$\not\to$Target indicates blocking attention edges from source tokens to target tokens.}
  \label{fig:qwen7B_video_knock_group1}
\end{figure}

\begin{figure}[htbp]
  \centering
  \includegraphics[width=\linewidth]{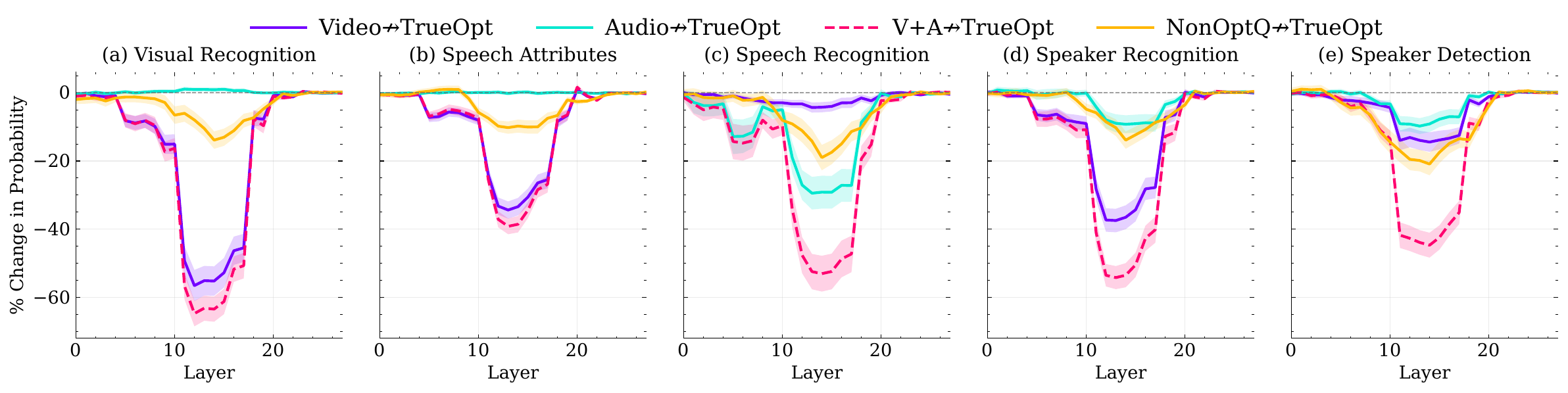}\\[4pt]
  \includegraphics[width=\linewidth]{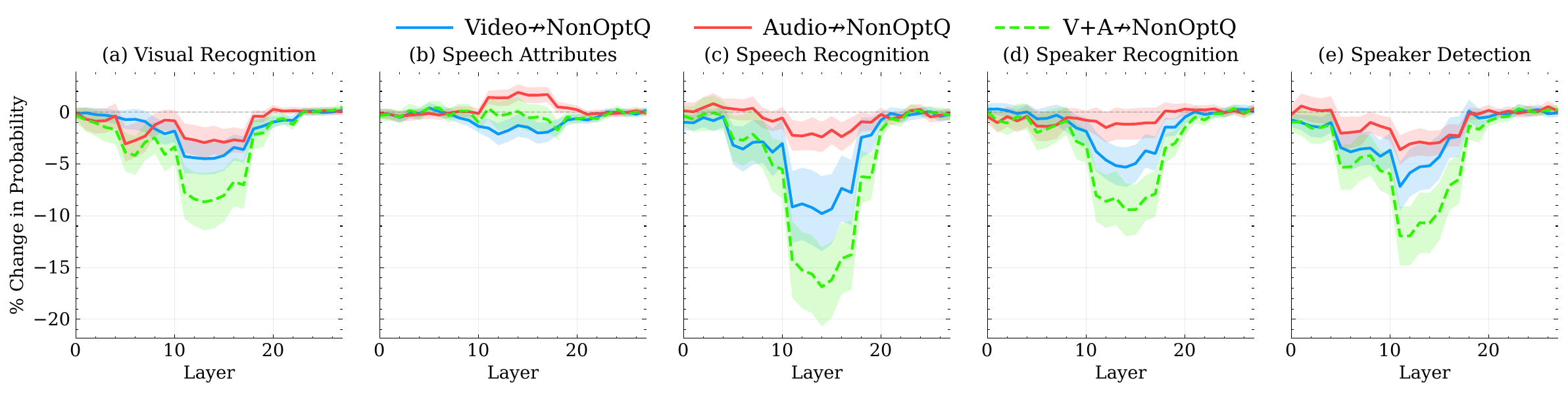}\\[4pt]
  \includegraphics[width=\linewidth]{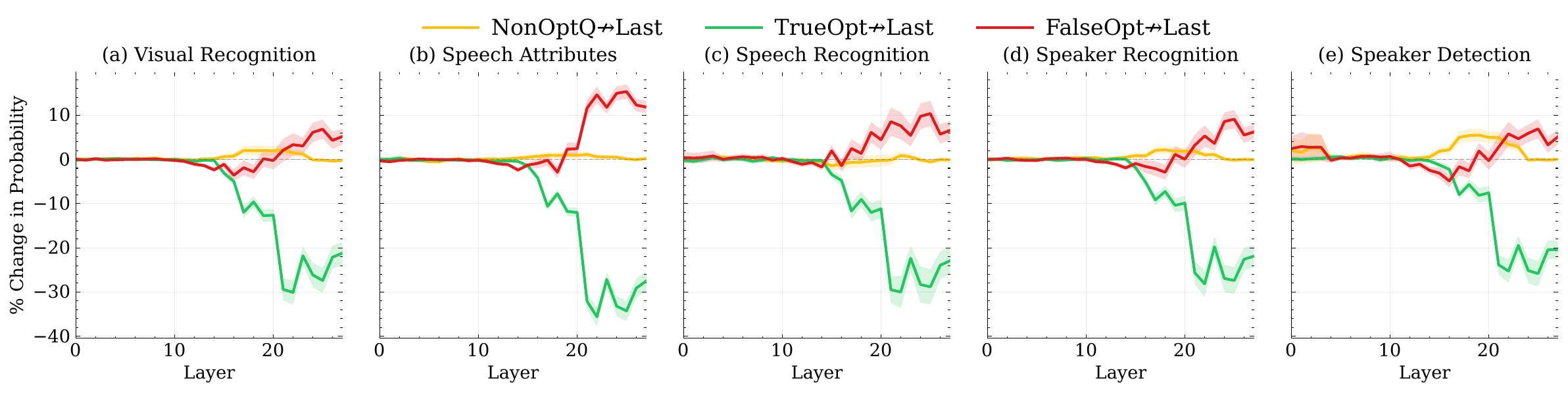}
\caption{\textbf{Qwen2.5-Omni 7B on AV-SpeakerBench.} Modality and question pathways into the correct option letter (Video$\not\to$TrueOpt, Audio$\not\to$TrueOpt, NonOptQ$\not\to$TrueOpt, V+A$\not\to$TrueOpt); modality pathways into the non-option question text (Video$\not\to$NonOptQ, Audio$\not\to$NonOptQ, V+A$\not\to$NonOptQ); and question-internal pathways into the last token (TrueOpt$\not\to$Last, FalseOpt$\not\to$Last, NonOptQ$\not\to$Last). Source$\not\to$Target indicates blocking attention edges from source tokens to target tokens.}
  \label{fig:qwen7B_video_knock_group2}
\end{figure}

Figures~\ref{fig:qwen7B_video_knock_group1} and~\ref{fig:qwen7B_video_knock_group2} report the AV-SpeakerBench knockout results. Figure~\ref{fig:qwen7B_video_knock_group1} covers within- and cross-modal pathways together with modality and question pathways into the last token, while Figure~\ref{fig:qwen7B_video_knock_group2} covers the question-internal pathways and the pathways into the correct option letter and the non-option question text.

\clearpage
\subsection{WorldSense (audio-visual video)}
\label{appendix:qwen7b-worldsense}

\begin{figure}[htbp]
  \centering
  \includegraphics[width=\linewidth]{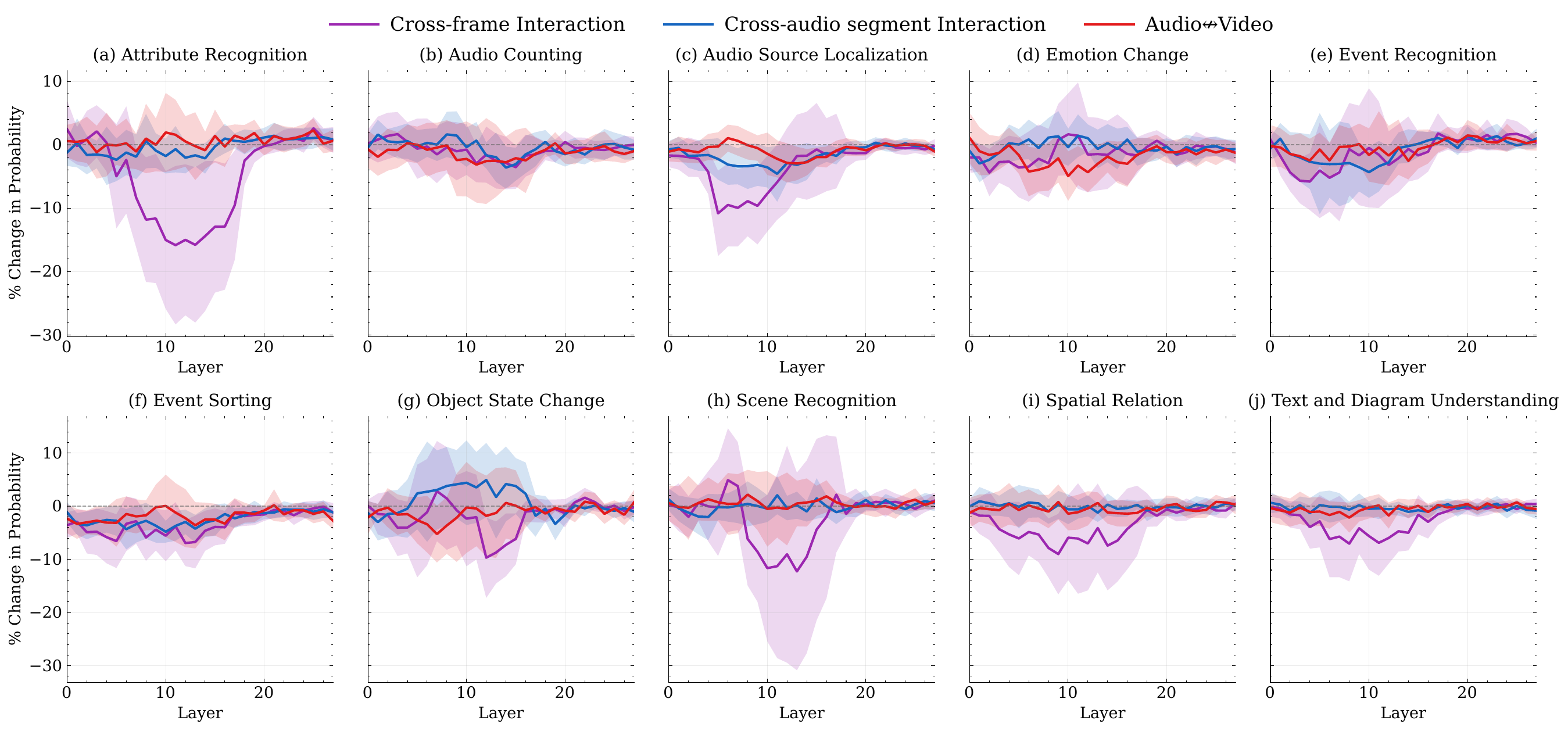}\\[4pt]
  \includegraphics[width=\linewidth]{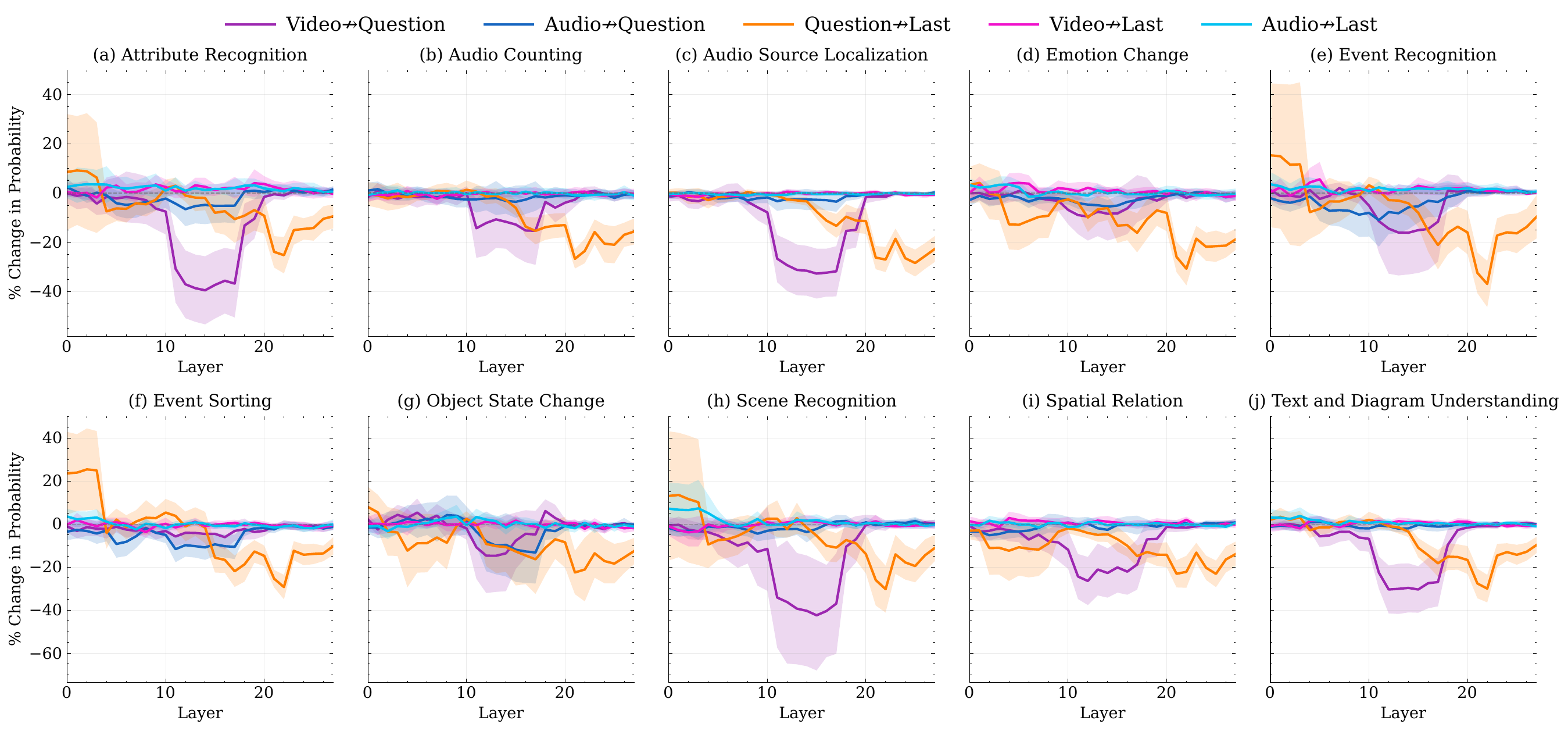}
  \caption{\textbf{Qwen2.5-Omni 7B on WorldSense.} Knockout of within- and cross-modal pathways (Cross-frame, Cross-audio segment, Audio$\leftrightarrow$Video) and of modality and question pathways into the last token (Video$\not\to$Question, Audio$\not\to$Question, Question$\not\to$Last, Video$\not\to$Last, Audio$\not\to$Last). Source$\not\to$Target indicates blocking attention edges from source tokens to target tokens.}
  \label{fig:qwen7B_worldsense_video_knock_group1}
\end{figure}

\begin{figure}[htbp]
  \centering
  \includegraphics[width=\linewidth]{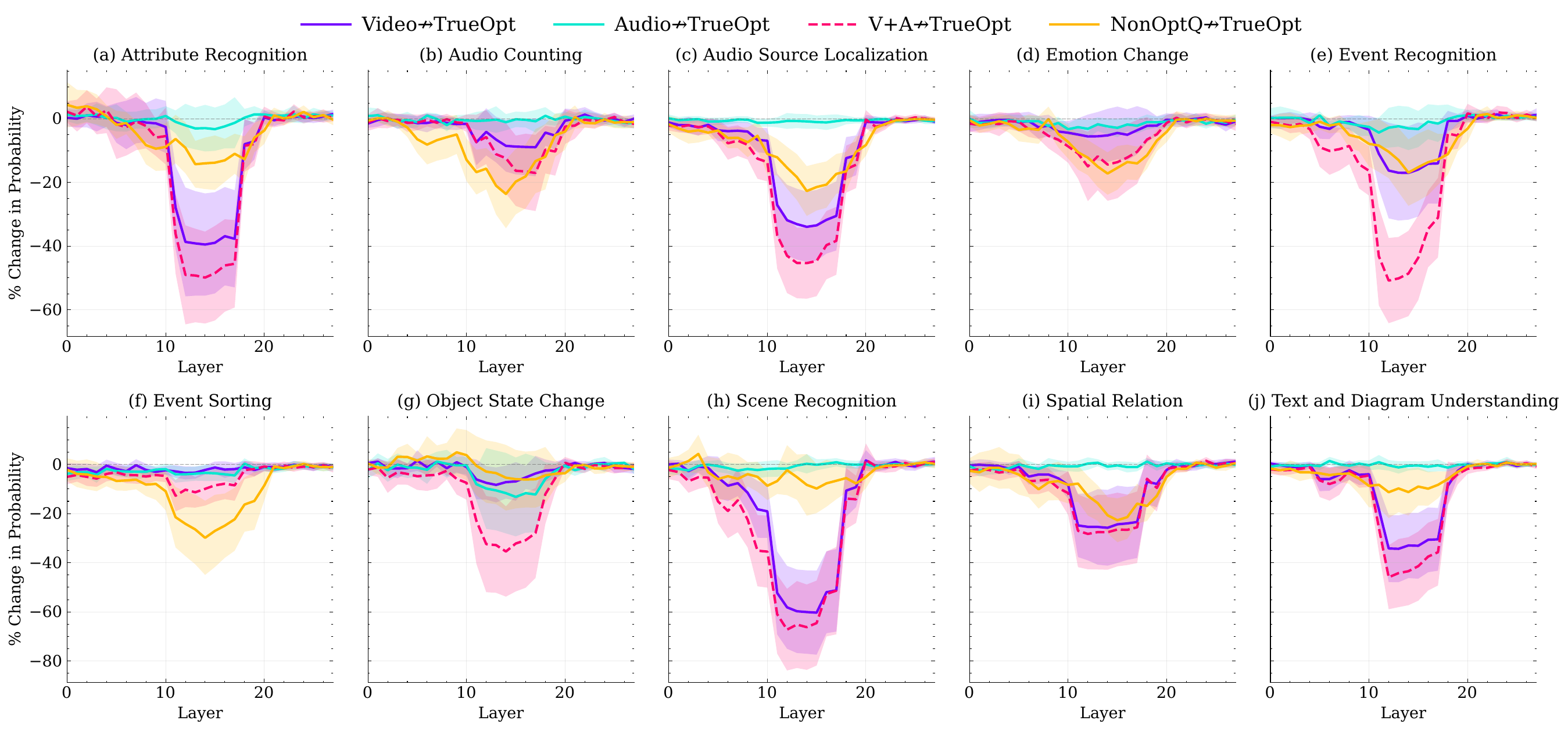}\\[4pt]
  \includegraphics[width=\linewidth]{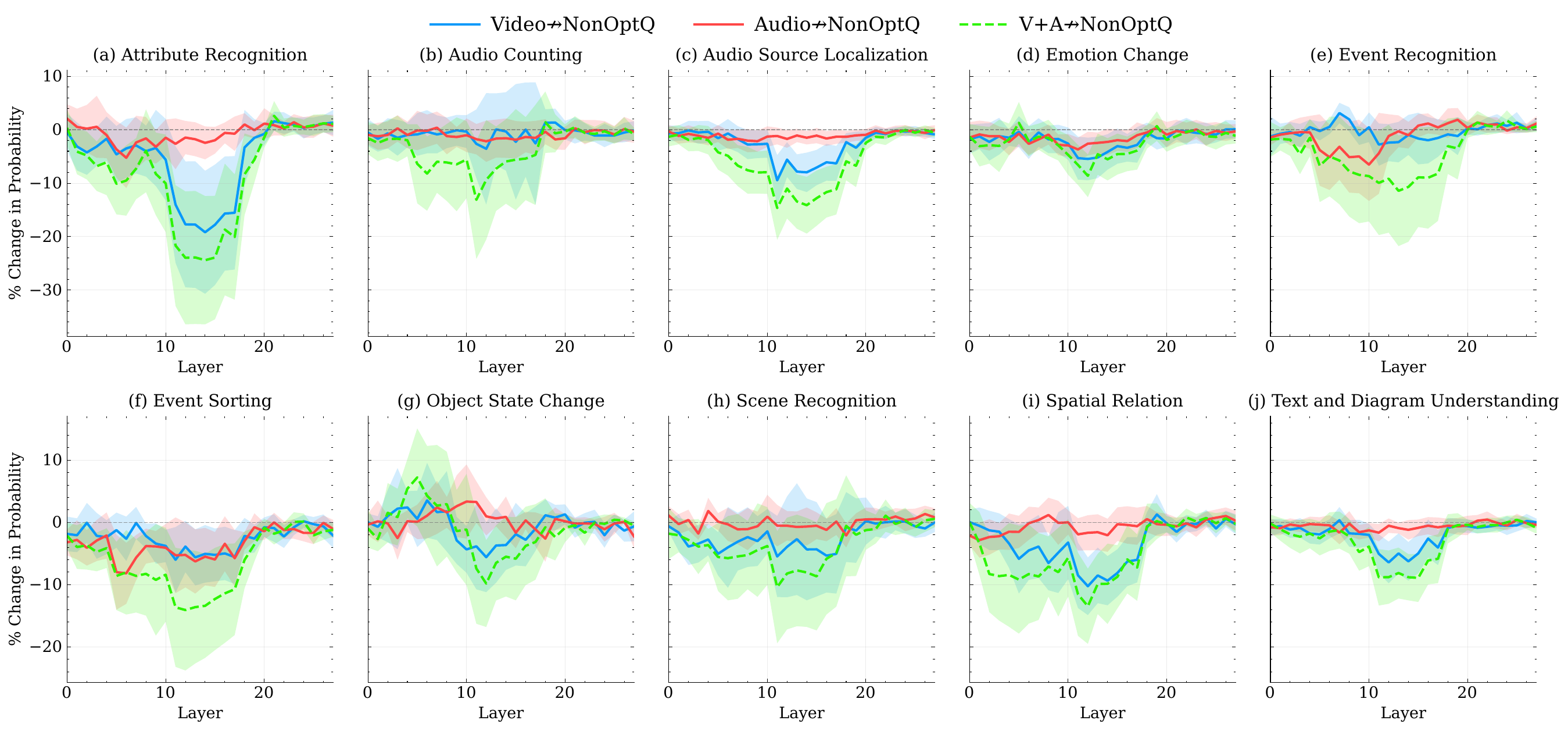}\\[4pt]
  \includegraphics[width=\linewidth]{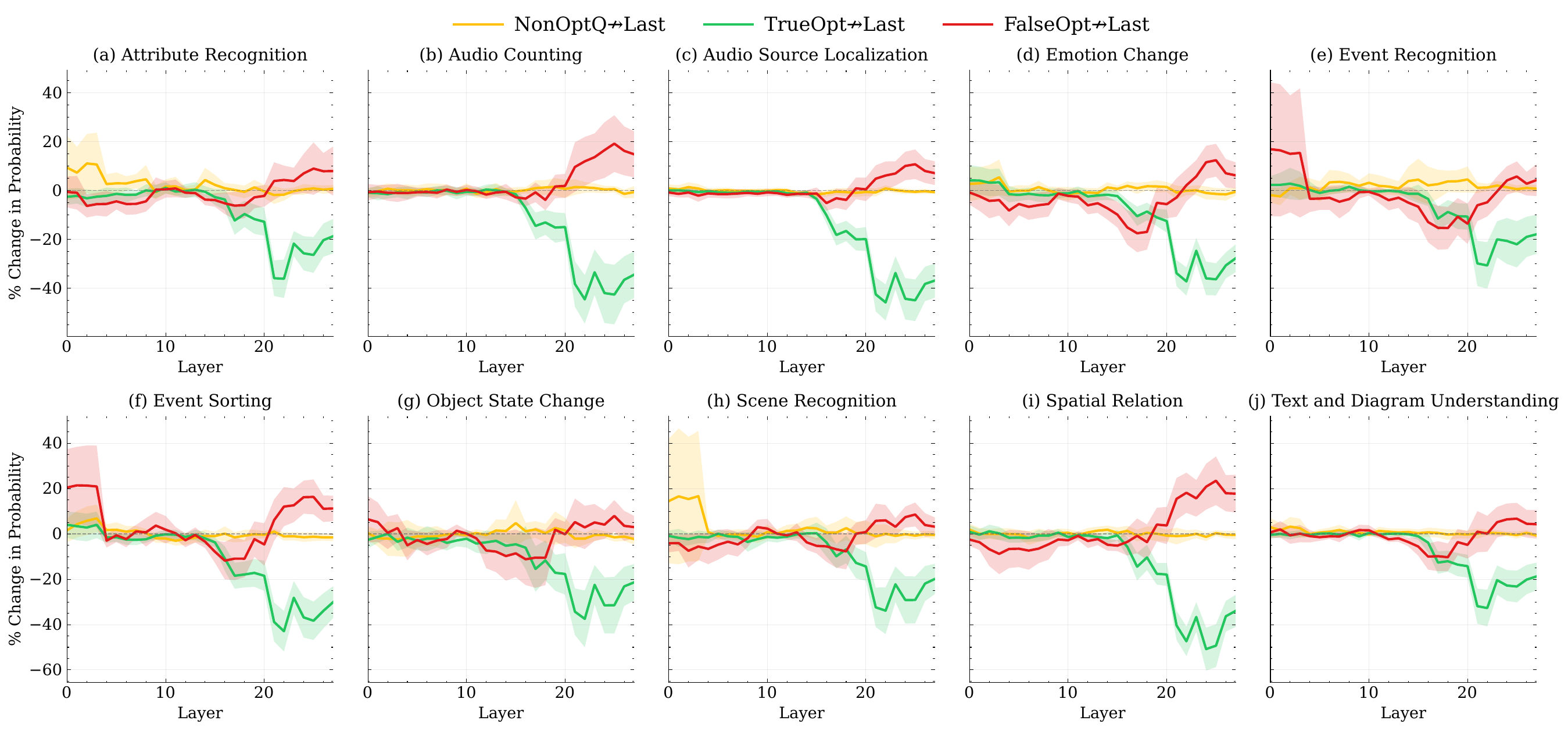}
    \caption{\textbf{Qwen2.5-Omni 7B on WorldSense.} Modality and question pathways into the correct option letter (Video$\not\to$TrueOpt, Audio$\not\to$TrueOpt, NonOptQ$\not\to$TrueOpt, V+A$\not\to$TrueOpt); modality pathways into the non-option question text (Video$\not\to$NonOptQ, Audio$\not\to$NonOptQ, V+A$\not\to$NonOptQ); and question-internal pathways into the last token (TrueOpt$\not\to$Last, FalseOpt$\not\to$Last, NonOptQ$\not\to$Last). Source$\not\to$Target indicates blocking attention edges from source tokens to target tokens.}
  \label{fig:qwen7B_worldsense_video_knock_group2}
\end{figure}

Figures~\ref{fig:qwen7B_worldsense_video_knock_group1} and~\ref{fig:qwen7B_worldsense_video_knock_group2} report the WorldSense knockout results. Figure~\ref{fig:qwen7B_worldsense_video_knock_group1} covers within- and cross-modal pathways together with modality and question pathways into the last token, while Figure~\ref{fig:qwen7B_worldsense_video_knock_group2} covers the question-internal pathways and the pathways into the correct option letter and the non-option question text.

\clearpage
\subsection{AV-Odyssey (multi-input audio-visual interleaved)}
\label{appendix:qwen7b-avodyssey}
This subsection reports multi-input knockouts on AV-Odyssey for Qwen2.5-Omni 7B in two views, averaged across tasks (Appendix~\ref{appendix:qwen7b-avodyssey-avg}) and broken down per task (Appendix~\ref{appendix:qwen7b-avodyssey-pertask}).
\subsubsection{Averaged Across Tasks}
\label{appendix:qwen7b-avodyssey-avg}

\begin{figure}[htbp]
  \centering
  \includegraphics[width=\linewidth]{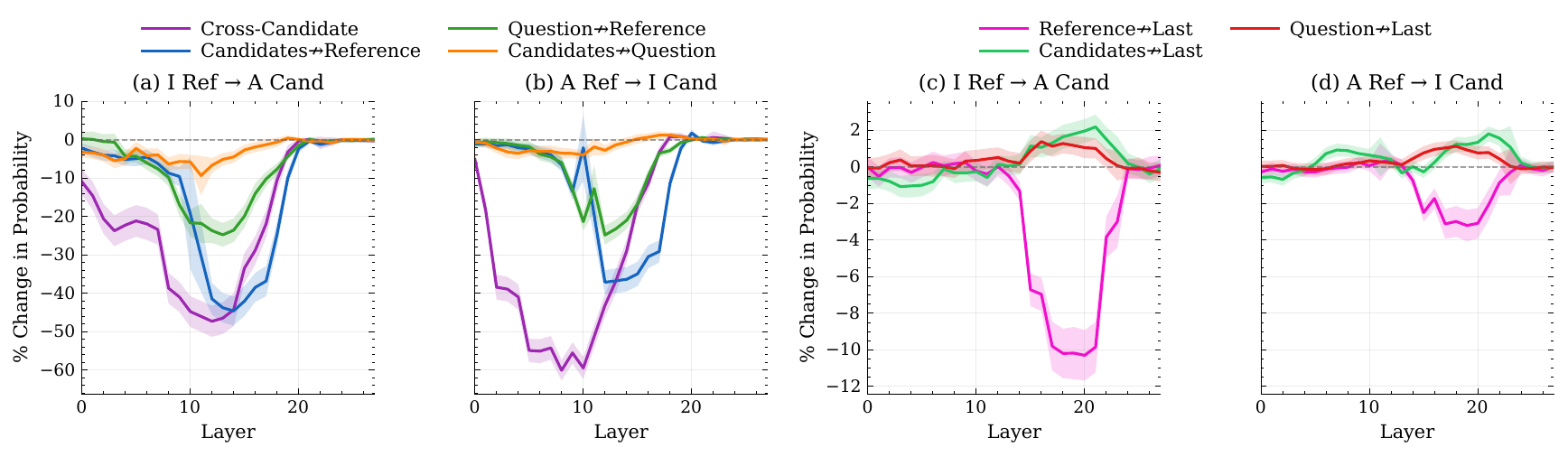}\\[4pt]
  \includegraphics[width=\linewidth]{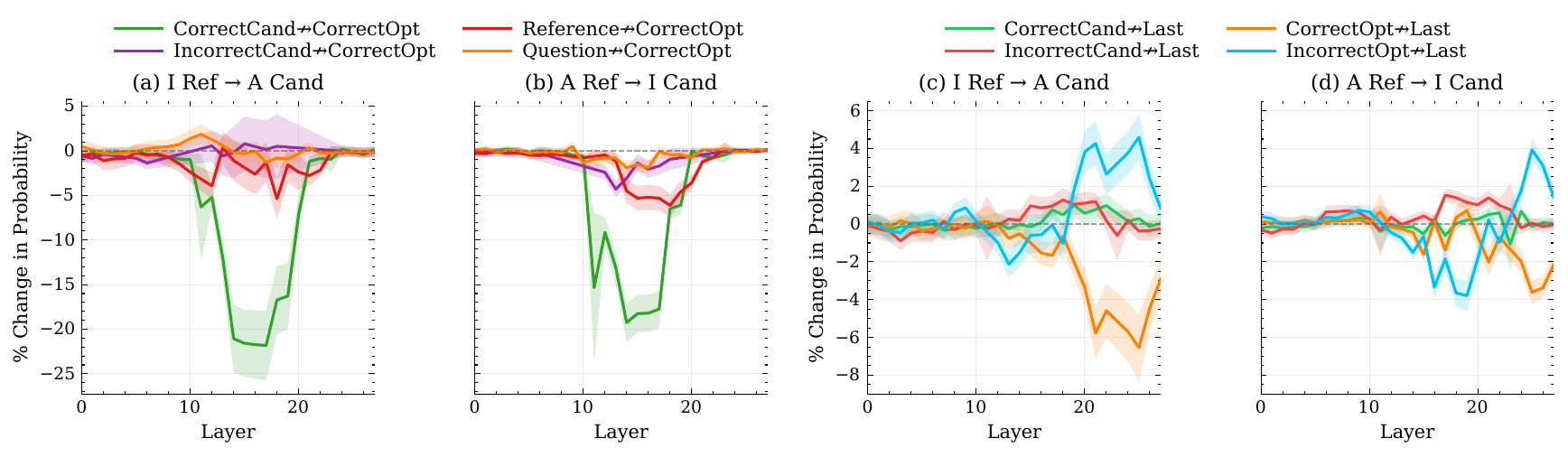}
  \caption{\textbf{Qwen2.5-Omni 7B multi-input knockout (AV-Odyssey), averaged across tasks.} Pathways into the Reference, Question, correct option letter, and last token, mirroring the analyses in the main paper for Qwen2.5-Omni 3B. Source$\not\to$Target indicates blocking attention edges from source tokens to target tokens.}
  \label{fig:qwen7B_overall}
\end{figure}

Figure~\ref{fig:qwen7B_overall} reports the multi-input knockout averaged across tasks, matching the reporting style of the main paper for Qwen2.5-Omni 3B. The same set of pathways into the Reference, Question, correct option letter, and last token are shown.

\subsubsection{Per-task breakdown}
\label{appendix:qwen7b-avodyssey-pertask}

\begin{figure}[htbp]
  \centering
  \includegraphics[width=\linewidth]{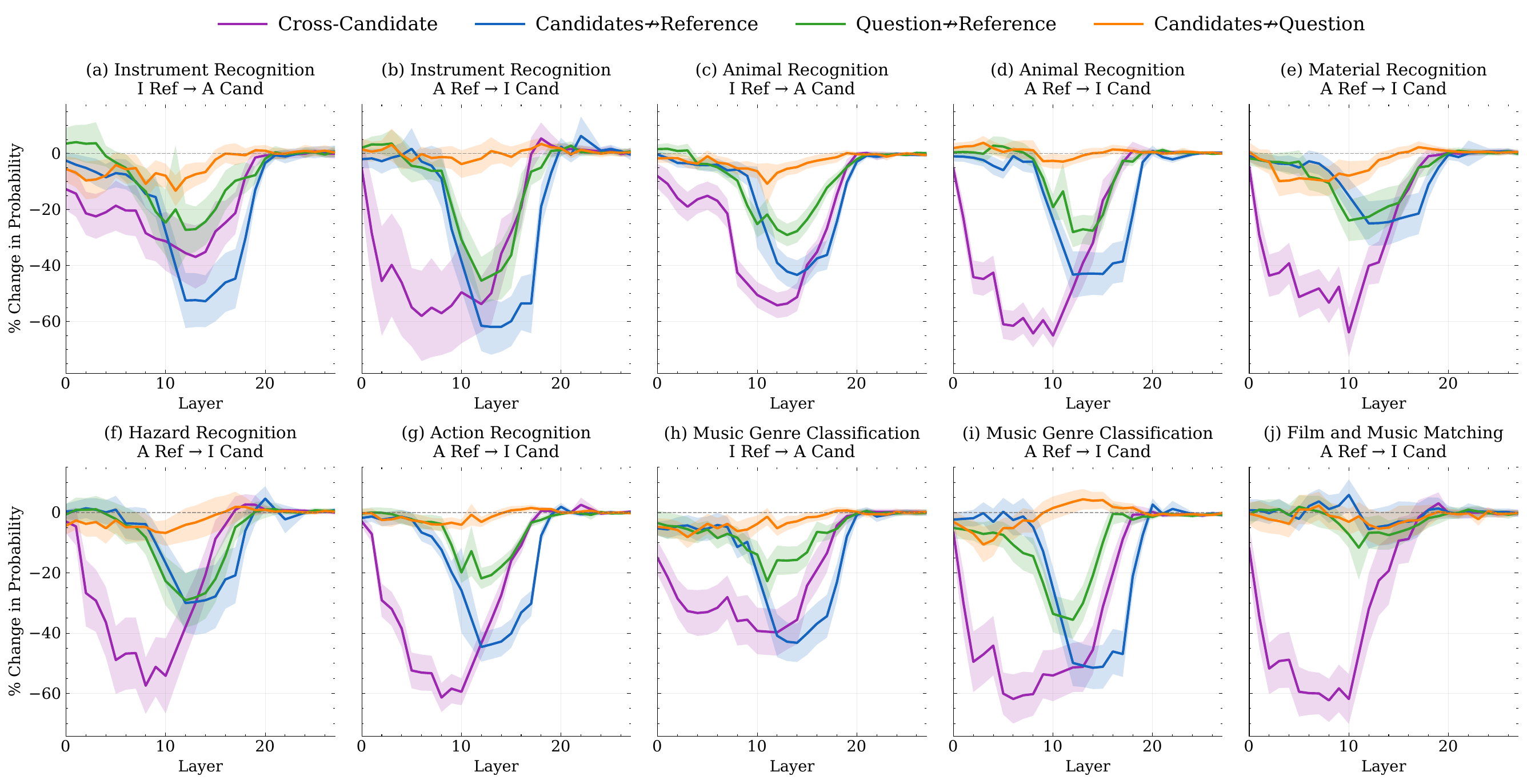}\\[4pt]
  \includegraphics[width=\linewidth]{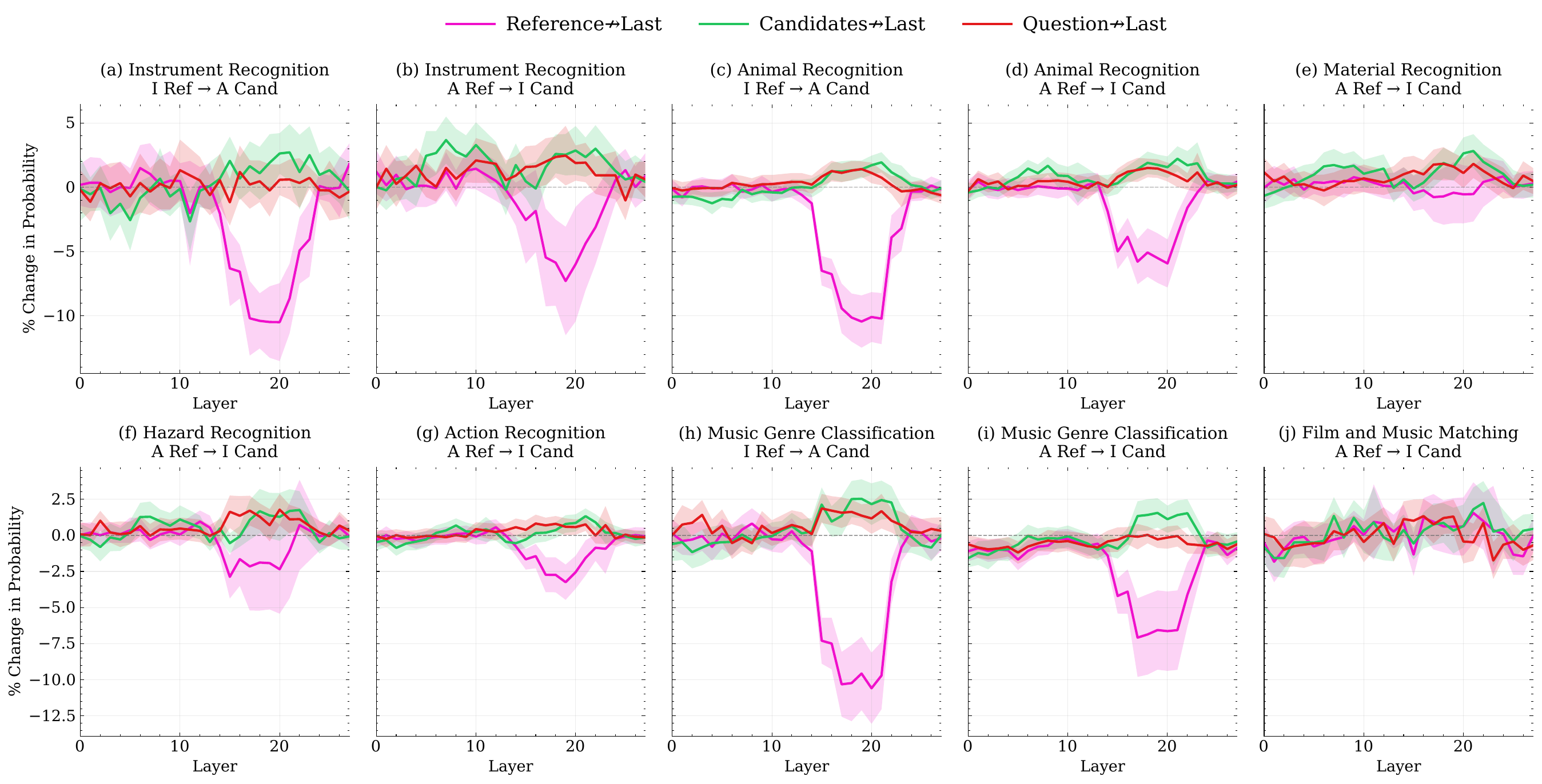}
  \caption{\textbf{Qwen2.5-Omni 7B per-task multi-input knockout (AV-Odyssey).} Pathways into the Reference and Question (Cross-Candidate, Candidates$\not\to$Reference, Question$\not\to$Reference, Candidates$\not\to$Question) and pathways into the last token (Reference$\not\to$Last, Candidates$\not\to$Last, Question$\not\to$Last). Each panel shows one task under one input ordering (I Ref $\to$ A Cand or A Ref $\to$ I Cand). Source$\not\to$Target indicates blocking attention edges from source tokens to target tokens.}
  \label{fig:qwen7B_task1_group1}
\end{figure}

\begin{figure}[htbp]
  \centering
  \includegraphics[width=\linewidth]{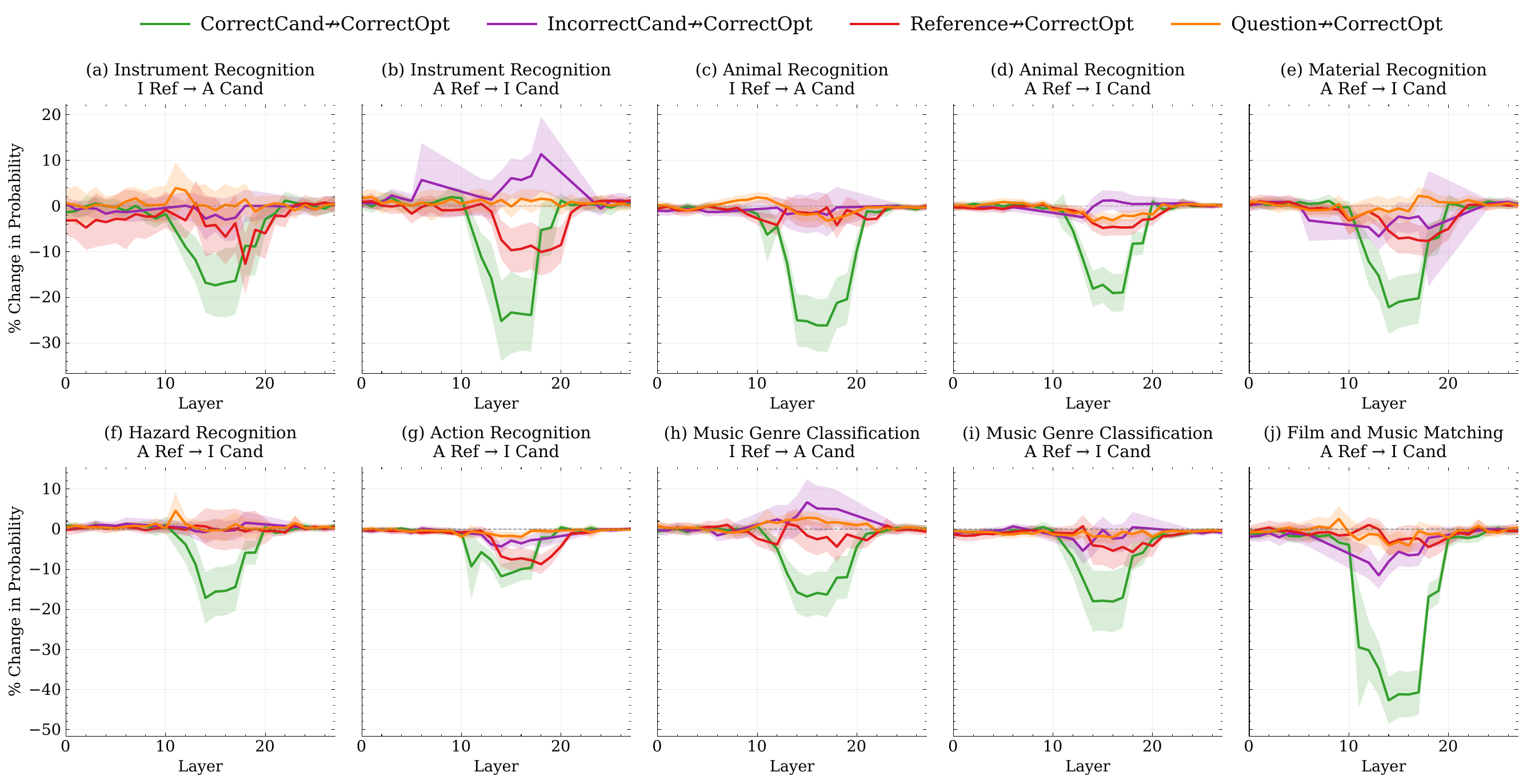}\\[4pt]
  \includegraphics[width=\linewidth]{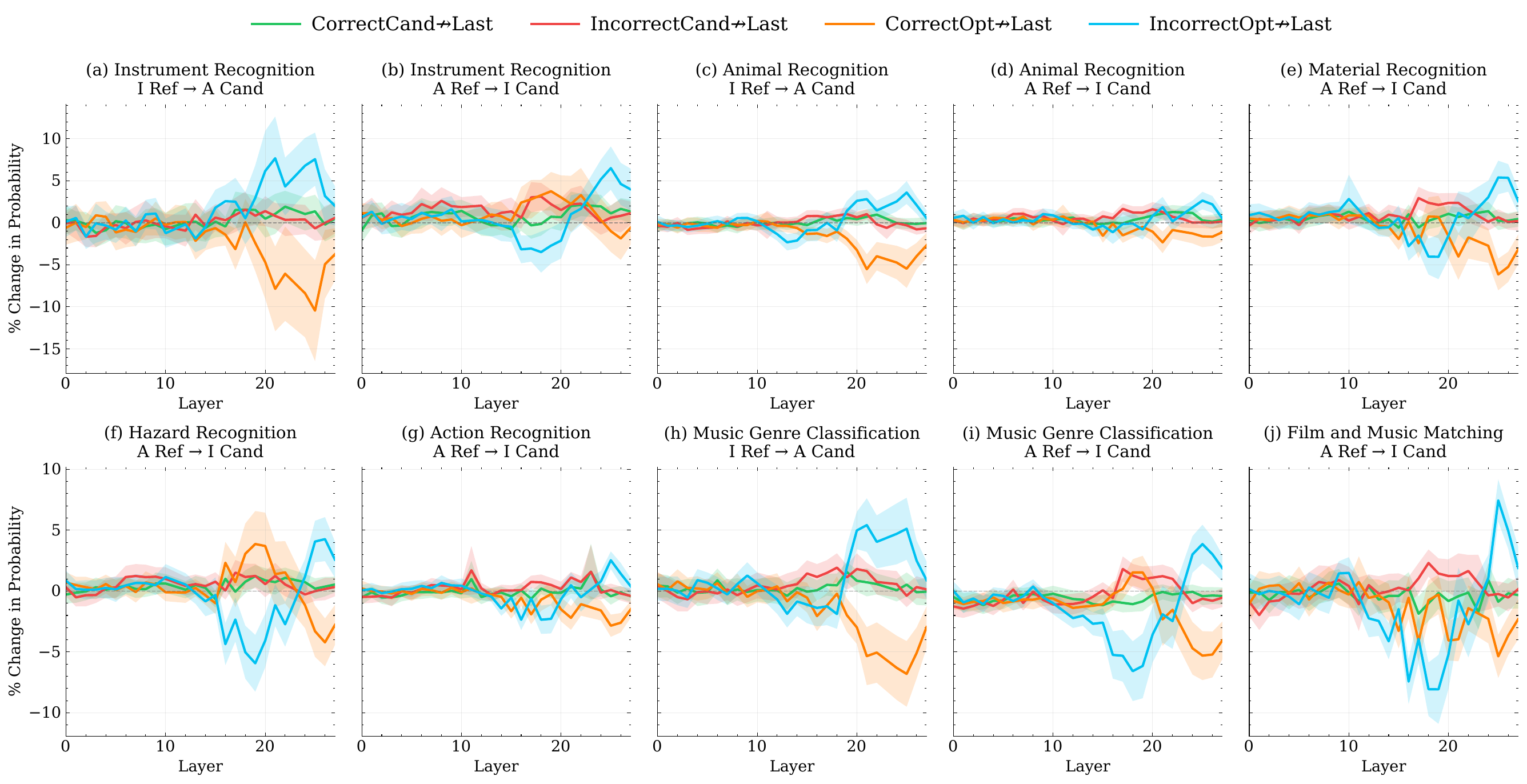}
  \caption{\textbf{Qwen2.5-Omni 7B per-task multi-input knockout (AV-Odyssey).} Pathways into the correct option letter (CorrectCand$\not\to$CorrectOpt, IncorrectCand$\not\to$CorrectOpt, Reference$\not\to$CorrectOpt, Question$\not\to$CorrectOpt) and finer-grained pathways into the last token (CorrectCand$\not\to$Last, IncorrectCand$\not\to$Last, CorrectOpt$\not\to$Last, IncorrectOpt$\not\to$Last). Each panel shows one task under one input ordering (I Ref $\to$ A Cand or A Ref $\to$ I Cand). Source$\not\to$Target indicates blocking attention edges from source tokens to target tokens.}
  \label{fig:qwen7B_task1_group2}
\end{figure}

Figures~\ref{fig:qwen7B_task1_group1} and~\ref{fig:qwen7B_task1_group2} break the same knockouts down by individual task and input ordering (I Ref $\to$ A Cand and A Ref $\to$ I Cand) for completeness. Figure~\ref{fig:qwen7B_task1_group1} covers pathways into the Reference and Question and pathways into the last token, while Figure~\ref{fig:qwen7B_task1_group2} covers pathways into the correct option letter and the finer-grained pathways into the last token.

\clearpage

\section{Generalization to Video-SALMONN2 3B Plus}
\label{appendix:salmonn3b}

This appendix reports knockout analyses for Video-SALMONN2 3B Plus on AV-SpeakerBench~\cite{nguyen2025see} and WorldSense~\cite{hong2025worldsense}, mirroring the analyses on Qwen2.5-Omni 3B in the main paper and Appendix~\ref{appendix:qwen3b-additional}.

\subsection{AV-SpeakerBench (audio-visual video)}
\label{appendix:salmonn3b-avspeakerbench}

\begin{figure}[htbp]
  \centering
  \includegraphics[width=\linewidth]{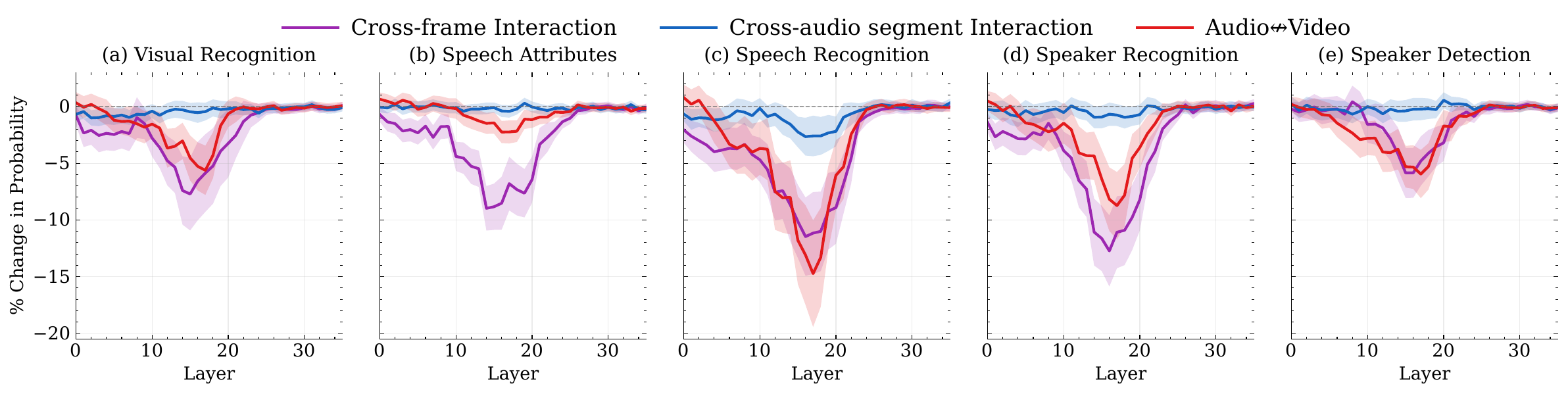}\\[4pt]
  \includegraphics[width=\linewidth]{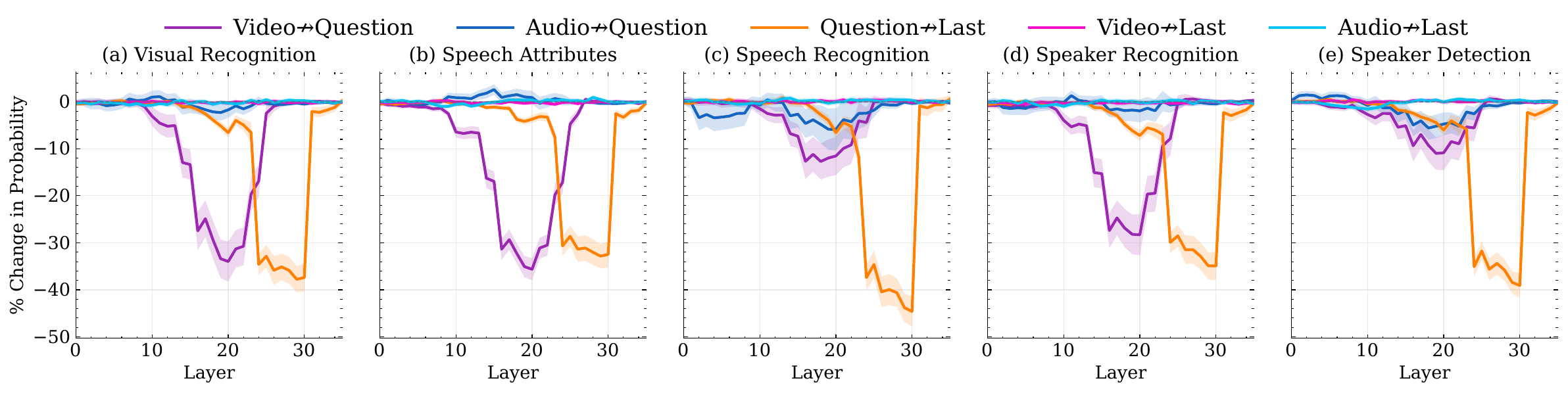}
  \caption{\textbf{Video-SALMONN2 3B Plus on AV-SpeakerBench.} Knockout of within- and cross-modal pathways (Cross-frame, Cross-audio segment, Audio$\leftrightarrow$Video) and of modality and question pathways into the last token (Video$\not\to$Question, Audio$\not\to$Question, Question$\not\to$Last, Video$\not\to$Last, Audio$\not\to$Last). Source$\not\to$Target indicates blocking attention edges from source tokens to target tokens.}
  \label{fig:salmonn3B_video_knock_group1}
\end{figure}

\begin{figure}[htbp]
  \centering
  \includegraphics[width=\linewidth]{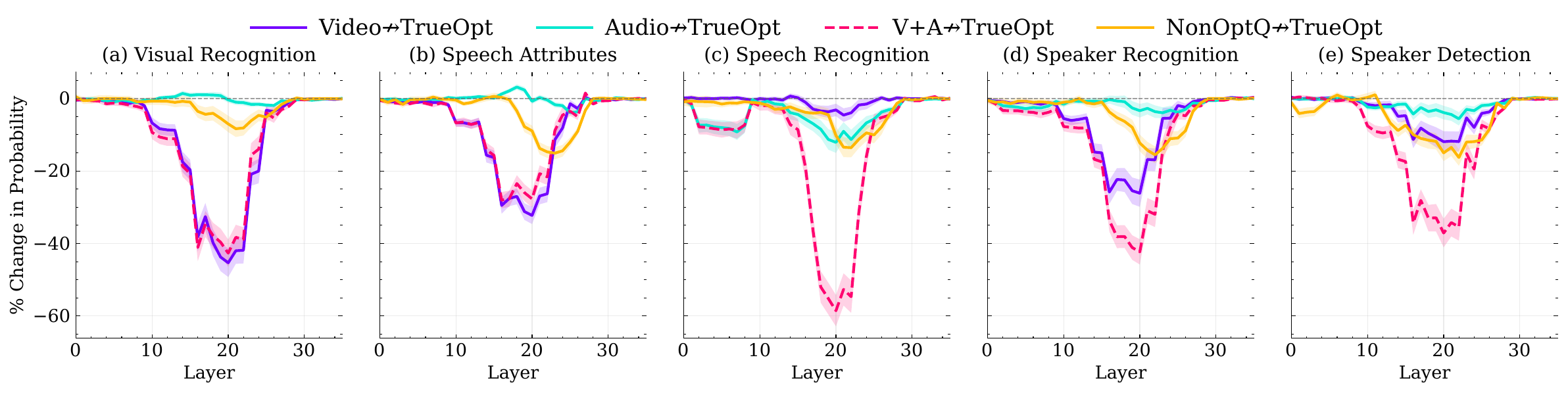}\\[4pt]
  \includegraphics[width=\linewidth]{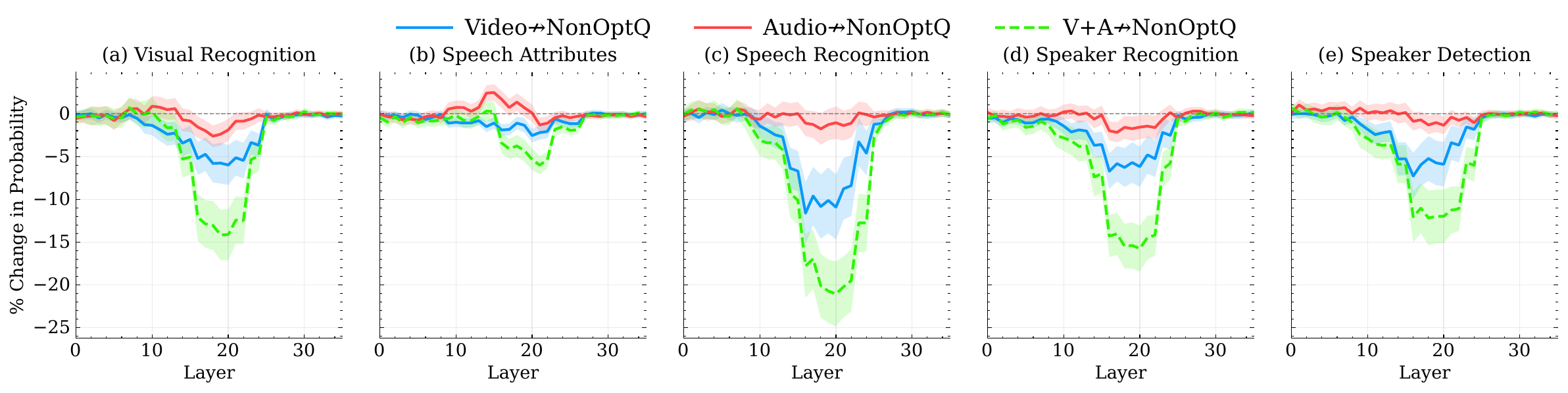}\\[4pt]
  \includegraphics[width=\linewidth]{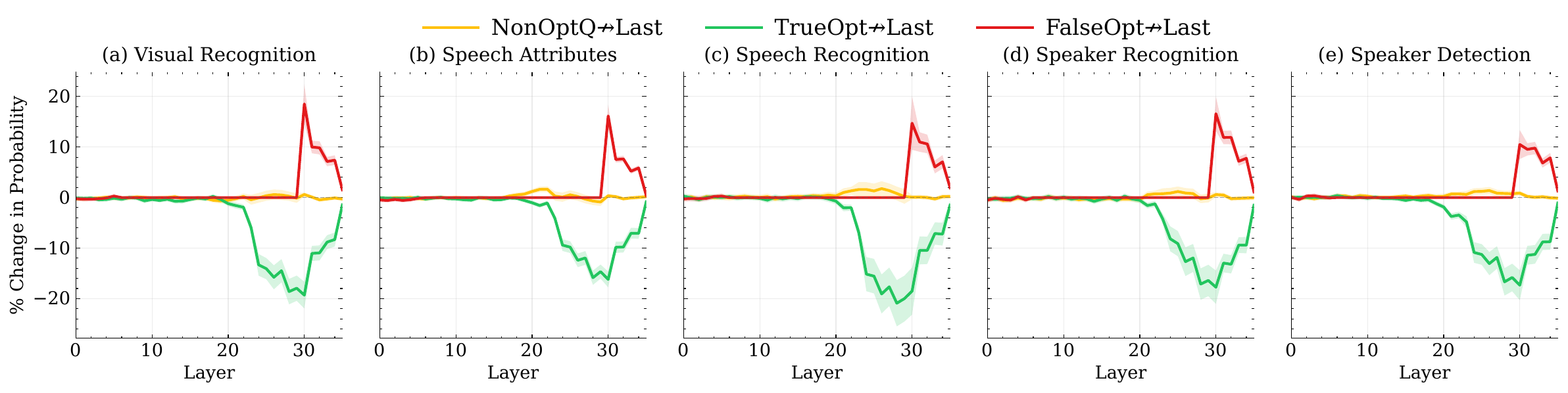}
  \caption{\textbf{Video-SALMONN2 3B Plus on AV-SpeakerBench.} Modality and question pathways into the correct option letter (Video$\not\to$TrueOpt, Audio$\not\to$TrueOpt, NonOptQ$\not\to$TrueOpt, V+A$\not\to$TrueOpt); modality pathways into the non-option question text (Video$\not\to$NonOptQ, Audio$\not\to$NonOptQ, V+A$\not\to$NonOptQ); and question-internal pathways into the last token (TrueOpt$\not\to$Last, FalseOpt$\not\to$Last, NonOptQ$\not\to$Last). Source$\not\to$Target indicates blocking attention edges from source tokens to target tokens.}
  \label{fig:salmonn3B_video_knock_group2}
\end{figure}

Figures~\ref{fig:salmonn3B_video_knock_group1} and~\ref{fig:salmonn3B_video_knock_group2} report the AV-SpeakerBench knockout results. Figure~\ref{fig:salmonn3B_video_knock_group1} covers within- and cross-modal pathways together with modality and question pathways into the last token, while Figure~\ref{fig:salmonn3B_video_knock_group2} covers the question-internal pathways and the pathways into the correct option letter and the non-option question text.

\clearpage
\subsection{WorldSense (audio-visual video)}
\label{appendix:salmonn3b-worldsense}

\begin{figure}[htbp]
  \centering
  \includegraphics[width=\linewidth]{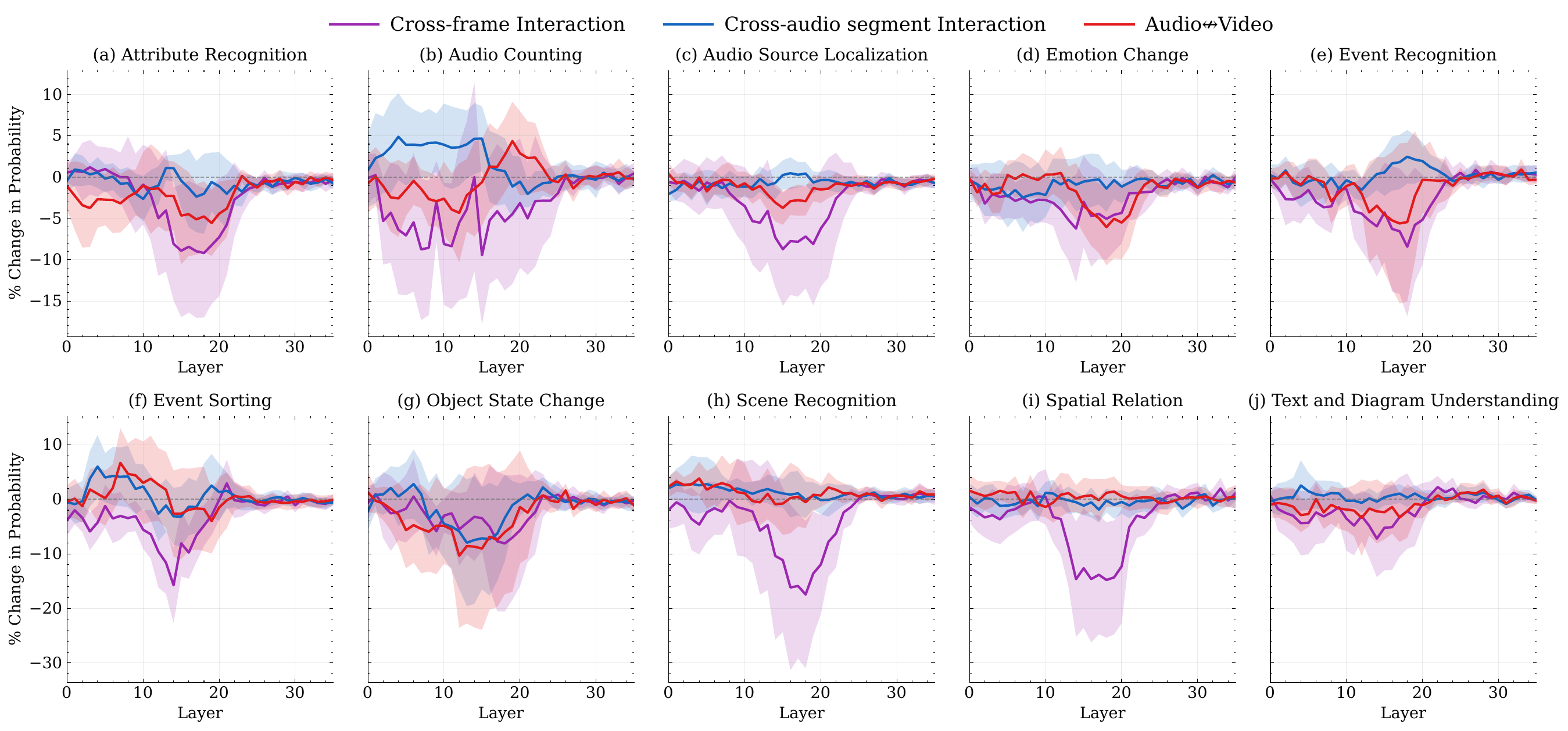}\\[4pt]
  \includegraphics[width=\linewidth]{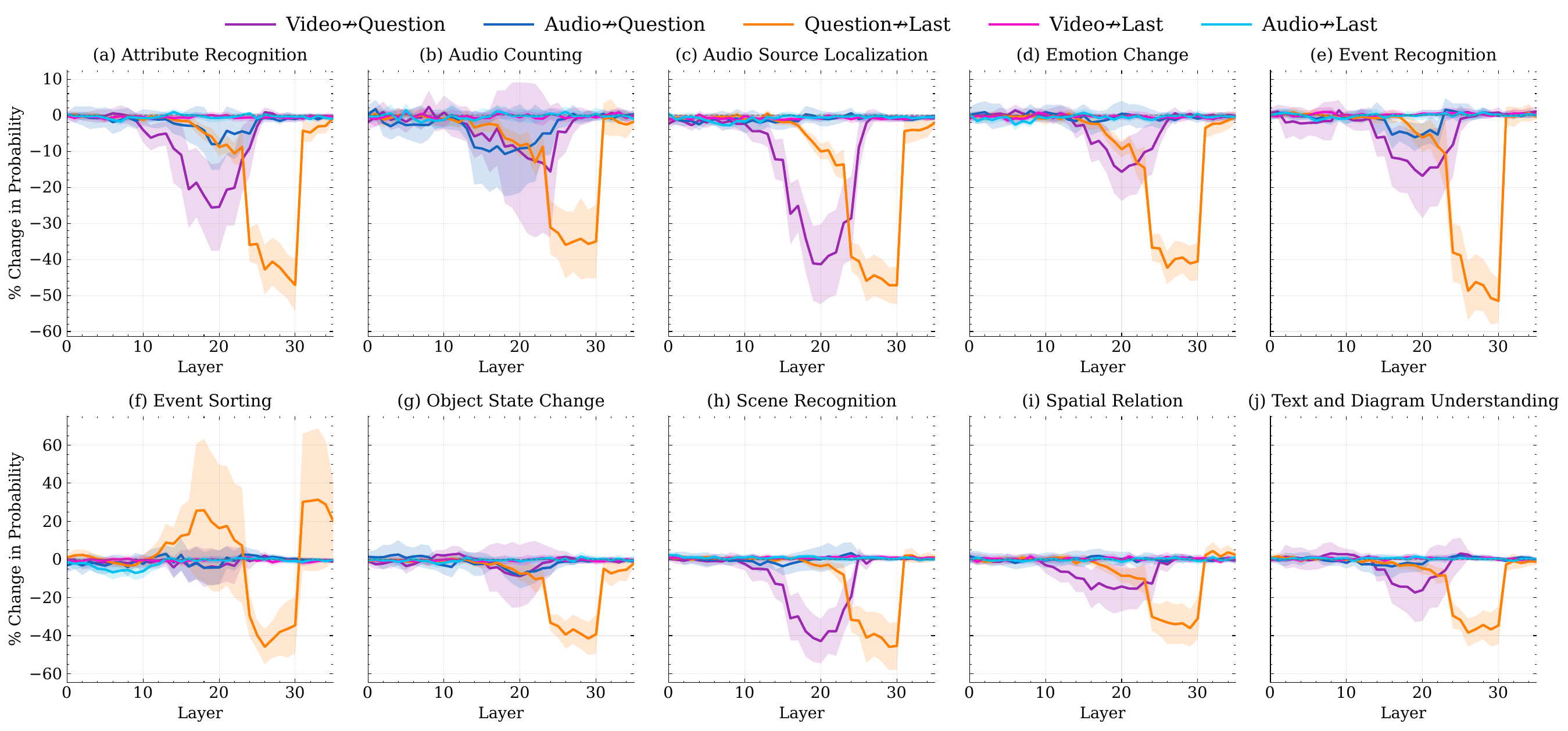}
  \caption{\textbf{Video-SALMONN2 3B Plus on WorldSense.} Knockout of within- and cross-modal pathways (Cross-frame, Cross-audio segment, Audio$\leftrightarrow$Video) and of modality and question pathways into the last token (Video$\not\to$Question, Audio$\not\to$Question, Question$\not\to$Last, Video$\not\to$Last, Audio$\not\to$Last). Source$\not\to$Target indicates blocking attention edges from source tokens to target tokens.}
  \label{fig:salmonn3B_worldsense_video_knock_group1}
\end{figure}

\begin{figure}[htbp]
  \centering
  \includegraphics[width=\linewidth]{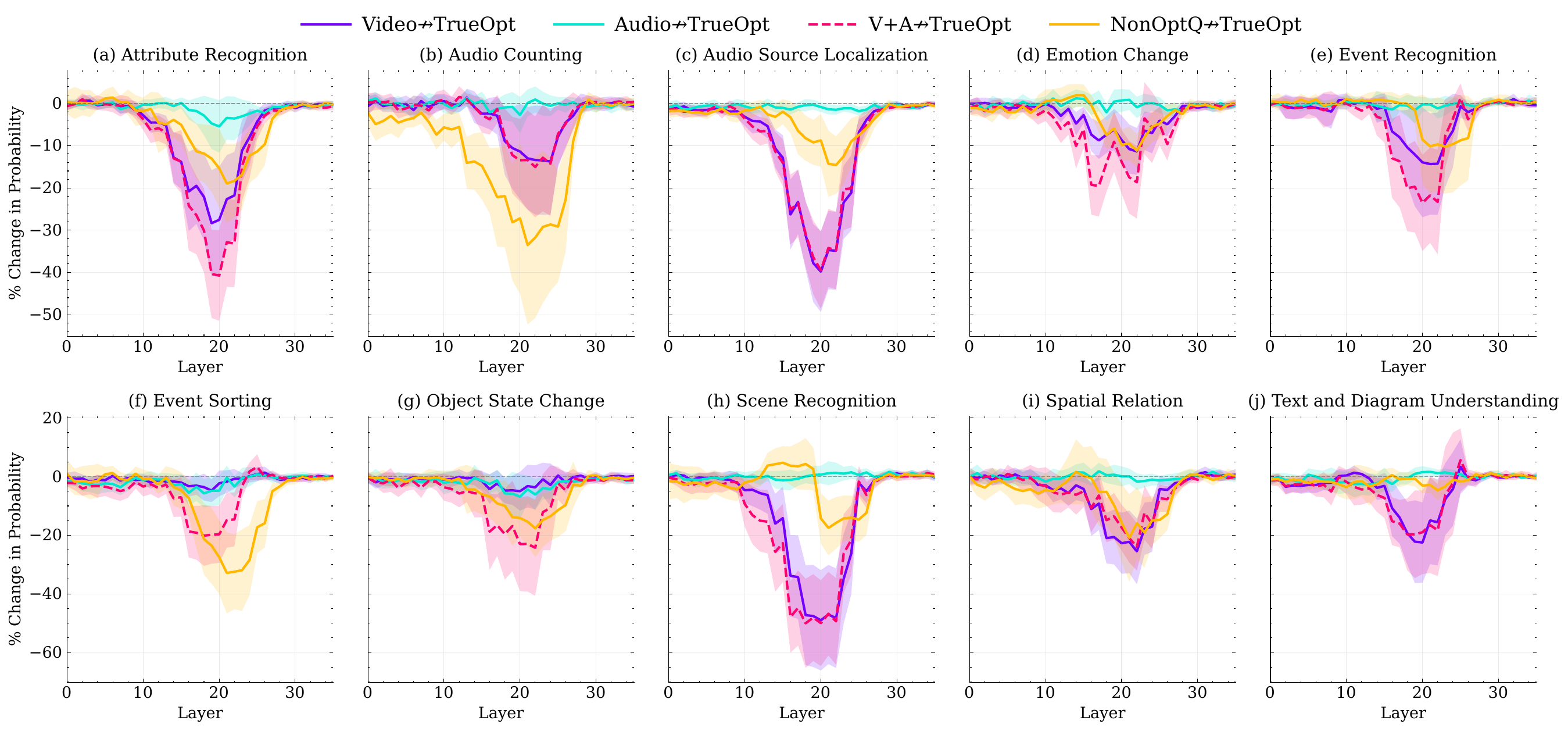}\\[4pt]
  \includegraphics[width=\linewidth]{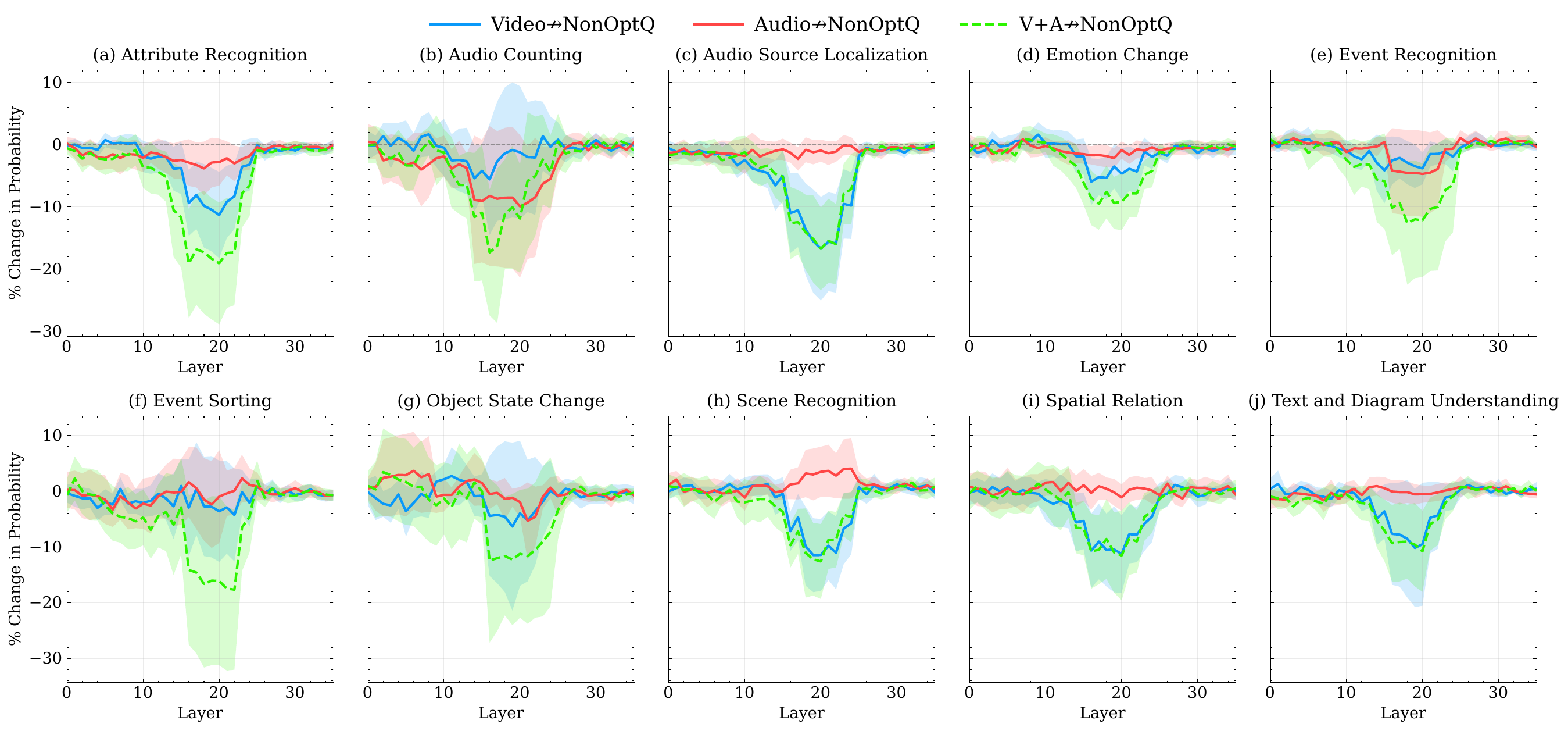}\\[4pt]
  \includegraphics[width=\linewidth]{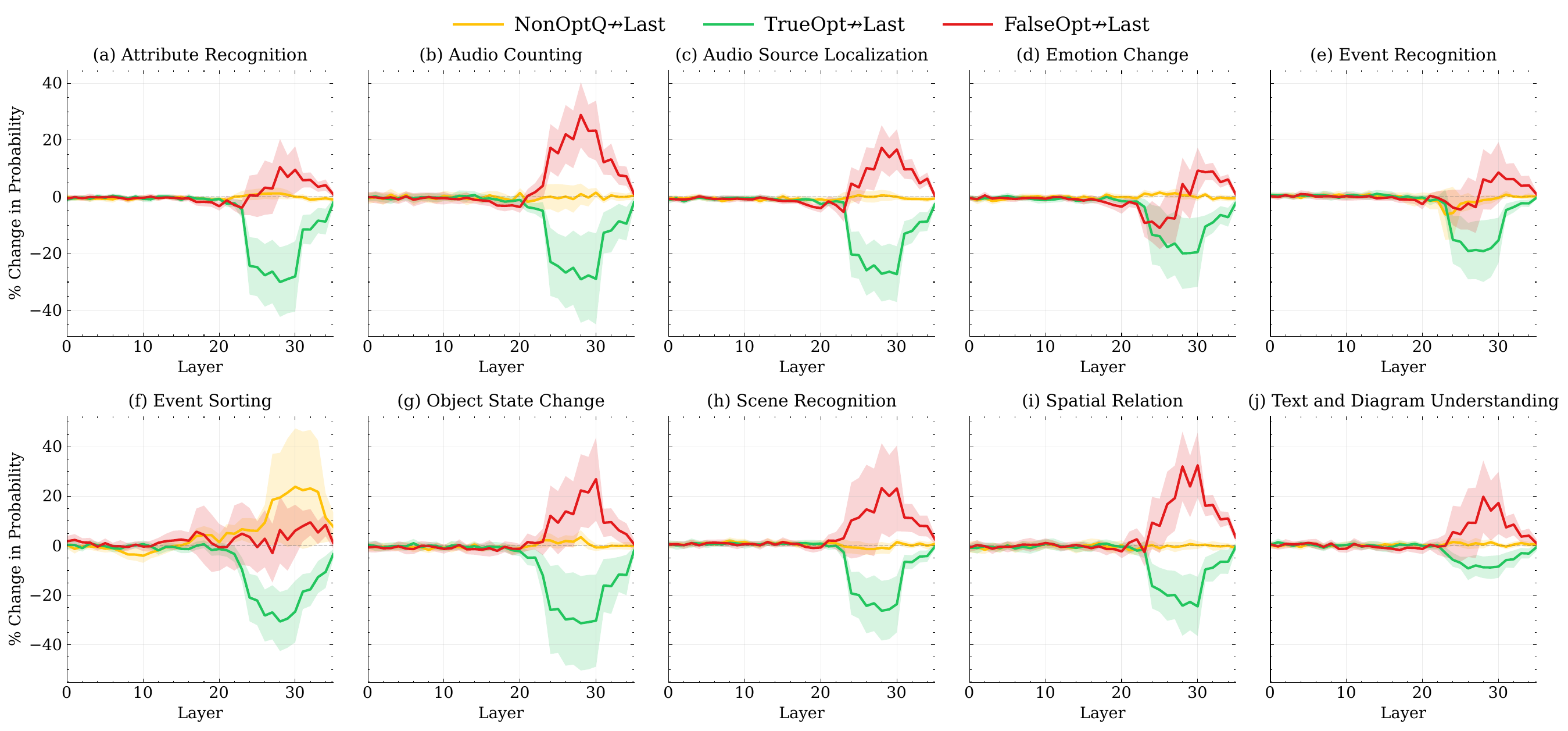}
  \caption{\textbf{Video-SALMONN2 3B Plus on WorldSense.} Modality and question pathways into the correct option letter (Video$\not\to$TrueOpt, Audio$\not\to$TrueOpt, NonOptQ$\not\to$TrueOpt, V+A$\not\to$TrueOpt); modality pathways into the non-option question text (Video$\not\to$NonOptQ, Audio$\not\to$NonOptQ, V+A$\not\to$NonOptQ); and question-internal pathways into the last token (TrueOpt$\not\to$Last, FalseOpt$\not\to$Last, NonOptQ$\not\to$Last). Source$\not\to$Target indicates blocking attention edges from source tokens to target tokens.}
  \label{fig:salmonn3B_worldsense_video_knock_group2}
\end{figure}

Figures~\ref{fig:salmonn3B_worldsense_video_knock_group1} and~\ref{fig:salmonn3B_worldsense_video_knock_group2} report the WorldSense knockout results. Figure~\ref{fig:salmonn3B_worldsense_video_knock_group1} covers within- and cross-modal pathways together with modality and question pathways into the last token, while Figure~\ref{fig:salmonn3B_worldsense_video_knock_group2} covers the question-internal pathways and the pathways into the correct option letter and the non-option question text.

\clearpage

\section{Generalization to Video-SALMONN2 7B Plus}
\label{appendix:salmonn7b}

This appendix reports knockout analyses for Video-SALMONN2 7B Plus on AV-SpeakerBench~\cite{nguyen2025see} and WorldSense~\cite{hong2025worldsense}, mirroring the analyses on Qwen2.5-Omni 3B in the main paper and Appendix~\ref{appendix:qwen3b-additional}.

\subsection{AV-SpeakerBench (audio-visual video)}
\label{appendix:salmonn7b-avspeakerbench}

\begin{figure}[htbp]
  \centering
  \includegraphics[width=\linewidth]{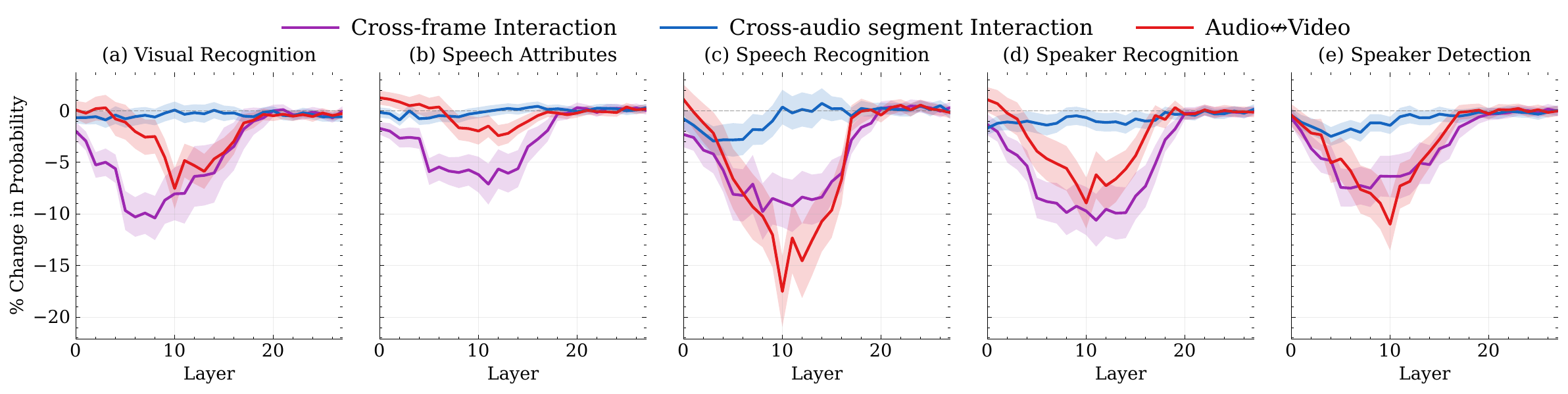}\\[4pt]
  \includegraphics[width=\linewidth]{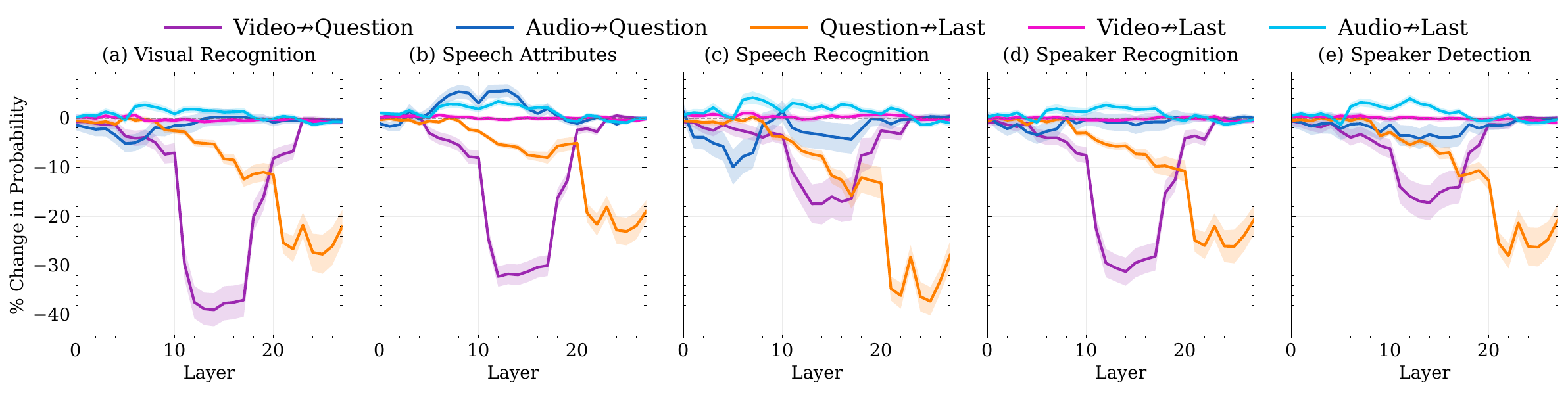}
  \caption{\textbf{Video-SALMONN2 7B Plus on AV-SpeakerBench.} Knockout of within- and cross-modal pathways (Cross-frame, Cross-audio segment, Audio$\leftrightarrow$Video) and of modality and question pathways into the last token (Video$\not\to$Question, Audio$\not\to$Question, Question$\not\to$Last, Video$\not\to$Last, Audio$\not\to$Last). Source$\not\to$Target indicates blocking attention edges from source tokens to target tokens.}
  \label{fig:salmonn7B_video_knock_group1}
\end{figure}

\begin{figure}[htbp]
  \centering
  \includegraphics[width=\linewidth]{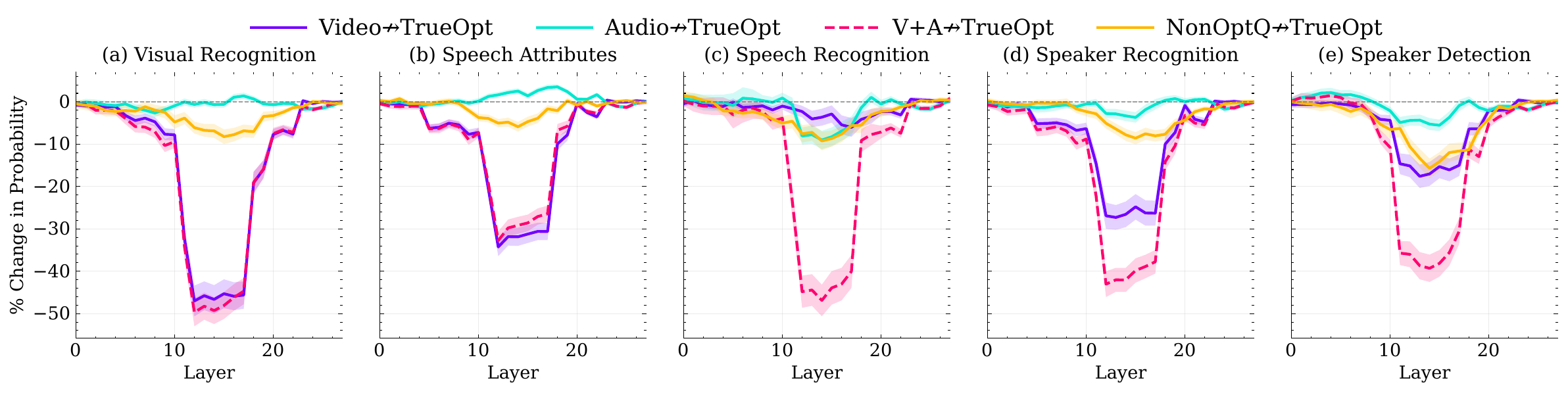}\\[4pt]
  \includegraphics[width=\linewidth]{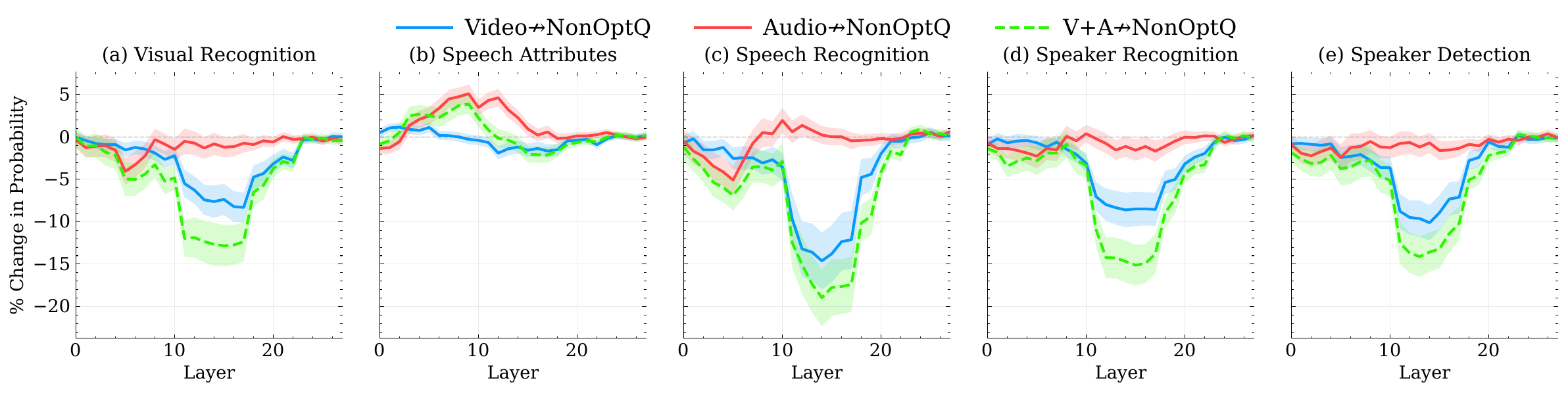}\\[4pt]
  \includegraphics[width=\linewidth]{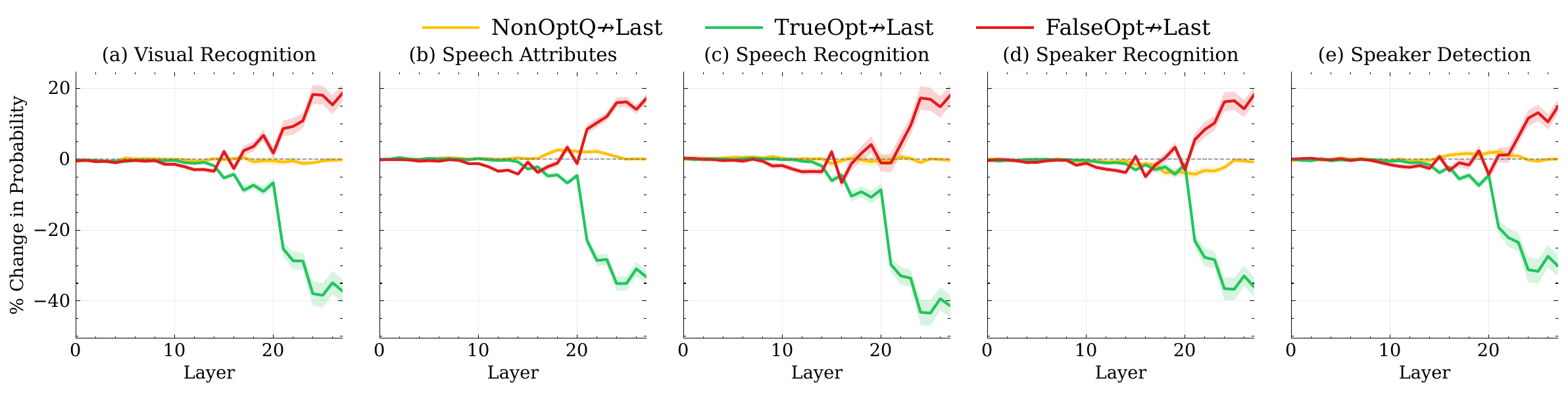}
  \caption{\textbf{Video-SALMONN2 7B Plus on AV-SpeakerBench.} Modality and question pathways into the correct option letter (Video$\not\to$TrueOpt, Audio$\not\to$TrueOpt, NonOptQ$\not\to$TrueOpt, V+A$\not\to$TrueOpt); modality pathways into the non-option question text (Video$\not\to$NonOptQ, Audio$\not\to$NonOptQ, V+A$\not\to$NonOptQ); and question-internal pathways into the last token (TrueOpt$\not\to$Last, FalseOpt$\not\to$Last, NonOptQ$\not\to$Last). Source$\not\to$Target indicates blocking attention edges from source tokens to target tokens.}
  \label{fig:salmonn7B_video_knock_group2}
\end{figure}

Figures~\ref{fig:salmonn7B_video_knock_group1} and~\ref{fig:salmonn7B_video_knock_group2} report the AV-SpeakerBench knockout results. Figure~\ref{fig:salmonn7B_video_knock_group1} covers within- and cross-modal pathways together with modality and question pathways into the last token, while Figure~\ref{fig:salmonn7B_video_knock_group2} covers the question-internal pathways and the pathways into the correct option letter and the non-option question text.

\clearpage
\subsection{WorldSense (audio-visual video)}
\label{appendix:salmonn7b-worldsense}

\begin{figure}[htbp]
  \centering
  \includegraphics[width=\linewidth]{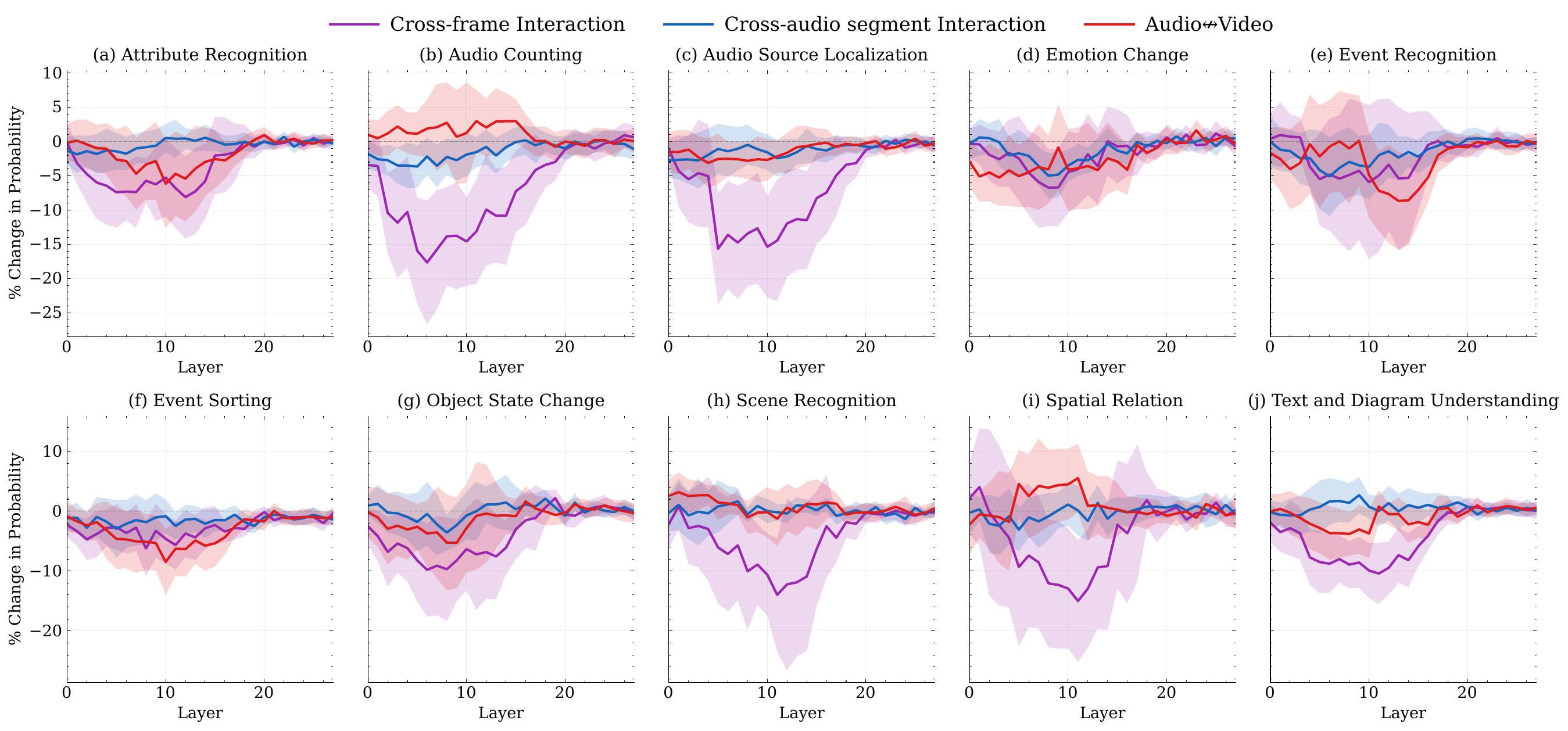}\\[4pt]
  \includegraphics[width=\linewidth]{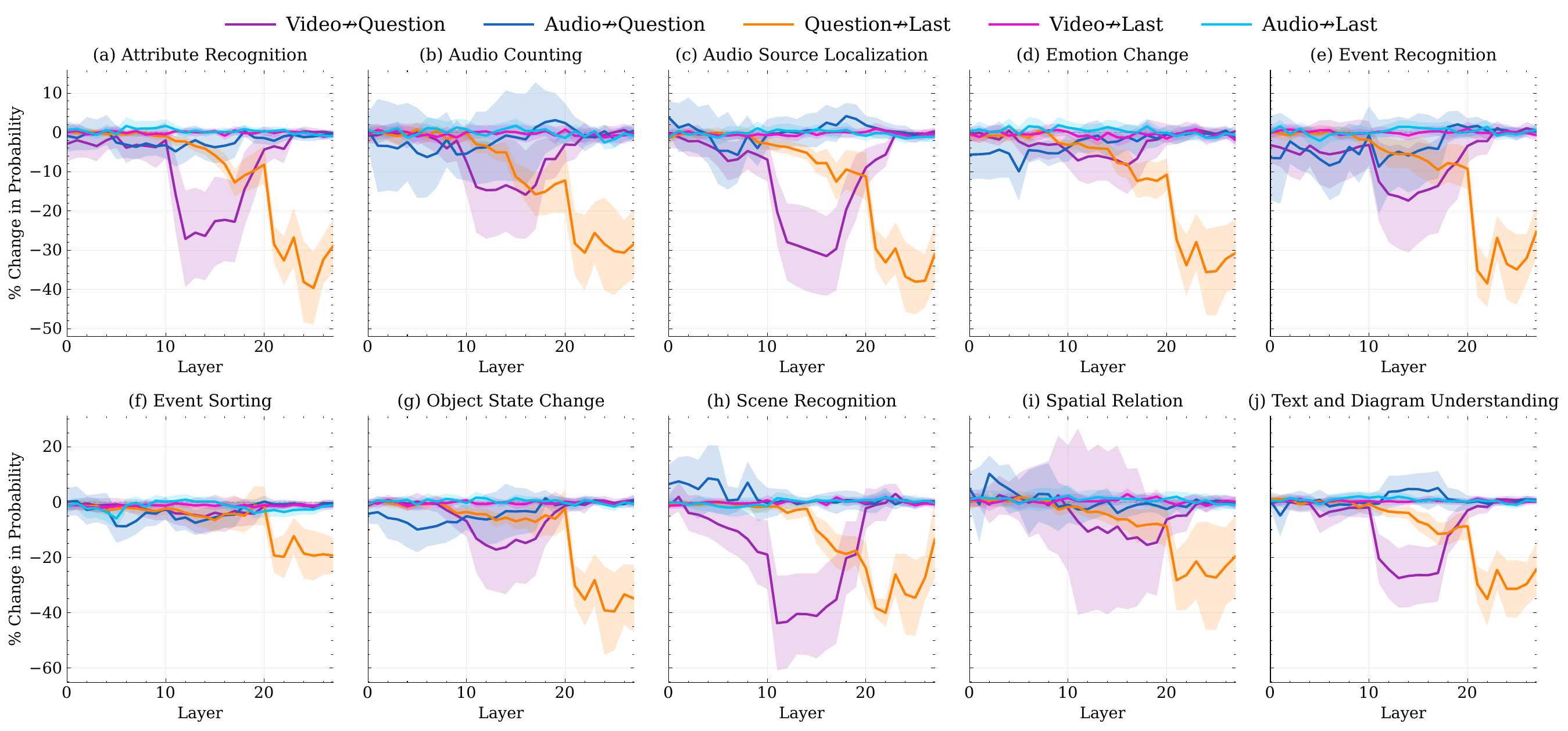}
  \caption{\textbf{Video-SALMONN2 7B Plus on WorldSense.} Knockout of within- and cross-modal pathways (Cross-frame, Cross-audio segment, Audio$\leftrightarrow$Video) and of modality and question pathways into the last token (Video$\not\to$Question, Audio$\not\to$Question, Question$\not\to$Last, Video$\not\to$Last, Audio$\not\to$Last). Source$\not\to$Target indicates blocking attention edges from source tokens to target tokens.}
  \label{fig:salmonn7B_worldsense_video_knock_group1}
\end{figure}

\begin{figure}[htbp]
  \centering
  \includegraphics[width=\linewidth]{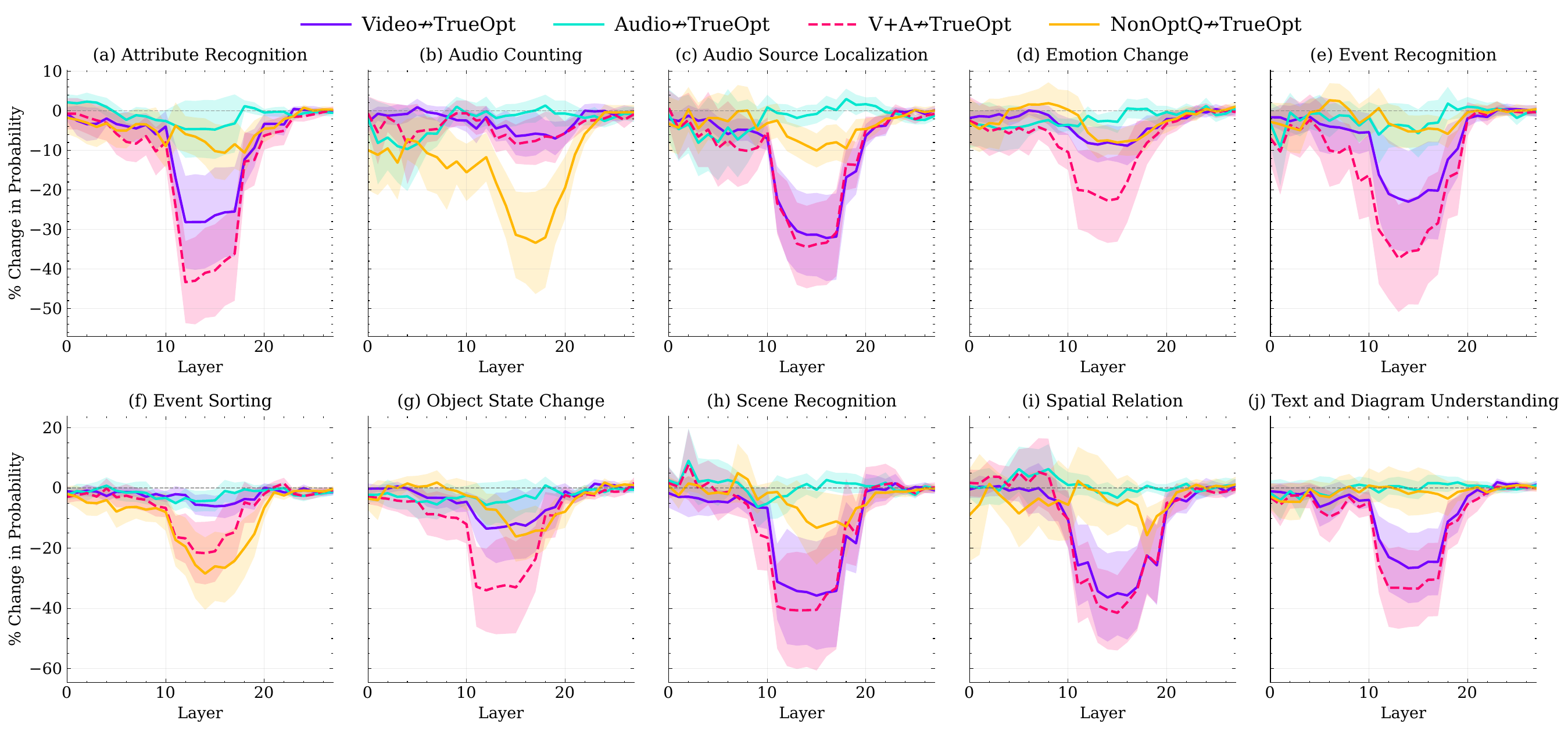}\\[4pt]
  \includegraphics[width=\linewidth]{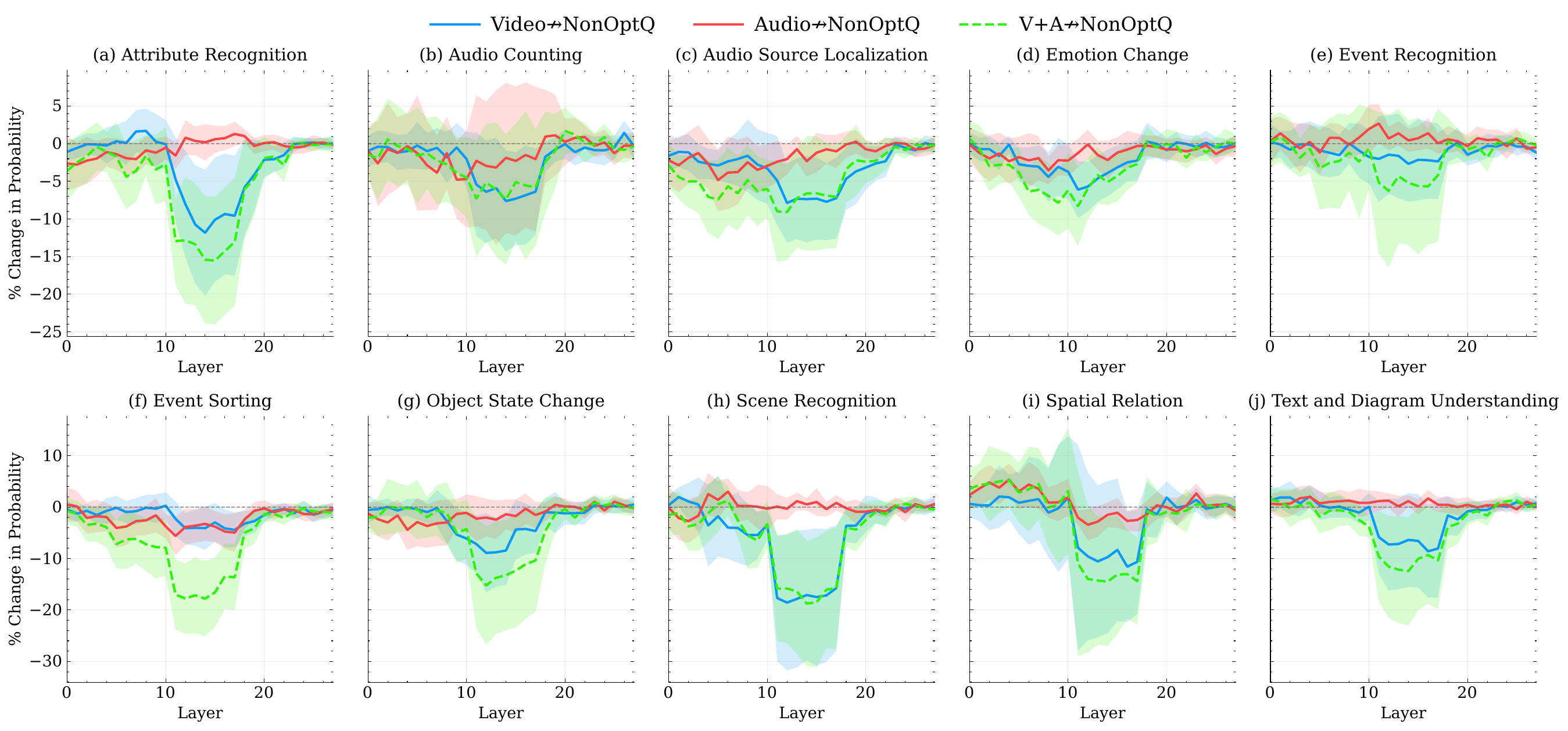}\\[4pt]
  \includegraphics[width=\linewidth]{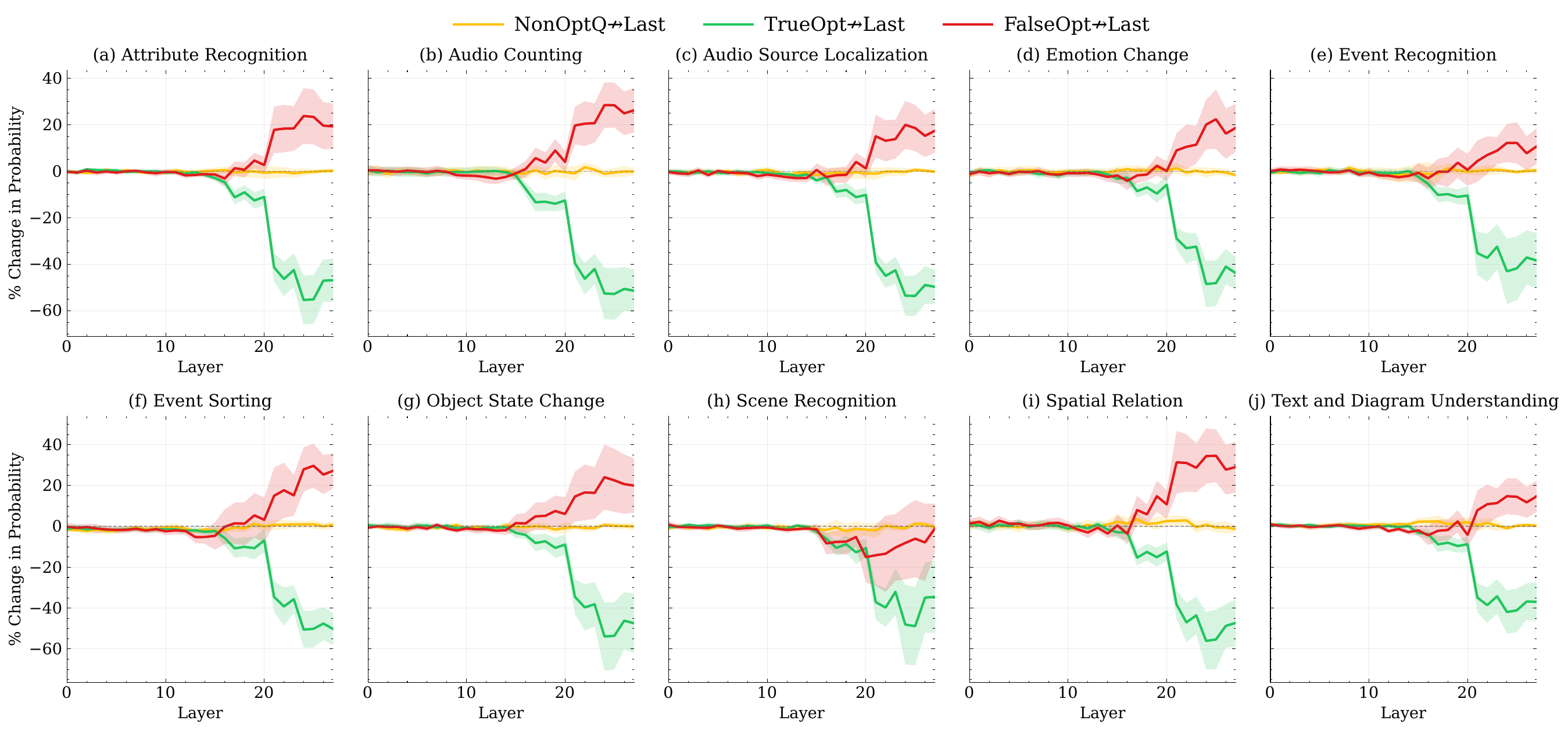}
  \caption{\textbf{Video-SALMONN2 7B Plus on WorldSense.} Modality and question pathways into the correct option letter (Video$\not\to$TrueOpt, Audio$\not\to$TrueOpt, NonOptQ$\not\to$TrueOpt, V+A$\not\to$TrueOpt); modality pathways into the non-option question text (Video$\not\to$NonOptQ, Audio$\not\to$NonOptQ, V+A$\not\to$NonOptQ); and question-internal pathways into the last token (TrueOpt$\not\to$Last, FalseOpt$\not\to$Last, NonOptQ$\not\to$Last). Source$\not\to$Target indicates blocking attention edges from source tokens to target tokens.}
  \label{fig:salmonn7B_worldsense_video_knock_group2}
\end{figure}

Figures~\ref{fig:salmonn7B_worldsense_video_knock_group1} and~\ref{fig:salmonn7B_worldsense_video_knock_group2} report the WorldSense knockout results. Figure~\ref{fig:salmonn7B_worldsense_video_knock_group1} covers within- and cross-modal pathways together with modality and question pathways into the last token, while Figure~\ref{fig:salmonn7B_worldsense_video_knock_group2} covers the question-internal pathways and the pathways into the correct option letter and the non-option question text.

\newpage
\section*{NeurIPS Paper Checklist}

\begin{enumerate}

\item {\bf Claims}
    \item[] Question: Do the main claims made in the abstract and introduction accurately reflect the paper's contributions and scope?
    \item[] Answer: \answerYes{} 
    \item[] Justification: The abstract and introduction state the main claims, which are reflected in Sections 4, 5, and 6.  

\item {\bf Limitations}
    \item[] Question: Does the paper discuss the limitations of the work performed by the authors?
    \item[] Answer: \answerYes{} 
    \item[] Justification: In Section 7 discusses the limitations scope of the study. 

\item {\bf Theory assumptions and proofs}
    \item[] Question: For each theoretical result, does the paper provide the full set of assumptions and a complete (and correct) proof?
    \item[] Answer: \answerNA{}  
    \item[] Justification: The paper does not include theoretical results. 

    \item {\bf Experimental result reproducibility}
    \item[] Question: Does the paper fully disclose all the information needed to reproduce the main experimental results of the paper to the extent that it affects the main claims and/or conclusions of the paper (regardless of whether the code and data are provided or not)?
    \item[] Answer: \answerYes{} 
    \item[] Justification: The main paper provides a comprehensive description of the experimental setup, and the Appendix includes the additional details necessary to reproduce the main experimental results. 

\item {\bf Open access to data and code}
    \item[] Question: Does the paper provide open access to the data and code, with sufficient instructions to faithfully reproduce the main experimental results, as described in supplemental material?
    \item[] Answer:  \answerNo{} 
    \item[] Justification: Code will be released publicly upon publication. All datasets and models used in this work are already publicly available and properly cited. The experimental methodology is described in Section 4, and full experimental details, including model checkpoints, inference setup, dataset processing, and task selection, are provided in the Appendix to support reproduction. 

\item {\bf Experimental setting/details}
    \item[] Question: Does the paper specify all the training and test details (e.g., data splits, hyperparameters, how they were chosen, type of optimizer) necessary to understand the results?
    \item[] Answer:   \answerYes{} 
    \item[] Justification: Experimental details are described across the paper, and specifically in the Appendix on Experimental Details. 

\item {\bf Experiment statistical significance}
    \item[] Question: Does the paper report error bars suitably and correctly defined or other appropriate information about the statistical significance of the experiments?
    \item[] Answer:  \answerNo{} 
    \item[] Justification: The experiments use deterministic decoding on the full dataset in a single execution, so no statistical variability is reported. 

\item {\bf Experiments compute resources}
    \item[] Question: For each experiment, does the paper provide sufficient information on the computer resources (type of compute workers, memory, time of execution) needed to reproduce the experiments?
    \item[] Answer:  \answerYes{} 
    \item[] Justification: Compute resources are described in the Experimental Details appendix. 
    
\item {\bf Code of ethics}
    \item[] Question: Does the research conducted in the paper conform, in every respect, with the NeurIPS Code of Ethics \url{https://neurips.cc/public/EthicsGuidelines}?
    \item[] Answer: \answerYes{} 
    \item[] Justification: The paper does not deal with any research that might deviate from the NeurIPS
code of ethics 

\item {\bf Broader impacts}
    \item[] Question: Does the paper discuss both potential positive societal impacts and negative societal impacts of the work performed?
    \item[] Answer: \answerNA{}  
    \item[] Justification: The paper does not introduce new models or technologies with potential malicious uses. 
    
\item {\bf Safeguards}
    \item[] Question: Does the paper describe safeguards that have been put in place for responsible release of data or models that have a high risk for misuse (e.g., pre-trained language models, image generators, or scraped datasets)?
    \item[] Answer: \answerNA{}. 
    \item[] Justification: The paper does not release any models or datasets that pose risks of misuse. 

\item {\bf Licenses for existing assets}
    \item[] Question: Are the creators or original owners of assets (e.g., code, data, models), used in the paper, properly credited and are the license and terms of use explicitly mentioned and properly respected?
    \item[] Answer: \answerYes{} 
    \item[] Justification: All models and datasets used are publicly available and properly cited. The assets are released under the Qwen Research License, Apache 2.0, and CC BY-NC 4.0. Each asset is used in accordance with its license and intended research use. 

\item {\bf New assets}
    \item[] Question: Are new assets introduced in the paper well documented and is the documentation provided alongside the assets?
    \item[] Answer: \answerNA{} 
    \item[] Justification: We are not releasing new assets. 

\item {\bf Crowdsourcing and research with human subjects}
    \item[] Question: For crowdsourcing experiments and research with human subjects, does the paper include the full text of instructions given to participants and screenshots, if applicable, as well as details about compensation (if any)? 
    \item[] Answer: \answerNA{} 
    \item[] Justification: The paper does not deal in research with human subjects. 

\item {\bf Institutional review board (IRB) approvals or equivalent for research with human subjects}
    \item[] Question: Does the paper describe potential risks incurred by study participants, whether such risks were disclosed to the subjects, and whether Institutional Review Board (IRB) approvals (or an equivalent approval/review based on the requirements of your country or institution) were obtained?
    \item[] Answer: \answerNA{} 
    \item[] Justification: The paper does not deal in research with human subjects. 

\item {\bf Declaration of LLM usage}
    \item[] Question: Does the paper describe the usage of LLMs if it is an important, original, or non-standard component of the core methods in this research? Note that if the LLM is used only for writing, editing, or formatting purposes and does \emph{not} impact the core methodology, scientific rigor, or originality of the research, declaration is not required.
    \item[] Answer: \answerNA{} 
    \item[] Justification: We used LLM only for writing, editing, or formatting purposes, and does not impact the core methodology.  

\end{enumerate}

\end{document}